\documentclass[conference]{IEEEtran}
\IEEEoverridecommandlockouts

\usepackage{cite}
\usepackage{amsmath,amssymb,amsfonts}
\usepackage{algorithmic}
\usepackage{graphicx}
\usepackage{textcomp}
\usepackage{xcolor}
\def\BibTeX{{\rm B\kern-.05em{\sc i\kern-.025em b}\kern-.08em
    T\kern-.1667em\lower.7ex\hbox{E}\kern-.125emX}}

\usepackage{xcolor}
\usepackage{amsfonts}
\usepackage{dsfont}
\usepackage{subcaption}
\usepackage{multirow}
\usepackage{stmaryrd}
\usepackage{booktabs}
\usepackage{soul}
\usepackage{placeins}
\usepackage{wrapfig}
\usepackage{varwidth}
\usepackage{hyperref}
\usepackage{url}

\usepackage{amsmath}
\usepackage{amssymb}
\usepackage{mathtools}
\usepackage{amsthm}

\begin{document}
\def\nexamples{N}
\def\nattributes{M}
\def\attributes[#1]{\mathbf{x}_{#1}}
\def\classes{\mathcal{C}}
\def\class{c}
\def\oheclass[#1,#2]{z_{#2#1}}
\def\example{k}
\def\feature{i}
\def\singleattribute[#1]{f_{#1}} %
\def\classset[#1]{Z_{#1}}
\def\dataset{D}

\def\lexicorder{\leq_L}
\def\ohevects{vects}
\def\ohegroup{w}

\def\forest{\mathcal{T}}
\def\tree{t}
\def\depth{d}
\def\node{v}
\def\noisevar{Y}
\def\sumnoisevars{\zeta} %
\def\query{q}
\def\Laplace{\mathcal{L}}
\def\noiseval{l}

\def\internalnodes[#1]{\smash{\mathcal{V}_{#1}^{I}}}
\def\leaves[#1]{\smash{\mathcal{V}_{#1}^{L}}}
\def\leftchild[#1]{l(#1)}
\def\rightchild[#1]{r(#1)}
\def\nodesupport[#1,#2,#3]{n_{#1#2#3}} %
\def\nodesforfeat[#1,#2]{\mathcal{V}_{#1#2}^{I}}
\def\positivesplits[#1,#2]{\Phi_{#2}^+} %
\def\negativesplits[#1,#2]{\Phi_{#2}^-} %
\def\varlambda[#1,#2,#3]{\lambda_{#1#2#3}}
\def\vary[#1,#2,#3]{y_{#1#2#3}}
\def\varyb[#1,#2,#3,#4]{y_{#1#2#3#4}}
\def\varx[#1,#2]{x_{#1#2}}
\def\varz[#1,#2]{z_{#1#2}}
\def\varq[#1,#2,#3]{q_{#1#2#3}}
\def\varqalt[#1,#2,#3,#4]{q_{#1#2#3#4}}
\def\attrib[#1]{f_{#1}}
\def\lb[#1]{l_{#1}}
\def\ub[#1]{u_{#1}}
\def\root[#1]{r_{#1}}
\def\parent[#1]{f_{#1}}
\def\treepath[#1,#2]{P^{#1}_{#2}}
\def\fixedatt[#1]{\phi(#1)}
\def\sensitivity#1{S(#1)}

\def\Z{\mathbb{Z}}
\def\np{\mathcal{NP}}

\def\sklearn{\texttt{scikit-learn}}
\def\ortools{\texttt{OR-Tools}}
\def\gurobi{\texttt{Gurobi}}
\def\jul#1{\textcolor{orange}{#1}}

\def\probname{DRP}
\def\mlprobname{MLDRP}

\def\scipy{\texttt{Scipy}}
\def\draft{DRAFT}

\def\updated#1{\textcolor{black}{#1}}
\def\rebuttalsatml#1{\textcolor{black}{#1}}
\def\rebuttalsatmlmultiline{\color{black}}

\title{Training Set Reconstruction from Differentially Private Forests: How Effective is DP?}

\author{\IEEEauthorblockN{1\textsuperscript{st} Alice Gorgé$^{\ast}$}
\IEEEauthorblockA{\textit{École Polytechnique} \\
Palaiseau, France \\
alice.gorge@polytechnique.edu}
\and
\IEEEauthorblockN{1\textsuperscript{st} Julien Ferry$^{\ast}$}
\IEEEauthorblockA{\textit{Polytechnique Montréal} \\
Montréal, Canada \\
julien.ferry@polymtl.ca}
\and
\IEEEauthorblockN{3\textsuperscript{rd} Sébastien Gambs}
\IEEEauthorblockA{\textit{Université du Québec à Montréal} \\
Montréal, Canada \\
gambs.sebastien@uqam.ca}
\and
\IEEEauthorblockN{4\textsuperscript{th} Thibaut Vidal}
\IEEEauthorblockA{\textit{Polytechnique Montréal} \\
Montréal, Canada \\
thibaut.vidal@polymtl.ca}
\thanks{$^{\ast}$Alice Gorgé and Julien Ferry contributed equally to this work.}%
\thanks{This work has been accepted for publication at the 2026 IEEE Conference on Secure and Trustworthy Machine Learning (SaTML). The final version will be available on IEEE Xplore.}
}

\maketitle

\begin{abstract}
Recent research has shown that structured machine learning models such as tree ensembles are vulnerable to privacy attacks targeting their training data. 
To mitigate these risks, differential privacy (DP) has become a widely adopted countermeasure, as it offers rigorous privacy protection.  
In this paper, we introduce a reconstruction attack targeting state-of-the-art $\varepsilon$-DP random forests. 
By leveraging a constraint programming model that incorporates knowledge of the forest's structure and DP mechanism characteristics, our approach formally reconstructs the \emph{most likely dataset} that could have produced a given forest.
Through extensive computational experiments, we examine the interplay between model utility, privacy guarantees and reconstruction accuracy across various configurations.
Our results reveal that random forests trained with meaningful DP guarantees can still leak portions of their training data. 
Specifically, while DP reduces the success of reconstruction attacks, the only forests fully robust to our attack exhibit predictive performance no better than a constant classifier.
Building on these insights, we also provide practical recommendations for the construction of DP random forests that are more resilient to reconstruction attacks while maintaining a non-trivial predictive performance.
\end{abstract}


\section{Introduction}
The success of machine learning algorithms in critical applications, such as medicine, kidney attribution and credit scoring \cite{dildar2021skin,DBLP:conf/aaai/0001CDM21,DBLP:journals/asc/DastileCP20}, hinges on access to large volumes of sensitive data, including medical records, personal histories and financial information. 
This extensive data is crucial for training models capable of achieving high accuracy. 
At the same time, trustworthiness in these applications demands transparency, as emphasized by regulations like Article 13 of the EU AI Act.\footnote{\href{https://artificialintelligenceact.eu/article/13/}{https://artificialintelligenceact.eu/article/13/}}
Transparency involves the ability to audit models and explain their outputs and inner workings, which fosters public confidence and ensures compliance with legal requirements.

However, transparency can inadvertently create vulnerabilities by exposing new attack surfaces. 
When models are made public, even via black-box APIs, they become susceptible to privacy attacks that can infer sensitive information about their training data~\cite{rigaki2023}. 
Recent studies on reconstruction attacks~\cite{dwork2017exposed} demonstrated that large language models~\cite{carlini2021}, diffusion models~\cite{DBLP:conf/uss/CarliniHNJSTBIW23}, neural networks~\cite{DBLP:conf/nips/HaimVYSI22}, random forests (RFs)~\cite{ferry2024trained} and a wide range of interpretable models~\cite{ferry2024probabilistic} inherently leak part of their training data through their structure or outputs.

To mitigate such privacy risks, \emph{differential privacy} (DP)~\cite{dwork2014algorithmic} has emerged as the \emph{de facto} standard, as it limits the amount of information an adversary can gain regarding any single individual in the dataset. 
DP theoretically bounds the success of membership inference attacks (which aim at inferring the absence or presence of a given individual in a model's training data), but also empirically provides a form of protection against other attacks. 
For instance, recent work on deep learning for medical imaging~\cite{ziller2024reconciling} has shown that DP mitigates the success of some reconstruction attacks, even for loose privacy guarantees. 
However, DP imposes significant trade-offs as ensuring privacy involves adding noise or other modifications that degrade a model's predictive performance. 
Consequently, practitioners must carefully balance privacy guarantees and model utility in sensitive domains.

In this paper, we focus on reconstruction attacks targeting tree ensembles, such as RFs. 
These models frequently achieve state-of-the-art performance on tabular data, as evidenced in recent works~\cite{borisov2022deep} and competitions~\cite{bojer2021kaggle}. 
Furthermore, they are widely used in real-world deployments as they offer competitive storage costs and inference times. 
Their structured nature makes them particularly suitable to model audits, further facilitating their use in sensitive contexts.
However, recent research~\cite{ferry2024trained} has revealed that their structure can also inherently expose their training data. 
While several DP tree-based models have been designed~\cite{fletcher2019decision}, it remains unclear how well DP can effectively prevent such privacy leaks.

To address this gap, our setting is designed to investigate the following fundamental question: under what conditions can a trained model be shared in a white-box manner without compromising the privacy of its training data? 
By considering state-of-the-art machine learning models for tabular data (RFs),  alongside standard privacy protection mechanisms (pure $\varepsilon$-DP using the Laplace mechanism and basic composition), our results establish meaningful milestones for addressing this question. More precisely, our main contributions are:
\begin{itemize}
    \item We introduce an effective reconstruction attack against state-of-the-art DP RFs.
    Building on the attack paradigm pioneered by~\cite{ferry2024trained}, our approach frames the reconstruction attack as a %
    Constraint Programming (CP) model, which identifies a \emph{most likely dataset} capable of generating the observed RF and noisy counts.
    \item Through extensive experiments, we show that DP RFs \rebuttalsatml{satisfying pure $\varepsilon$-DP guarantees} can still be exploited to retrieve portions of their training data. 
    In particular, all investigated DP RFs with non-trivial predictive performance reveal information about their training data, often extending beyond distributional patterns to specificities of the dataset such as complete training examples.
    The source code of our method
    is available at \url{https://github.com/vidalt/DRAFT-DP} under an MIT license.
    \item We provide key insights %
    regarding the factors influencing the trade-offs between DP RFs' predictive performance and their vulnerability to reconstruction attacks. 
    These insights aim to guide practitioners in effectively deploying DP RFs in real-world scenarios. 
\end{itemize}

The remainder of the paper is organized as follows. 
First, in Section~\ref{sec:preliminaries}, we provide the necessary background on DP and describe the considered DP RFs algorithm. 
Then, we introduce our reconstruction attack model and formulation in Section~\ref{sec:attack}. 
Afterwards, we empirically evaluate its effectiveness in Section~\ref{sec:experiments}. 
Finally, we review the relevant related works on reconstruction attacks, post-processing of DP values and DP RFs in Section~\ref{sec:related_work} before concluding and discussing promising research avenues in Section~\ref{sec:conclusion}.

\section{Background Notions}
\label{sec:preliminaries}

\subsection{Differential Privacy}\label{sec:dp}

Originally formalized in~\cite{dwork2006differential}, DP enables the computation of useful statistics on private data while strictly limiting the information that can be inferred about any single individual.  
A randomized mechanism \(\mathcal{K}\) satisfies \( \varepsilon \)-DP if, for any pair of datasets \( \dataset_1 \) and \( \dataset_2 \) differing by the addition or removal of at most one record, and all \( U \subseteq \text{Range}(\mathcal{K}) \):
\[
\mathbb{P}(\mathcal{K}(\dataset_1) \in U) \leq e^\varepsilon \times \mathbb{P}(\mathcal{K}(\dataset_2) \in U).
\]
The parameter \( \varepsilon \), called the privacy budget, quantifies the trade-off between privacy and utility. 
Smaller values of \( \varepsilon \) ensure stronger privacy at the expense of reduced utility.

Let \(\query\) represent a deterministic query such as counts, sums, averages, or, in the context of tree ensembles, the number of samples of a particular class in a data subset. 
DP can give protected access to this query through a probabilistic mechanism \(\mathcal{K}\), which outputs \(\query\) with additional noise. 
As detailed in~\cite{dwork2006calibrating}, the amount of noise needed to protect the data depends on the \emph{global sensitivity} of the query, defined as
\(
\sensitivity{\query} = \max ||\query(\dataset_1) - \query(\dataset_2)||_1,
\)
in which the maximum is taken over all pairs of datasets $\dataset_1$ and $\dataset_2$ differing in at most one element. 
Thus, \(\sensitivity{\query}\) quantifies the largest possible change in the query output due to a single record's addition or removal (\emph{e.g.}, $\sensitivity{\query} = 1$ for counts).
For instance, the Laplace mechanism, a widely used DP method, adds noise sampled from a Laplace distribution:
\[
\mathcal{K}(\dataset) = \query(\dataset) + (\noisevar_1, \ldots, \noisevar_k),
\]
in which \( \noisevar_i \) are random variables i.i.d. following the Laplace distribution \( \text{Lap}(\sensitivity{\query} / \varepsilon) \) with $\varepsilon$ the privacy budget granted to the mechanism.
When multiple DP mechanisms are combined, the overall privacy budget is governed by two fundamental composition theorems~\cite{mcsherry2007mechanism}.
First, \emph{sequential composition} states that applying several \( \varepsilon_i \)-DP mechanisms to the same dataset results in \( \sum_i \varepsilon_i \)-DP.
In contrast, \emph{parallel composition} states that applying several \( \varepsilon_i \)-DP mechanisms to disjoint subsets of data results in  \( \max_i \varepsilon_i \)-DP.
Finally, the \emph{postprocessing property} ensures that any function applied to the output of \(\mathcal{K}\) does not affect the DP guarantee, providing flexibility for downstream analysis.

DP mechanisms have been widely adopted to protect privacy when releasing machine learning models.
For example, noise can be added to gradients during neural network training~\cite{Abadi_2016}, or split thresholds can be perturbed in decision tree predictors~\cite{fletcher2019decision}. 
In particular, DP can mitigate privacy risks such as membership inference or model inversion~\cite{DBLP:journals/tifs/YeSZLZ22}, 
and implementations of DP mechanisms have been integrated within popular libraries, such as IBM's \texttt{diffprivlib}~\cite{DBLP:journals/corr/abs-1907-02444} or Google's \texttt{differential-privacy}~\cite{googledp}. %

\subsection{Differentially-Private Random Forests}
\label{sec:dp_rf}

We consider a training set $\dataset = {\{\attributes[\example];\class_{\example}\}}^{\nexamples}_{\example=1}$ in which each example $\example$ is characterized by a vector $\smash{\attributes[\example] \in \{0,1\}^{\nattributes}}$ of $\nattributes$ binary attributes and a class $\class_{\example} \in \classes$. 
Categorical features can be transformed into this format using one-hot encoding, as is standard practice in tree ensemble training, while continuous features can be discretized through binning.
Based on this dataset, a random forest \(\forest\) is built as an ensemble of decision trees \(\tree \in \forest\). 
Each tree consists of a set of internal nodes \(\internalnodes[\tree]\), in which each node $\node \in \internalnodes[\tree]$ contains a binary condition on the value of an attribute, and a set of leaves \(\leaves[\tree]\). 
When an example is evaluated, it traverses the tree by descending to the left child \(\leftchild[\node]\) if the condition is satisfied, or to the right child \(\rightchild[\node]\) otherwise. 
Once a leaf \(\node \in \leaves[\tree]\) (terminal node) is reached, it contributes to the class distribution associated with that leaf, which is determined by the per-class counts of training examples reaching the leaf.

In most random forest implementations, including popular libraries such as \texttt{scikit-learn}~\cite{scikit-learn}, predictions are made using \emph{soft voting}: each tree outputs class probabilities based on the normalized class counts in its leaves and the ensemble prediction averages these probabilities.
Specifically, for each leaf~\(\node \in \leaves[\tree]\) in tree $\tree$, the model stores~\(\nodesupport[\tree,\node,\class]\), the number of training examples of class~\(\class\) that reach~\(\node\). 
This information is essential for on-device inferences and audits but also introduces risks, as it exposes model parameters to potential adversaries~\cite{ferry2024trained}.

\begin{figure}
  \centering
    \includegraphics[width=0.95\linewidth]{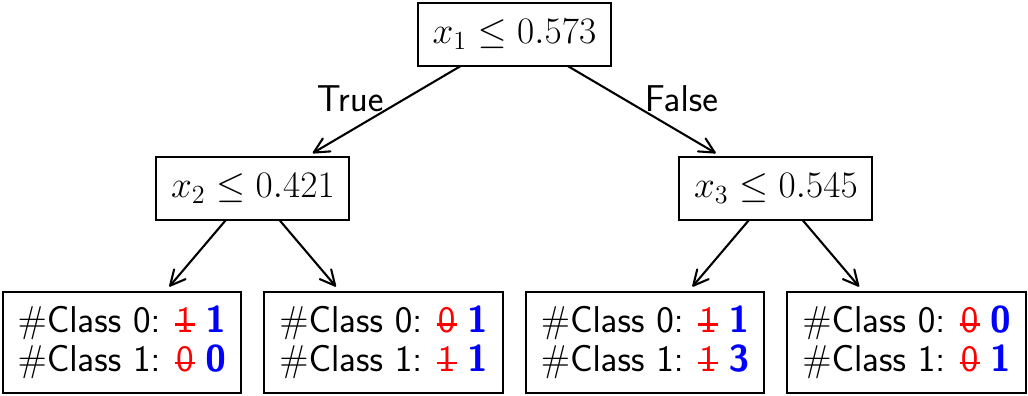}
  \caption{Example decision tree $\tree$, {\color{red}\st{before}} and {\color{blue}\textbf{after}} adding Laplace noise to comply with 1-DP. The rightmost leaf, originally empty, now reports a positive sample count.}
    \label{fig:comparaison}
\end{figure}

To address these risks, an extensive body of research has focused on constructing DP RFs to conciliate model accessibility and privacy protection, with many of these seminal methods being surveyed by
\cite{fletcher2019decision}.
Among these, the method proposed by~\cite{jagannathan2012enabling}, which combines randomized tree structure and noisy counts publication, was the first to achieve strong empirical predictive performance and formal DP guarantees. 
Consequently, we use it as a foundation of our experimental study.
In this approach, each tree is trained on the complete dataset~$\dataset$.
Each tree of a chosen depth $\depth$ is grown by randomly selecting attributes and split values for all internal nodes \(\node \in \internalnodes[\tree]\), avoiding direct access to $\dataset$ during this phase to preserve the DP budget.
Subsequently, the class counts~\(\nodesupport[\tree,\node,\class]\) at each leaf \(\node \in \leaves[\tree]\) are computed using $\dataset$. Finally, they are obfuscated using the Laplace mechanism~\cite{dwork2006calibrating} and cast to integer values, as illustrated in Figure~\ref{fig:comparaison}.

Let $\varepsilon$ denote the total privacy budget. 
Since all trees $\tree \in \forest$ are trained on the same dataset $\dataset$, sequential composition applies, %
such that each tree has a budget of $\varepsilon_{\tree} = \varepsilon / |\forest|$.
Since the attributes and split values within each tree are chosen randomly without accessing the training data, the noisy computation of class counts is the sole operation consuming the DP budget. 
Within each tree, \emph{parallel composition} applies since the leaves have disjoint supports.
Similarly, within a leaf, the counting operations for each class are independent, also satisfying parallel composition. 

Accordingly, the DP budget allocated to publish each class count $ \nodesupport[\tree,\node,\class]$  for $\tree \in \forest, \node \in \leaves[\tree], \class \in \classes$ is $\varepsilon_\node = \varepsilon_\tree = \varepsilon / \vert \forest \rvert$. 
Since class counts have a global sensitivity $\sensitivity{\query}=1$, the noisy class counts are computed~as:
\[
\nodesupport[\tree,\node,\class]^* = \textrm{int}(\nodesupport[\tree,\node,\class] + \noisevar_{\tree\node\class}) = \nodesupport[\tree,\node,\class]  +  \textrm{int}( \noisevar_{\tree\node\class}) ,
\]
in which $\noisevar_{\tree\node\class}$ is a random variable sampled from a Laplace distribution \( \text{Lap}(1 / \varepsilon_\node)\) and
$\textrm{int}(.)$ is the integer part function. \rebuttalsatml{Appendix~\ref{appendix:design_dp_rf_algo} provides qualitative and quantitative evidence that the Laplace mechanism is preferable to approximate DP for privatizing per-class per-leaf counts in this setting.}

\section{Reconstruction Attack against DP RFs}
\label{sec:attack}

\textbf{Adversary model.} To the best of our knowledge, our attack is \rebuttalsatml{\emph{the first method to target full dataset reconstruction against DP-protected models}}. We assume \emph{white-box} access to an $\varepsilon$-DP RF constructed using the algorithm of~\cite{jagannathan2012enabling}, as described in Section~\ref{sec:dp_rf}. 
This includes access to the structure of the forest (split attributes and values) and the (noisy) counts at the leaves. 
Consistent with the literature on reconstruction attacks~\cite{dwork2017exposed,ferry2024trained}, we also assume knowledge of the number of attributes $\nattributes$ and training examples $\nexamples$.
While the knowledge of $\nattributes$ is necessary for inference, the knowledge of $\nexamples$ simplifies the attack but is not strictly required.
We briefly discuss how the proposed CP formulation can be adapted to handle this setting at the end of this section, and provide the detailed modified formulation in Appendix~\ref{appendix:cp_model_n_unknown}.
We also assume the privacy budget~$\varepsilon$ is known. 
This does not degrade the DP guarantees, is generally disclosed by practitioners\footnote{For instance, \cite{desfontainesblog20211001} documents real-world applications of DP and their associated budgets.}, and researchers have advocated this for transparency~\cite{DBLP:journals/jpc/DworkKM19}.

\vspace{0.25\baselineskip}

\textbf{Reconstruction approach.} Our approach is grounded on the CP formulation introduced in~\cite{ferry2024trained}, with additional constraints, variables and an objective designed to model the (observed) noisy counts, the underlying (guessed) true counts and their likelihood.
Solving the original problem formulation (DRAFT) was shown to be NP-Hard.
Our reconstruction problem for DP RFs can be seen as a generalization (\emph{i.e.}, DRAFT is a special case with no noise) and is also NP-hard, though scalable with state-of-the-art CP solvers~\cite{cpsatlp}.

Our formulation revolves around three main groups of \emph{decision variables}. 
The objective of a CP solver is to find compatible variable values that satisfy a given set of \emph{constraints}, yielding a feasible solution. 
Moreover, as explained later in this section, an \emph{objective function} steering the search towards likely solutions (reconstructions) will be included.
First, for each tree $\tree \in \forest$, the (guessed) true number of examples of class $\class \in \mathcal{\classes}$ in each leaf $\node \in \leaves[\tree]$ are encoded using a variable $\nodesupport[\tree,\node,\class] \in \mathbb{N}$. 
Next, a set of variables encodes each reconstructed example $\example \in \{1,\dots,\nexamples\}$: $\varx[\example,\feature] \in \{ 0,1 \}$ encodes the value of its $i^{\text{th}}$ attribute $\feature \in \{1,\dots,\nattributes\}$, while $\varz[\example,\class] \in \{ 0,1 \}$ indicates if it belongs to class $\class \in \classes$.
Third, the path of each example $\example \in \{1,\dots,\nexamples\}$ through each tree $\tree \in \forest$ is modelled using a binary variable $\varyb[\tree,\node,\example,\class] \in \{0,1\}$, which is $1$ if it is classified by leaf $\node \in \leaves[\tree]$ as class $\class \in \classes$, and $0$ otherwise.
The following constraints linking these variables are then enforced:
\vspace{-1.25em}
\begin{center}
\begin{varwidth}{0.47\textwidth}
        \begin{align}
            &\sum\limits_{\node \in \leaves[\tree]} \sum\limits_{\class \in \classes} \nodesupport[\tree,\node,\class] = \nexamples &\tree \in \forest \label{constr_count_sum_N}\\
            &\sum\limits_{\class \in \classes}\varz[\example,\class]=1 &\example \in \{1,\dots,\nexamples\}\label{constr_exactly_one_class}\\
            &\varz[\example,\class] =  \sum\limits_{\smash{\mathclap{\node \in \leaves[\tree]}}} \varyb[\tree,\node,\example,\class]  &   \example \in \{1,\dots,\nexamples\}, \tree \in \forest, \class \in \classes\label{constr_not_in_class_counts}\\
            & \sum_{\example=1}^{\smash{\nexamples}} \varyb[\tree,\node,\example,\class] =\nodesupport[\tree,\node,\class] &\tree \in \forest,~\node \in \leaves[\tree],~\class \in \classes. \label{constr_counts_flow}
        \end{align}
\end{varwidth}
\end{center}
\vspace{-0.25em}

In a nutshell, Constraint~\eqref{constr_count_sum_N} ensures that the example counts sum up to $\nexamples$ within each tree.
Constraint~\eqref{constr_exactly_one_class} guarantees that each example is associated to exactly one class.
Constraint~\eqref{constr_not_in_class_counts} enforces that each example contributes only to the counts of its assigned class.
Moreover, Constraint~\eqref{constr_counts_flow} ensures consistency between the counts at each leaf and the examples assigned to it. 
Finally, for each leaf $\node \in \leaves[\tree]$ of each tree $\tree \in \forest$, let $\Phi_\node^+$ (respectively $\Phi_\node^-$) be the set of binary attributes that must equal $1$ (respectively $0$) for an example to fall into $\node$. 
Then, the following constraint enforces consistency between the examples' assignments to the leaves and the value of their attributes:
For $\tree \in \forest,~\node \in \leaves[\tree],~\example \in \{1,\dots,\nexamples\},$
\begin{align}
\sum\limits_{\class \in \classes} \varyb[\tree,\node,\example,\class] = 1 \Rightarrow &\left(\displaystyle\bigwedge_{i\in\Phi_v^+} \varx[k,i] = 1 \right) \wedge
\left( \displaystyle\bigwedge_{i\in \Phi_v^-} \varx[k,i] = 0 \right). \label{constr_flow_attr_values}
\end{align}
Optionally, if additional structural knowledge about the features is available, \emph{e.g.}, one-hot encoding or logical/linear relationships between their values, this information can be incorporated by adding the corresponding constraints on the~$\varx[k,i]$ values for each sample $k$.

The aforementioned variables and constraints define the \textbf{domain of all possible datasets and associated counts~$\nodesupport[\tree,\node,\class]$ compatible with the structure and split values of an observed forest}. 
Among all solutions in this domain, we aim at retrieving the one that is the \emph{most likely} to have generated the observed noisy counts $\nodesupport[\tree,\node,\class]^{*}$ through the DP mechanism.
To this end, we formulate through Constraint~\eqref{constr_count_link_noise} the noise $\Delta_{\tree\node\class}$ that must have been added to each count to achieve the observed values.
\begin{align}
&\Delta_{\tree\node\class} = \nodesupport[\tree,\node,\class]^{*} - \nodesupport[\tree,\node,\class] & \tree \in \forest,~\node \in \leaves[\tree],~\class \in \classes \label{constr_count_link_noise}
\end{align}
We know that these noise values correspond to i.i.d realizations of random variables $\noisevar_{\tree\node\class}$ from a Laplace distribution \( \text{Lap}(1 / \varepsilon_\node)\) cast to integer. 
Given this information, the \emph{likelihood} of a complete solution associated to a set of $\Delta_{\tree\node\class}$ values is defined in Equation~\eqref{obj:nolog}. 
Maximizing this likelihood is equivalent to maximizing its logarithm, leading to the objective function in Equation~\eqref{eq:obj_f_th}.
\begin{minipage}[t]{0.47\textwidth}
        \begin{align}
          &\mathcal{P}_{\boldsymbol\Delta} = \prod\limits_{\tree,\node,\class} \mathbb{P}(\textrm{int}(\noisevar_{\tree\node\class}) = \Delta_{\tree\node\class}) \label{obj:nolog} \\
          &\max \log(\mathcal{P}_{\boldsymbol\Delta}) = \sum\limits_{\tree,\node,\class} \log(\mathbb{P}(\textrm{int}(\noisevar_{\tree\node\class}) = \Delta_{\tree\node\class}))\label{eq:obj_f_th}
        \end{align}
\end{minipage}%

The Laplace distribution density with parameter \((1 / \varepsilon_\node)\) is given by $\Laplace(u) = \frac{\varepsilon_\node}{2}e^{(-|u|\varepsilon_\node)}$. 
Since the noisy counts are cast to integer values, the probability of a specific (integer) noise value $\noiseval$ is:
\begin{equation*}
p_\noiseval = \mathbb{P}(\textrm{int}(\noisevar_{\tree\node\class}) = l) = 
\begin{cases}
\int_{\noiseval}^{\noiseval+1} \frac{\varepsilon_\node}{2} e^{-\lvert u \rvert \varepsilon_\node} \, du & \text{if } \noiseval > 0, \\
\int_{-1}^{1} \frac{\varepsilon_\node}{2} e^{-\lvert u \rvert \varepsilon_\node} \, du & \text{if } \noiseval = 0, \\
\int_{l-1}^{l} \frac{\varepsilon_\node}{2} e^{-\lvert u \rvert \varepsilon_\node} \, du & \text{if } \noiseval < 0.
\end{cases}
\end{equation*}

In practice, we pre-compute constants $p_\noiseval$ for all values $\noiseval \in \{ -\gamma,\dots, \gamma \}$ such that $\mathbb{P}(\textrm{int}(\noisevar_{\tree\node\class}) \in \{ -\gamma,\dots, \gamma \} ) \geq 0.999$. 
Details on the computation of $\gamma$ are provided in Appendix~\ref{appendix:width_delta_search_interval}, but typically, $\gamma = \lceil 12 / \varepsilon_\node \rceil$ is sufficient.
With this, we can additionally restrict the feasible domain of the variables to $\nodesupport[\tree,\node,\class] \in \{\max(0,\nodesupport[\tree,\node,\class]^*-\gamma),\dots,\nodesupport[\tree,\node,\class]^*+\gamma\}$ and $\Delta_{\tree\node\class} \in \{ -\gamma,\dots, \gamma \}$ and formulate the objective as:
\begin{align}
\max \ \sum\limits_{\tree \in \forest}\sum\limits_{\node \in \leaves[\tree]} \sum\limits_{\class \in \classes} \sum\limits_{\noiseval = -\gamma}^{\gamma} \log(p_\noiseval) \mathds{1}_{\Delta_{\tree\node\class} = \noiseval},\label{eq:obj_f}
\end{align}
in which the indicator $\mathds{1}_{\Delta_{\tree\node\class} = \noiseval}$ can efficiently be expressed in CP using \emph{domain mapping} constraints based on the values of the $\Delta_{\tree\node\class}$ variables.

\vspace{0.25\baselineskip}
\textbf{Reconstruction approach when $\nexamples$ is unknown.} While the CP formulation introduced earlier assumes knowledge of the number of training examples $\nexamples$, it can actually be adapted to the more challenging case where $\nexamples$ is not known.
Appendix~\ref{appendix:cp_model_n_unknown} provides the full modified formulation along with a detailed empirical evaluation. 
The high-level idea is as follows. First, we leverage the knowledge of the distribution of the noise added to each individual leaf count to estimate the variance of their sum for each individual tree (whose true value without noise should be the unknown $\nexamples$). 
We consequently use the Central Limit Theorem to determine a $95$\% confidence interval $\smash{\llbracket \nexamples_{\textsc{min}}, \nexamples_{\textsc{max}} \rrbracket}$ for $\nexamples$, given the average sum of the noisy counts for the different trees in the forest.
In the modified CP formulation, we allow up to $\nexamples_{\textsc{max}}$ examples to be reconstructed, with the actual number determined by an additional integer variable $\smash{\widetilde{\nexamples}}\in{\nexamples_{\textsc{min}},\dots, \nexamples_{\textsc{max}}}$. 
In the encoded reconstruction, only examples $\example$ such that $\smash{\example \leq \widetilde{\nexamples}}$ contribute to the inferred leaf counts. 
The objective function~\eqref{eq:obj_f} then drives $\smash{\widetilde{\nexamples}}$ towards the true $\nexamples$ by maximizing the likelihood of the inferred noise values.

Finally, these modifications increase the size of the CP formulation search space, making the problem computationally more challenging. 
Nevertheless, our empirical evaluation shows that it can still be solved efficiently and achieves reconstruction errors comparable to those obtained when $\nexamples$ is known.

\section{Experimental Study}
\label{sec:experiments}

We now empirically evaluate our proposed reconstruction attack, and investigate whether it succeeds in accurately inferring a RF training data despite the DP protection.
The source code and data needed to reproduce all our experiments and figures
are available at~\url{https://github.com/vidalt/DRAFT-DP}. %

\subsection{Experimental Setup}
\label{subsec:setup}

\textbf{Datasets.} We rely on three datasets, using the same data preprocessing as in~\cite{ferry2024trained}. %
The COMPAS dataset~\cite{angwin2016machine} contains records of $7{,}206$ criminal offenders from Florida, each characterized by $15$ binary attributes, and is used for recidivism prediction.
The UCI Adult Income dataset~\cite{Dua:2019} gathers data on $48{,}842$ individuals from the 1994 U.S. census. 
Each individual is described by $20$ binary features, with the objective of predicting whether an individual earns more than \$50K/year. 
Finally, the Default of Credit Card Clients dataset~\cite{yeh2009comparisons} (Default Credit) contains $29{,}986$ customer records from Taiwan, each with $22$ binary attributes, and is used to predict payment defaults.

\vspace{0.25\baselineskip}

\textbf{Target models.} The DP RFs are trained as described in Section~\ref{sec:dp_rf}. \rebuttalsatml{In particular, we note that they satisfy pure $\varepsilon$-DP, which is therefore the focus of our experiments.}
In practice, we build upon the efficient and modular implementation of IBM's \texttt{diffprivlib} library~\cite{DBLP:journals/corr/abs-1907-02444}, which we slightly adapt for our experiments.
Following~\cite{ferry2024trained}, each experiment uses a randomly sampled training set of $\nexamples=100$ examples.
While this sample size is in line with the literature and allows all experiments to be conducted in a moderate time frame, we also report results with larger training datasets in Appendix~\ref{appendix:additional_experiments_scalability}, demonstrating the scalability of the approach. 
We build DP RFs of size $\lvert \forest \rvert \in \{1, 5, 10, 20, 30\}$ and of maximum depth $\depth \in \{3,5,7\}$.  
Empirical evidence from~\cite{fan2003is} suggests that the predictive accuracy of a random forest typically plateaus beyond $30$ trees, with $10$ to $30$ trees being sufficient for robust performance.
This is particularly true in the DP setting, in which adding trees consumes additional privacy budget~\cite{fletcher2019decision}. 
Finally, we evaluate privacy parameters $\varepsilon \in \{0.1, 1, 5, 10, 20, 30\}$, ranging from very tight to large privacy budgets, which reflect commonly used values in practice~\cite{desfontainesblog20211001}. 

\begin{table*}[t!]
\centering
\caption{Average reconstruction error for different numbers of trees $\lvert \forest \rvert$ 
, tree depths $\depth$ and privacy budgets $\varepsilon$ for the three datasets. Reconstruction errors worse than the random baseline are 
\textbf{bold}. 
Cases where the solver failed to find a feasible reconstruction within the 
time limit are marked with~``--''.}\label{tab:error-values}
\begin{tabular}{llccccccccc}
\toprule
                                &                     & \multicolumn{3}{c}{Adult} & \multicolumn{3}{c}{COMPAS} & \multicolumn{3}{c}{Default Credit} \\
\cmidrule(lr){3-5} \cmidrule(lr){6-8} \cmidrule(lr){9-11}
                                &                     & $d=3$   & $d=5$  & $d=7$  & $d=3$   & $d=5$   & $d=7$  & $d=3$      & $d=5$      & $d=7$     \\
\midrule
\multirow{6}{*}{$\lvert \forest \rvert = 1$}  & $\varepsilon = 0.1$ & \textbf{0.26} & 0.24 & 0.22 & \textbf{0.29} & \textbf{0.28} & \textbf{0.21} & \textbf{0.35} & \textbf{0.31} & \textbf{0.29} \\
& $\varepsilon = 1$ & 0.25 & 0.20 & 0.17 & \textbf{0.26} & 0.20 & 0.14 & \textbf{0.32} & \textbf{0.29} & 0.24 \\
& $\varepsilon = 5$ & 0.24 & 0.23 & \textbf{0.30} & \textbf{0.25} & 0.20 & 0.18 & \textbf{0.32} & \textbf{0.31} & 0.25 \\
& $\varepsilon = 10$ & 0.24 & 0.22 & \textbf{0.33} & \textbf{0.25} & \textbf{0.21} & 0.18 & \textbf{0.32} & \textbf{0.30} & \textbf{0.27} \\
& $\varepsilon = 20$ & 0.24 & 0.19 & \textbf{0.27} & \textbf{0.25} & 0.17 & 0.20 & \textbf{0.32} & \textbf{0.28} & \textbf{0.30} \\
& $\varepsilon = 30$ & 0.24 & 0.19 & \textbf{0.27} & \textbf{0.25} & 0.16 & 0.18 & \textbf{0.32} & \textbf{0.29} & \textbf{0.31} \\
\midrule
\multirow{6}{*}{$\lvert \forest \rvert = 5$}  & $\varepsilon = 0.1$ & \textbf{0.26} & 0.24 & 0.24 & \textbf{0.23} & \textbf{0.24} & \textbf{0.23} & \textbf{0.32} & \textbf{0.27} & \textbf{0.29} \\
& $\varepsilon = 1$ & 0.19 & 0.18 & 0.20 & 0.14 & 0.13 & 0.12 & 0.23 & 0.23 & 0.22 \\
& $\varepsilon = 5$ & 0.17 & 0.14 & 0.11 & 0.10 & 0.07 & 0.05 & 0.22 & 0.16 & 0.15 \\
& $\varepsilon = 10$ & 0.17 & 0.13 & 0.10 & 0.10 & 0.06 & 0.04 & 0.22 & 0.16 & 0.13 \\
& $\varepsilon = 20$ & 0.16 & 0.12 & 0.08 & 0.09 & 0.05 & 0.02 & 0.21 & 0.15 & 0.11 \\
& $\varepsilon = 30$ & 0.16 & 0.13 & 0.09 & 0.10 & 0.05 & 0.03 & 0.21 & 0.16 & 0.11 \\
\midrule
\multirow{6}{*}{$\lvert \forest \rvert = 10$}  & $\varepsilon = 0.1$ & \textbf{0.28} & \textbf{0.29} & \textbf{0.34} & \textbf{0.25} & \textbf{0.27} & \textbf{0.28} & \textbf{0.32} & \textbf{0.31} & \textbf{0.30} \\
& $\varepsilon = 1$ & 0.17 & 0.20 & 0.23 & 0.12 & 0.13 & 0.16 & 0.21 & 0.24 & 0.24 \\
& $\varepsilon = 5$ & 0.15 & 0.12 & 0.15 & 0.07 & 0.06 & 0.06 & 0.15 & 0.16 & 0.19 \\
& $\varepsilon = 10$ & 0.13 & 0.10 & 0.11 & 0.06 & 0.05 & 0.04 & 0.14 & 0.14 & 0.13 \\
& $\varepsilon = 20$ & 0.14 & 0.09 & 0.09 & 0.06 & 0.03 & 0.03 & 0.14 & 0.12 & 0.11 \\
& $\varepsilon = 30$ & 0.13 & 0.09 & 0.09 & 0.05 & 0.03 & 0.02 & 0.14 & 0.12 & 0.09 \\
\midrule
\multirow{6}{*}{$\lvert \forest \rvert = 20$}  & $\varepsilon = 0.1$ & \textbf{0.28} & \textbf{0.34} & \textbf{0.30} & \textbf{0.29} & \textbf{0.34} & \textbf{0.33} & \textbf{0.29} & \textbf{0.35} & \textbf{0.33} \\
& $\varepsilon = 1$ & 0.19 & 0.22 & 0.22 & 0.16 & 0.19 & 0.19 & 0.23 & 0.25 & 0.26 \\
& $\varepsilon = 5$ & 0.12 & 0.16 & 0.19 & 0.09 & 0.08 & 0.14 & 0.17 & 0.19 & 0.21 \\
& $\varepsilon = 10$ & 0.11 & 0.11 & 0.14 & 0.06 & 0.06 & 0.10 & 0.14 & 0.14 & 0.18 \\
& $\varepsilon = 20$ & 0.11 & 0.09 & 0.10 & 0.05 & 0.05 & 0.05 & 0.12 & 0.12 & 0.13 \\
& $\varepsilon = 30$ & 0.10 & 0.09 & 0.08 & 0.04 & 0.03 & 0.04 & 0.12 & 0.11 & 0.12 \\
\midrule
\multirow{6}{*}{$\lvert \forest \rvert = 30$}  & $\varepsilon = 0.1$ & \textbf{0.34} & \textbf{0.35} & - & \textbf{0.33} & \textbf{0.33} & - & \textbf{0.36} & \textbf{0.38} & - \\
& $\varepsilon = 1$ & 0.21 & 0.22 & 0.25 & 0.17 & 0.20 & \textbf{0.21} & 0.23 & 0.26 & 0.25 \\
& $\varepsilon = 5$ & 0.15 & 0.16 & 0.22 & 0.09 & 0.10 & 0.16 & 0.17 & 0.20 & 0.24 \\
& $\varepsilon = 10$ & 0.13 & 0.14 & 0.18 & 0.07 & 0.08 & 0.13 & 0.14 & 0.14 & 0.19 \\
& $\varepsilon = 20$ & 0.10 & 0.10 & 0.13 & 0.06 & 0.06 & 0.09 & 0.13 & 0.13 & 0.17 \\
& $\varepsilon = 30$ & 0.10 & 0.09 & 0.11 & 0.04 & 0.04 & 0.06 & 0.11 & 0.11 & 0.13 \\
\midrule
\multicolumn{2}{c}{Random baseline} & \multicolumn{3}{c}{0.25} & \multicolumn{3}{c}{0.21} & \multicolumn{3}{c}{0.26} \\\bottomrule\end{tabular}
\end{table*}

In Appendix~\ref{appendix:rfs_perfs_comparisons}, we compare the predictive performance of the considered DP RFs (for a large privacy budget $\varepsilon=30$) with that of standard, non-DP RFs built using the popular CART implementation from the \texttt{scikit-learn}~\cite{scikit-learn} Python library. 
This comparison aims to separate the effect of the DP RF algorithmic design from the impact of DP itself. 
These design choices include omitting bootstrap sampling and selecting split values and attributes at random rather than using an information gain criterion. 
Our results show that, with identical hyperparameters, DP RFs with $\varepsilon=30$ achieve test accuracies comparable to those of traditional RFs, thereby confirming the soundness of the algorithmic design.

\vspace{0.25\baselineskip}

\textbf{Reconstruction attack.} We use the \texttt{OR-Tools} CP-SAT solver~\cite{cpsatlp} (v9) to solve the CP formulation introduced in Section~\ref{sec:attack}. 
Each model resolution is limited to a maximum of $2$ hours of CPU time using $16$ threads with up to $7$ GB of RAM per thread. 
\updated{Note that in practice, feasible solutions (\emph{i.e.}, reconstructions that are compatible with the RF's structure and the observed noisy counts) can be obtained using only a small fraction of the $2$-hours timeout (in average, less than two minutes), as reported in Table~\ref{tab:solution-times} in the Appendix~\ref{appendix:all_running_times}.}
All experiments are run on a computing cluster over a set of homogeneous nodes with AMD EPYC 7532 (Zen 2) @ 2.40 GHz CPU.

\vspace{0.25\baselineskip}

\textbf{Baselines.}  We consider two baselines in our experiments. 
As in~\cite{ferry2024trained}, a random reconstruction baseline is used to quantify the amount of information that can be inferred without knowledge of the RF (\emph{i.e.}, only based on the number of examples $\nexamples$, the different attributes $\nattributes$, their domains and one-hot encoding). 
This baseline randomly guesses the value of each attribute of each example while remaining consistent with one-hot encoding when applicable (we average its performance over $100$ runs). 
Note that since the reconstruction error is computed over the encoded attributes (provided as inputs to the RF), consistency with respect to one-hot-encoding naturally leads to reconstruction errors lower than $0.5$, even for the random baseline.
To quantify the effect of DP on the reconstruction error, we also run the state-of-the-art DRAFT attack against the trained RFs without the DP protection (\emph{i.e.}, with the true leaves' counts). 

\vspace{0.25\baselineskip}

\textbf{Reconstruction error measure.} In line with the literature~\cite{ferry2024trained}, while a reconstructed dataset is inferred, first it is aligned with the original training set using a minimum cost matching with the Manhattan distance, in order to identify which reconstructed examples correspond to which original ones.
To do so, the Manhattan distance between each reconstructed and original example is computed. 
The resulting distance matrix then instantiates a minimum weight matching in bipartite graphs, also known as linear sum assignment problem, which we solve using the \texttt{Scipy}~\cite{2020SciPy-NMeth} Python library. 
Then, the proportion of binary attributes that differ is averaged over the entire dataset to quantify the reconstruction error. 

Each experiment is repeated using five different random seeds and the results are averaged. 

\subsection{Results}
\label{subsec:results}

We now highlight our key empirical findings and illustrate each of them with a subset of representative results.

\vspace{0.25\baselineskip}
\textbf{Result 1. Reconstruction attacks succeed against DP RFs, even with meaningful privacy guarantees.} 
Table~\ref{tab:error-values} reports the reconstruction error for all our experiments. 
\begin{figure}
    \centering
    \includegraphics[width=\linewidth]{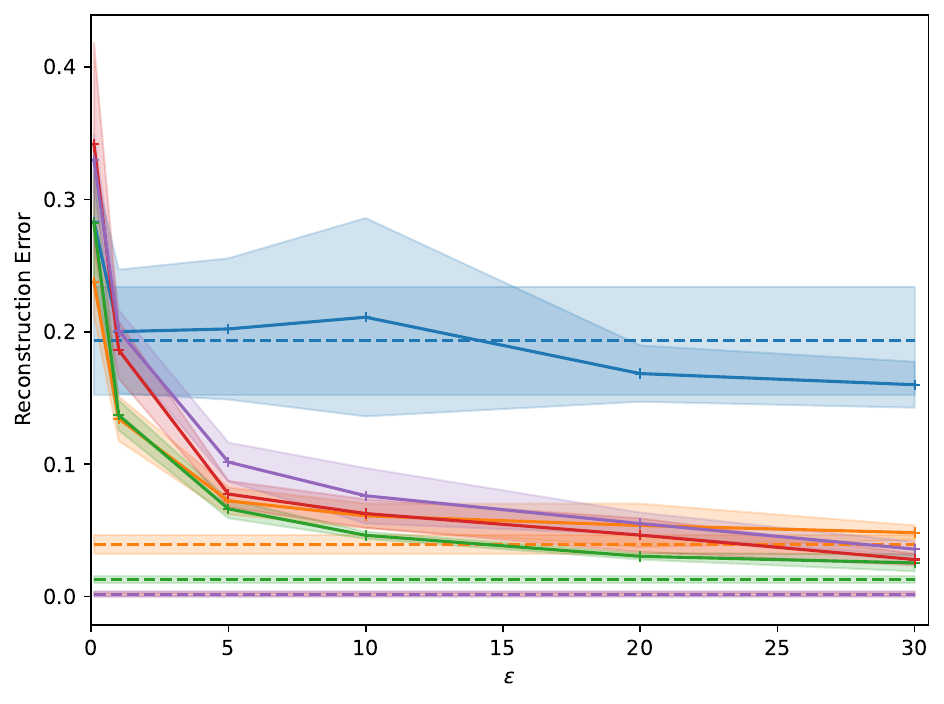}
    \includegraphics[width=\linewidth]{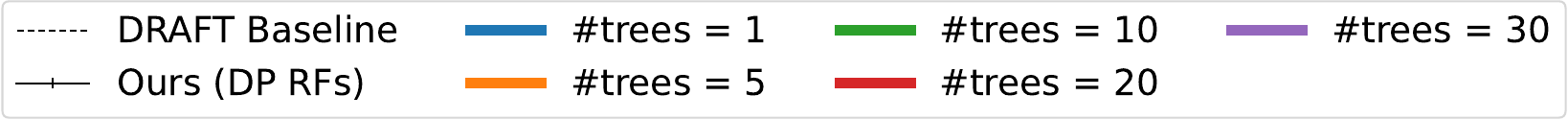}
    \caption{Average reconstruction error as a function of the privacy budget $\varepsilon$ used to fit the target DP RF, for different numbers of depth-5 trees $\vert \forest \rvert$ on the COMPAS dataset. 
    For comparison, we also report the reconstruction error of DRAFT applied to the same RFs without DP protection (the x-axis does not apply for this baseline), using a dashed line.
    }
    \label{fig:DRAFT_baseline}
\end{figure}
For commonly protective privacy budgets (\emph{e.g.}, $\varepsilon = 1$ or $5$), our attack achieves significantly lower reconstruction error compared to the random baseline. 
This indicates that the adversary was able to gain insights from the RF to better infer the training data.
Nonetheless, the reconstruction error consistently increases when reducing the privacy budget $\varepsilon$, which suggests that DP partially mitigates the reconstruction success. 
To assess this, Figure~\ref{fig:DRAFT_baseline} compares the effectiveness of our reconstruction to that of DRAFT~\cite{ferry2024trained} applied on the same RF (without DP\footnote{Note that the original DRAFT attack formulation cannot be directly applied to DP RFs, since any inconsistency between leaf counts (likely to arise from the added random noise) would render the underlying optimization problem infeasible.}). 
For very tight privacy budgets (\emph{e.g.}, small~$\varepsilon$), DP significantly increases the reconstruction error compared to the DRAFT baseline. 
However, as $\varepsilon$ increases, the protective effect of DP diminishes, and the difference in reconstruction error between our attack and DRAFT becomes much smaller, even for $\varepsilon$ values commonly considered protective.
Overall, this comparison confirms that the proposed attack asymptotically approaches the original DRAFT method for large privacy budgets $\varepsilon$.

We nevertheless observe from Table~\ref{tab:error-values} that no feasible reconstructions were found within the given time limit for the largest and deepest ($\vert \forest \rvert = 30$ and $\depth = 7$) DP RFs with the tightest privacy budget $\varepsilon = 0.1$. 
We emphasize that this does not indicate that the reconstruction problem is infeasible, as by design, the true DP RF training set is always part of the CP formulation search space with very high probability. 
Rather, it suggests that exploring the attack search space and identifying feasible reconstructions becomes computationally more challenging and would require an increase of the solver's running time.
Two main factors are at stake. 
First, increasing either the number of trees or their depth directly increases the number of variables. Second, tighter privacy budgets lead to larger domains for the noise variables. 
Together, these effects substantially enlarge the search space.

\vspace{0.25\baselineskip}
\textbf{Result 2. DP leads to a complex trade-off between the RFs' size and the reconstruction attack success.} 
As can be seen in Figure~\ref{fig:DRAFT_baseline}, larger DP RFs do not necessarily provide more information to the attacker, unlike non-DP RFs. 
A threshold effect is observed: the reconstruction error decreases as the number of trees increases, up to a point (typically around 5-10 trees), before increasing again. 
This behaviour is consistent across all privacy budgets and is likely due to the privacy budget being distributed across more trees, resulting in noisier counts for individual leaves. 
Similarly, while deeper trees in the non-DP setting provide more information to the attacker, this is not always the case for DP RFs (see also, Table~\ref{tab:error-values}) since smaller leaves' counts are proportionally more affected by the noise.
These observations highlight the importance of carefully tuning the hyperparameters of DP RFs, as different trade-offs between predictive accuracy and empirical vulnerability to our reconstruction attack can arise even under a fixed privacy budget $\varepsilon$. 
For example, on the COMPAS dataset and for $\varepsilon=1$, a DP RF composed of $\lvert \forest \rvert = 10$ depth-$3$ trees and another composed of $\lvert \forest \rvert = 5$ depth-$7$ trees yield approximately the same reconstruction error under our attack (0.12, see Table~\ref{tab:error-values}). However, their training accuracies differ substantially ($0.57$ versus $0.61$), as reported in Figures~\ref{fig:results_tradeoffs_acc_reconstr_COMPAS_depth_3} and~\ref{fig:results_tradeoffs_acc_reconstr_COMPAS_depth_7} in Appendix~\ref{appendix:tradeoffs_detailed}.

\begin{figure}
    \centering
    \includegraphics[width=\linewidth]{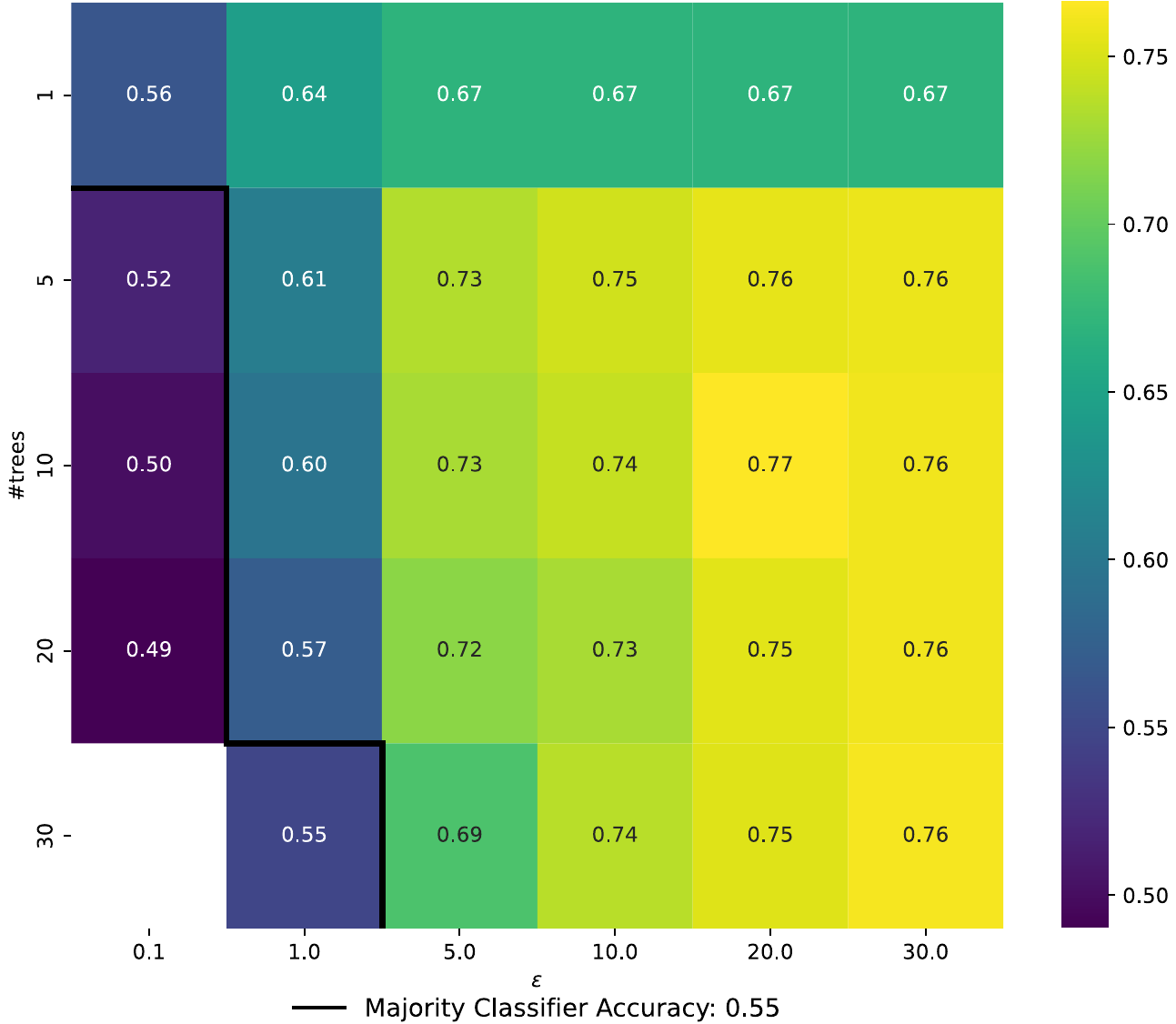}
     \caption{Average training accuracy of $\varepsilon$-DP RFs with depth-7 trees as a function of the privacy budget $\varepsilon$, for different forest sizes $\lvert \forest \rvert$ on the COMPAS dataset}
    \label{fig:heatmap}
\end{figure}

\begin{figure}
    \centering
    \includegraphics[width=\linewidth]{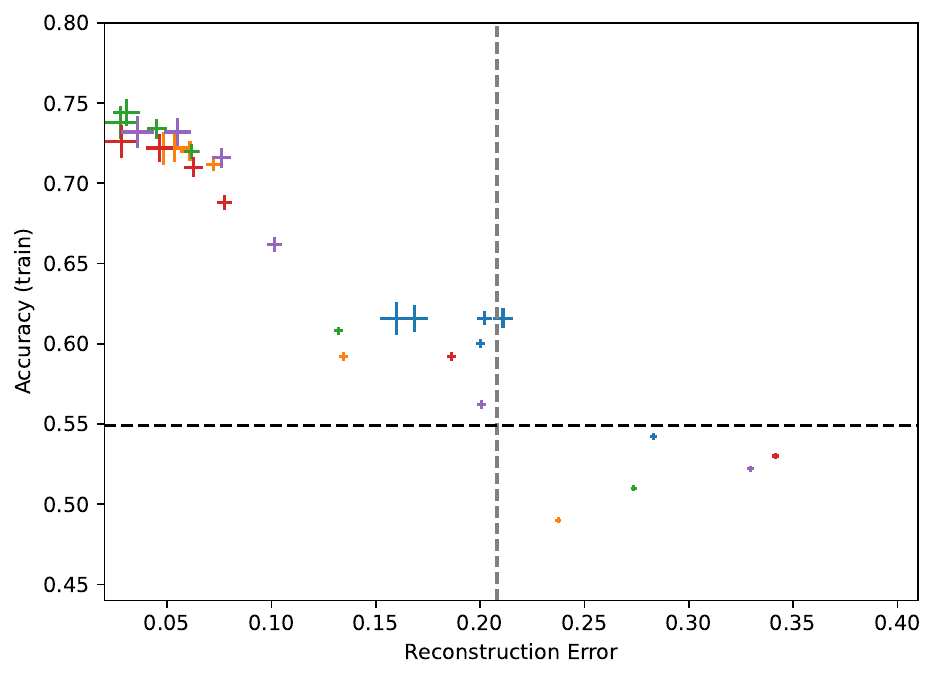}
    
    \vspace{0.6em}
    
    \includegraphics[width=\linewidth]{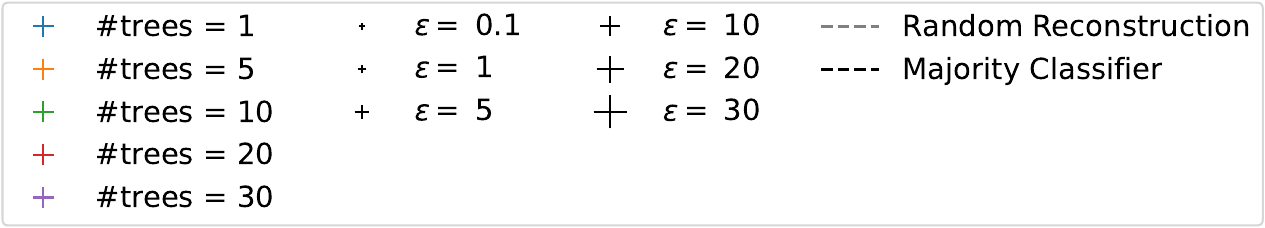}
    
    \caption{Average training accuracy of $\varepsilon$-DP RFs with depth-5 trees as a function of the reconstruction error, for different privacy budgets $\varepsilon$ and forest sizes $\lvert \forest \rvert$ on the COMPAS dataset.} \label{fig:acc_vs_error_compas}
\end{figure}

\vspace{\baselineskip}
\textbf{Result 3. Protection against reconstruction attacks comes at the expense of predictive accuracy.} 
As shown in Table~\ref{tab:error-values}, the considered DP mechanism effectively mitigates the success of reconstruction attacks at very low $\varepsilon$ values (\emph{e.g.}, $\varepsilon = 0.1$). 
However, at these privacy levels, the resulting RFs have poor predictive accuracy, often performing no better than a majority prediction classifier (constantly predicting the majority class), as illustrated in Figure~\ref{fig:heatmap}. %
Notably, only DP RFs with trivial predictive performance provide no advantage to the reconstruction adversary (compared to the random reconstruction baseline). 
This is also visible in Figure~\ref{fig:acc_vs_error_compas}, in which most DP RFs either perform better than a majority classifier but provide a significant advantage to the reconstruction attacker (upper left corner), or provide no advantage to the attacker but perform worse than a majority classifier (lower right corner). 
This is observed across all considered datasets and maximum depths, as evidenced in the detailed results reported in Appendix~\ref{appendix:tradeoffs_detailed}. 
Notably, the only DP FRs not falling within this dichotomy consist of a single tree, which prevents information from being combined across learners. 
While we included these special cases for completeness, the probabilistic reconstruction attack of~\cite{ferry2024probabilistic} would be more appropriate in such settings, although it is limited to single decision trees and does not natively handle DP.

\begin{table*}[htb]
\centering
\caption{\updated{Privacy leak quantification: the reported values quantify \emph{the likelihood of observing a reconstruction error lower or equal to the one measured on the actual RF training set, using a dataset randomly drawn from the same data distribution}.
More precisely, we first estimate the distribution of the errors measured between the dataset reconstructed from the trained RF and datasets randomly drawn from the data distribution. 
Then, we compute and report in this table the cumulative density function (CDF) at the point corresponding to the reconstruction error measured using the actual RF training dataset. 
We display it for different numbers of trees $\lvert \forest \rvert$ within the target forest, tree depths $\depth$ and privacy budgets $\varepsilon$ for the three datasets. 
The situations in which the solver failed to find a feasible reconstruction within the $2$-hour time limit are marked with~``--''. 
Values below $5$\% are \textbf{bold}, indicating that there is a very small-to-negligible probability that the actual RF training set lies as close to the reconstructed dataset by chance (given the data distribution).}}\label{tab:p-values}
\begin{tabular}{llccccccccc}
\toprule
                                &                     & \multicolumn{3}{c}{Adult} & \multicolumn{3}{c}{COMPAS} & \multicolumn{3}{c}{Default Credit} \\
\cmidrule(lr){3-5} \cmidrule(lr){6-8} \cmidrule(lr){9-11}
                                &                     & $d=3$   & $d=5$  & $d=7$  & $d=3$   & $d=5$   & $d=7$  & $d=3$      & $d=5$      & $d=7$     \\
\midrule
\multirow{6}{*}{$\lvert \forest \rvert = 1$}  & $\varepsilon = 0.1$ & 0.208 & 0.544 & 0.344 & 0.614 & 0.480 & 0.464 & 0.852 & 0.884 & 0.914 \\
& $\varepsilon = 1$ & 0.106 & 0.225 & 0.144 & 0.528 & 0.357 & 0.212 & 0.803 & 0.771 & 0.535 \\
& $\varepsilon = 5$ & 0.074 & 0.277 & 0.067 & 0.464 & 0.271 & 0.102 & 0.764 & 0.613 & 0.146 \\
& $\varepsilon = 10$ & 0.072 & 0.229 & 0.147 & 0.461 & 0.274 & 0.158 & 0.761 & 0.622 & 0.211 \\
& $\varepsilon = 20$ & 0.072 & 0.205 & 0.201 & 0.454 & 0.328 & 0.262 & 0.761 & 0.762 & 0.081 \\
& $\varepsilon = 30$ & 0.071 & 0.205 & 0.172 & 0.461 & 0.254 & 0.286 & 0.762 & 0.790 & 0.074 \\
\midrule
\multirow{6}{*}{$\lvert \forest \rvert = 5$}  & $\varepsilon = 0.1$ & 0.164 & 0.308 & 0.264 & 0.358 & 0.342 & 0.177 & 0.823 & 0.748 & 0.778 \\
& $\varepsilon = 1$ & 0.087 & 0.266 & 0.079 & 0.331 & \textbf{{0.042}} & 0.208 & 0.644 & 0.534 & 0.651 \\
& $\varepsilon = 5$ & \textbf{{0.035}} & 0.061 & \textbf{{0.002}} & 0.114 & \textbf{3.01E-05} & \textbf{3.37E-05} & 0.606 & \textbf{{0.034}} & \textbf{{0.027}} \\
& $\varepsilon = 10$ & 0.053 & \textbf{{0.007}} & \textbf{1.14E-04} & \textbf{{0.023}} & \textbf{9.09E-07} & \textbf{5.01E-08} & 0.618 & \textbf{{0.033}} & \textbf{1.77E-04} \\
& $\varepsilon = 20$ & \textbf{{0.008}} & \textbf{{0.007}} & \textbf{8.75E-06} & \textbf{{0.035}} & \textbf{5.66E-08} & \textbf{1.16E-11} & 0.715 & \textbf{{0.002}} & \textbf{1.98E-05} \\
& $\varepsilon = 30$ & \textbf{{0.008}} & \textbf{{0.011}} & \textbf{8.03E-07} & \textbf{{0.022}} & \textbf{2.52E-07} & \textbf{1.60E-10} & 0.685 & \textbf{{0.002}} & \textbf{4.24E-04} \\
\midrule
\multirow{6}{*}{$\lvert \forest \rvert = 10$}  & $\varepsilon = 0.1$ & 0.304 & 0.332 & 0.321 & 0.326 & 0.512 & 0.101 & 0.652 & 0.438 & 0.446 \\
& $\varepsilon = 1$ & 0.066 & 0.144 & \textbf{{0.021}} & \textbf{{0.030}} & 0.190 & 0.127 & 0.442 & 0.543 & 0.463 \\
& $\varepsilon = 5$ & 0.077 & \textbf{{0.001}} & \textbf{{0.003}} & \textbf{{0.022}} & \textbf{{0.001}} & \textbf{2.74E-04} & 0.297 & 0.068 & 0.087 \\
& $\varepsilon = 10$ & \textbf{{0.031}} & \textbf{{0.001}} & \textbf{1.31E-04} & \textbf{{0.002}} & \textbf{5.08E-06} & \textbf{4.65E-05} & 0.149 & \textbf{{0.011}} & \textbf{7.00E-06} \\
& $\varepsilon = 20$ & \textbf{{0.039}} & \textbf{9.42E-04} & \textbf{1.80E-07} & \textbf{1.19E-04} & \textbf{5.59E-07} & \textbf{1.70E-11} & 0.173 & \textbf{1.46E-04} & \textbf{7.26E-04} \\
& $\varepsilon = 30$ & \textbf{{0.007}} & \textbf{3.82E-07} & \textbf{1.92E-06} & \textbf{2.65E-05} & \textbf{4.89E-09} & \textbf{8.05E-13} & 0.200 & \textbf{8.57E-05} & \textbf{1.23E-08} \\
\midrule
\multirow{6}{*}{$\lvert \forest \rvert = 20$}  & $\varepsilon = 0.1$ & 0.143 & 0.395 & 0.219 & 0.443 & 0.436 & 0.239 & 0.548 & 0.546 & 0.559 \\
& $\varepsilon = 1$ & 0.176 & 0.132 & 0.093 & 0.233 & 0.372 & 0.342 & 0.379 & 0.540 & 0.571 \\
& $\varepsilon = 5$ & \textbf{{0.016}} & 0.056 & \textbf{{0.008}} & 0.069 & \textbf{{0.022}} & 0.101 & 0.170 & 0.245 & 0.195 \\
& $\varepsilon = 10$ & \textbf{{0.009}} & \textbf{{0.006}} & \textbf{8.69E-04} & \textbf{{0.008}} & \textbf{{0.008}} & \textbf{{0.032}} & 0.092 & \textbf{{0.005}} & 0.131 \\
& $\varepsilon = 20$ & \textbf{{0.021}} & \textbf{2.84E-04} & \textbf{6.46E-06} & \textbf{1.66E-04} & \textbf{7.96E-05} & \textbf{6.69E-04} & \textbf{{0.008}} & \textbf{7.05E-04} & \textbf{{0.004}} \\
& $\varepsilon = 30$ & \textbf{{0.002}} & \textbf{1.20E-05} & \textbf{6.32E-08} & \textbf{2.68E-05} & \textbf{5.86E-10} & \textbf{1.18E-07} & \textbf{{0.004}} & \textbf{8.61E-05} & \textbf{1.31E-05} \\
\midrule
\multirow{6}{*}{$\lvert \forest \rvert = 30$}  & $\varepsilon = 0.1$ & 0.383 & 0.364 & - & 0.274 & 0.302 & - & 0.630 & 0.501 & - \\
& $\varepsilon = 1$ & 0.179 & 0.069 & 0.064 & 0.309 & 0.389 & 0.271 & 0.461 & 0.452 & 0.453 \\
& $\varepsilon = 5$ & 0.125 & \textbf{{0.007}} & \textbf{{0.028}} & 0.054 & \textbf{{0.036}} & 0.119 & 0.113 & 0.141 & 0.291 \\
& $\varepsilon = 10$ & 0.064 & \textbf{{0.004}} & \textbf{{0.008}} & 0.050 & \textbf{{0.001}} & 0.107 & \textbf{{0.017}} & \textbf{{0.001}} & 0.071 \\
& $\varepsilon = 20$ & \textbf{{0.006}} & \textbf{5.75E-04} & \textbf{{0.001}} & \textbf{{0.001}} & \textbf{1.07E-04} & 0.084 & \textbf{{0.006}} & \textbf{7.83E-04} & \textbf{{0.032}} \\
& $\varepsilon = 30$ & \textbf{2.50E-04} & \textbf{2.86E-04} & \textbf{{0.001}} & \textbf{3.84E-06} & \textbf{1.09E-08} & \textbf{1.63E-04} & \textbf{4.84E-04} & \textbf{5.11E-05} & \textbf{5.08E-04} \\
\bottomrule\end{tabular}
\end{table*}

\vspace{0.25\baselineskip}
\textbf{Result 4. For typical values of $\varepsilon$, one can infer information specific to the DP RF training set beyond general distributional patterns.} 
The results in Table~\ref{tab:error-values} demonstrate that the proposed attack extracts information useful for accurately reconstructing the DP RF training dataset $\dataset{}$, which may reflect general distributional patterns and/or dataset-specific details.
While the former is still valuable (in particular, since our attack assumes no knowledge of the data distribution), privacy leakages are typically defined in terms of recovering information specific to the actual training set~\cite{DBLP:conf/innovations/CohenKMMNST25}. Appendix~\ref{appendix:privacy_leak} presents a complementary analysis in which we assess \emph{whether the reconstructed dataset is closer to $\dataset{}$ than to other datasets from the same data distribution, in a statistically significant manner}. 
More precisely, we randomly sample $100$ datasets containing $\nexamples$ examples from the same distribution as $\dataset$, and compute the errors between each of them and the reconstructed dataset. 
We then fit a normal distribution from this list of errors, estimating how likely it is to measure a given reconstruction error for a dataset randomly drawn from the data distribution. 
We finally compute the cumulative density function (CDF) of this fitted distribution for the actual reconstruction error, which quantifies \emph{the likelihood of observing a reconstruction error lower or equal to the one measured on the actual RF training set, using a dataset randomly drawn from the same data distribution}. 
Hence, it is somewhat analogous to a $p$-value, with very small values corresponding to observations that are unlikely to be observed purely by chance.
We illustrate this process on Figure~\ref{fig:results_privacy_leak} in the Appendix~\ref{appendix:privacy_leak}.
The detailed results provided in Table~\ref{tab:p-values} suggest that for very tight DP budgets (\emph{i.e.}, $\varepsilon = 0.1$), only distributional information is inferred. 
However, larger DP budgets (\emph{i.e.}, $\varepsilon \geq 5$) usually allow the retrieval of dataset-specific details. For instance, the likelihood of observing purely by chance (given the data distribution) a reconstruction error as low as the one obtained by our attack (namely, $0.06$) against DP RFs composed of $\lvert \forest \rvert = 10$ depth-$7$ trees trained on the COMPAS dataset with privacy budget $\varepsilon=5$ is roughly $2.74 \times 10^{-4}$.
Several configurations with small DP budgets (\emph{i.e.}, $\varepsilon=1$) can also lead to the inference of information specific to $\dataset{}$. 
Importantly, for a given DP budget, different DP RFs configurations can lead to significantly different privacy leaks likelihood, which highlights the importance of carefully tuning all hyperparameters beyond the DP budget alone. 
This observation is consistent with the literature as previous work~\cite{hayes2023bounding} on reconstruction attacks against differentially private deep learning models similarly found that for a fixed DP budget $\varepsilon$, varying the DP mechanism’s hyperparameters can lead to substantially different reconstruction success rates. 
The authors of this study therefore concluded that the DP budget alone ``might not be a good proxy for controlling protection against reconstruction attacks'', which aligns with our findings.

\begin{table*}[h!]
\centering
\caption{Individual reconstruction error analysis for our experiments on the three considered datasets, using a representative DP RF configuration of $\lvert \forest \rvert = 10$ depth-$5$ trees. For different privacy budgets $\varepsilon$, we report the average reconstruction error, the proportion of individual examples perfectly reconstructed, and the worst individual reconstruction error across the entire training set. We further provide the average reconstruction error and the proportion of perfectly reconstructed examples separately for the subsets of training examples identified as inliers and as outliers.}\label{tab:individual_reconstruction_results}
\begin{subtable}{\textwidth}
\centering
\caption{UCI Adult Income dataset}\label{tab:individual_reconstruction_results_adult}
\begin{tabular}{@{}clllllll@{}}
\toprule
\multirow{2}{*}{\textbf{$\varepsilon$}} & \multicolumn{3}{c}{\textbf{Global}}                                                                    & \multicolumn{2}{c}{\textbf{Inliers}}                                & \multicolumn{2}{c}{\textbf{Outliers}}                         \\  \cmidrule(lr){2-4} \cmidrule(lr){5-6} \cmidrule(lr){7-8} 
                                        & \multicolumn{1}{c}{Avg error} & \multicolumn{1}{c}{\%Perfect} & \multicolumn{1}{c}{Worst reconstr.}    & \multicolumn{1}{c}{Avg error} & \multicolumn{1}{c}{\%Perfect}       & \multicolumn{1}{c}{Avg error} & \multicolumn{1}{c}{\%Perfect} \\ \midrule
\multicolumn{1}{c|}{30}                 & $0.088 \pm 0.002$             & $14.6 \pm 2.4$                & \multicolumn{1}{l|}{$0.274 \pm 0.021$} & $0.071 \pm 0.004$             & \multicolumn{1}{l|}{$17.7 \pm 3.1$} & $0.112 \pm 0.004$             & $10.0 \pm 3.5$                \\
\multicolumn{1}{c|}{20}                 & $0.090 \pm 0.003$             & $16.2 \pm 2.3$                & \multicolumn{1}{l|}{$0.337 \pm 0.042$} & $0.071 \pm 0.006$             & \multicolumn{1}{l|}{$22.0 \pm 2.9$} & $0.119 \pm 0.005$             & $7.5 \pm 4.7$                 \\
\multicolumn{1}{c|}{10}                 & $0.100 \pm 0.004$             & $12.0 \pm 3.6$                & \multicolumn{1}{l|}{$0.347 \pm 0.042$} & $0.082 \pm 0.007$             & \multicolumn{1}{l|}{$16.0 \pm 4.4$} & $0.127 \pm 0.007$             & $6.0 \pm 3.4$                 \\
\multicolumn{1}{c|}{5}                  & $0.122 \pm 0.005$             & $6.6 \pm 2.0$                 & \multicolumn{1}{l|}{$0.368 \pm 0.000$} & $0.099 \pm 0.002$             & \multicolumn{1}{l|}{$9.7 \pm 2.4$}  & $0.156 \pm 0.012$             & $2.0 \pm 1.9$                 \\
\multicolumn{1}{c|}{1}                  & $0.199 \pm 0.013$             & $2.2 \pm 1.5$                 & \multicolumn{1}{l|}{$0.442 \pm 0.026$} & $0.184 \pm 0.017$             & \multicolumn{1}{l|}{$3.3 \pm 2.8$}  & $0.222 \pm 0.013$             & $0.5 \pm 1.0$                 \\
\multicolumn{1}{c|}{0.1}                & $0.291 \pm 0.029$             & $0.0 \pm 0.0$                 & \multicolumn{1}{l|}{$0.474 \pm 0.047$} & $0.298 \pm 0.032$             & \multicolumn{1}{l|}{$0.0 \pm 0.0$}  & $0.280 \pm 0.026$             & $0.0 \pm 0.0$                 \\ \bottomrule
\end{tabular}
\end{subtable}

\vspace{\baselineskip} 

\begin{subtable}{\textwidth}
\centering
\caption{COMPAS dataset}\label{tab:individual_reconstruction_results_compas}
\begin{tabular}{@{}cccccccc@{}}
\toprule
\multirow{2}{*}{\textbf{$\varepsilon$}} & \multicolumn{3}{c}{\textbf{Global}}                                         & \multicolumn{2}{c}{\textbf{Inliers}}                     & \multicolumn{2}{c}{\textbf{Outliers}} \\
 \cmidrule(lr){2-4} \cmidrule(lr){5-6} \cmidrule(lr){7-8} 
                                        & Avg error         & \%Perfect      & Worst reconstr.                        & Avg error         & \%Perfect                            & Avg error           & \%Perfect       \\ \midrule
\multicolumn{1}{c|}{30}                 & $0.028 \pm 0.006$ & $72.8 \pm 3.1$ & \multicolumn{1}{c|}{$0.243 \pm 0.035$} & $0.016 \pm 0.005$ & \multicolumn{1}{c|}{$83.3 \pm 6.2$}  & $0.042 \pm 0.012$   & $60.0 \pm 9.2$  \\
\multicolumn{1}{c|}{20}                 & $0.031 \pm 0.004$ & $71.6 \pm 3.3$ & \multicolumn{1}{c|}{$0.257 \pm 0.057$} & $0.016 \pm 0.004$ & \multicolumn{1}{c|}{$82.9 \pm 2.5$}  & $0.049 \pm 0.007$   & $57.8 \pm 6.1$  \\
\multicolumn{1}{c|}{10}                 & $0.045 \pm 0.003$ & $62.4 \pm 1.9$ & \multicolumn{1}{c|}{$0.300 \pm 0.070$} & $0.025 \pm 0.008$ & \multicolumn{1}{c|}{$77.1 \pm 3.9$}  & $0.070 \pm 0.013$   & $44.4 \pm 4.2$  \\
\multicolumn{1}{c|}{5}                  & $0.062 \pm 0.003$ & $51.6 \pm 2.6$ & \multicolumn{1}{c|}{$0.286 \pm 0.000$} & $0.031 \pm 0.005$ & \multicolumn{1}{c|}{$70.9 \pm 2.6$}  & $0.100 \pm 0.006$   & $28.0 \pm 4.6$  \\
\multicolumn{1}{c|}{1}                  & $0.132 \pm 0.014$ & $22.0 \pm 6.8$ & \multicolumn{1}{c|}{$0.414 \pm 0.095$} & $0.112 \pm 0.024$ & \multicolumn{1}{c|}{$30.5 \pm 11.9$} & $0.157 \pm 0.015$   & $11.6 \pm 6.7$  \\
\multicolumn{1}{c|}{0.1}                & $0.274 \pm 0.040$ & $1.0 \pm 0.0$  & \multicolumn{1}{c|}{$0.514 \pm 0.070$} & $0.289 \pm 0.043$ & \multicolumn{1}{c|}{$0.4 \pm 0.7$}   & $0.255 \pm 0.036$   & $1.8 \pm 0.9$   \\ \bottomrule
\end{tabular}
\end{subtable}

\vspace{\baselineskip} 

\begin{subtable}{\textwidth}
\centering
\caption{Default of Credit Card Clients dataset}\label{tab:individual_reconstruction_result_default_credit}
\begin{tabular}{@{}clllllll@{}}
\toprule
\multirow{2}{*}{\textbf{$\varepsilon$}} & \multicolumn{3}{c}{\textbf{Global}}                                                                    & \multicolumn{2}{c}{\textbf{Inliers}}                                & \multicolumn{2}{c}{\textbf{Outliers}}                         \\ \cmidrule(lr){2-4} \cmidrule(lr){5-6} \cmidrule(lr){7-8} 
                                        & \multicolumn{1}{c}{Avg error} & \multicolumn{1}{c}{\%Perfect} & \multicolumn{1}{c}{Worst reconstr.}    & \multicolumn{1}{c}{Avg error} & \multicolumn{1}{c}{\%Perfect}       & \multicolumn{1}{c}{Avg error} & \multicolumn{1}{c}{\%Perfect} \\ \midrule
\multicolumn{1}{c|}{30}                 & $0.117 \pm 0.002$             & $10.0 \pm 2.6$                & \multicolumn{1}{l|}{$0.343 \pm 0.036$} & $0.078 \pm 0.005$             & \multicolumn{1}{l|}{$18.8 \pm 6.0$} & $0.154 \pm 0.008$             & $1.6 \pm 1.5$                 \\
\multicolumn{1}{c|}{20}                 & $0.119 \pm 0.005$             & $10.8 \pm 3.1$                & \multicolumn{1}{l|}{$0.362 \pm 0.049$} & $0.084 \pm 0.007$             & \multicolumn{1}{l|}{$19.6 \pm 6.3$} & $0.152 \pm 0.004$             & $2.4 \pm 1.5$                 \\
\multicolumn{1}{c|}{10}                 & $0.141 \pm 0.015$             & $7.8 \pm 3.8$                 & \multicolumn{1}{l|}{$0.410 \pm 0.049$} & $0.105 \pm 0.015$             & \multicolumn{1}{l|}{$13.1 \pm 6.3$} & $0.175 \pm 0.015$             & $2.7 \pm 2.0$                 \\
\multicolumn{1}{c|}{5}                  & $0.160 \pm 0.020$             & $4.6 \pm 2.0$                 & \multicolumn{1}{l|}{$0.419 \pm 0.056$} & $0.130 \pm 0.026$             & \multicolumn{1}{l|}{$8.6 \pm 4.4$}  & $0.189 \pm 0.015$             & $0.8 \pm 1.0$                 \\
\multicolumn{1}{c|}{1}                  & $0.239 \pm 0.022$             & $0.2 \pm 0.4$                 & \multicolumn{1}{l|}{$0.457 \pm 0.038$} & $0.241 \pm 0.026$             & \multicolumn{1}{l|}{$0.0 \pm 0.0$}  & $0.238 \pm 0.020$             & $0.4 \pm 0.8$                 \\
\multicolumn{1}{c|}{0.1}                & $0.312 \pm 0.035$             & $0.0 \pm 0.0$                 & \multicolumn{1}{l|}{$0.476 \pm 0.052$} & $0.323 \pm 0.043$             & \multicolumn{1}{l|}{$0.0 \pm 0.0$}  & $0.301 \pm 0.026$             & $0.0 \pm 0.0$                 \\ \bottomrule
\end{tabular}
\end{subtable}

\end{table*}

\vspace{0.25\baselineskip}
\textbf{Result 5. Complete training set examples can be retrieved even for meaningful DP budget.} 
While the dataset-level reconstruction errors presented earlier provide a general measure of attack success, their interpretation is not always straightforward when the error is non-zero, as is the case in all our experiments under DP (unlike in the original non-DP setup~\cite{ferry2024trained}). 
For instance, in many contexts, perfectly reconstructing a subset of training examples represents a more severe privacy leakage than partially reconstructing most of them. 
To capture this aspect, we also measure the proportion of training examples perfectly reconstructed, \emph{i.e.}, those for which every attribute value is exactly recovered. 

Table~\ref{tab:individual_reconstruction_results} presents example results obtained on the three considered datasets, for a representative DP RF configuration of $\lvert \forest \rvert = 10$ trees of depth $5$. 
For varying privacy budgets $\varepsilon$, the leftmost part (``Global'') of each sub-table reports the average reconstruction error (as already reported in Table~\ref{tab:error-values}), the proportion of training examples perfectly reconstructed (out of the $\nexamples=100$ original ones), and the worst individual reconstruction error across the entire training set. 
The results indicate that a substantial number of training examples are perfectly reconstructed in practice, with this number increasing as the privacy budget $\varepsilon$ grows. Even under conservative privacy budgets, the attack remains effective. 
For instance, an average of 22 training examples (over five random folds) are perfectly reconstructed for $\varepsilon = 1$ on the COMPAS dataset, representing nearly a quarter of the training set.
Interestingly, while the worst individual reconstruction error generally decreases when the privacy budget $\varepsilon$ increases, it never beats the random baseline (reported in Table~\ref{tab:error-values}), suggesting that some particular examples still struggle to be retrieved. 

We further investigate whether outliers are empirically more or less vulnerable to reconstruction than in-distribution examples (inliers).
To do so, we first classify each training example as either inlier or outlier using isolation forests as implemented in the \texttt{scikit-learn}~\cite{scikit-learn} Python library, with their default hyperparameters. 
Isolation forests are indeed a widely used method for quantifying the degree to which individual examples are outliers (referred to as the anomaly score) within a dataset. 
They work by recursively partitioning the data using random splits, creating a set of isolation trees.
Outliers are expected to be isolated more quickly, resulting in shorter average path lengths across the trees, which translates into higher anomaly scores.
Based on this analysis, the rightmost parts of each sub-table in Table~\ref{tab:individual_reconstruction_results} report the average reconstruction error and the proportion of perfectly reconstructed training examples for the subsets identified as inliers and outliers.
The results for the three considered datasets indicate that outliers tend to be less vulnerable to reconstruction than in-distribution examples, as reflected by their higher reconstruction errors and lower rates of perfect reconstruction. 

Interestingly, these observations are consistent with prior results: a recent work~\cite{DBLP:journals/corr/abs-2202-07623} theoretically and empirically shows, in the context of language models, that under a variant of DP, sentences with low likelihood (so-called \emph{rare secrets}) have a very low probability of being reconstructed.
This is also an adverse consequence of DP: by theoretically bounding the contribution of each individual example to the algorithm’s output, differential privacy generally causes machine learning models to learn less effectively and to perform worse on outliers and minority groups~\cite{DBLP:conf/nips/BagdasaryanPS19}. Our analysis confirms that one consequence is that outliers are also less at risk of being reconstructed.

\vspace{0.5\baselineskip}
\textbf{Result 6. Knowledge of the training set size $\nexamples$ is unnecessary.} 
Appendix~\ref{appendix:cp_model_n_unknown} presents a modified version of our CP formulation that enables reconstruction without knowledge of the training set size $\nexamples$. 
Although the resulting search space is considerably larger (as the exact number of examples to be reconstructed is not known in advance, a sufficiently large number of reconstruction variables must be included), our experiments demonstrate that accurate reconstruction remains feasible under this assumption. 
More precisely, although the computational cost and reconstruction error increase slightly compared to the traditional setup in which~$\nexamples$ is known, the attack remains highly effective for most practically relevant $\varepsilon$ values. 
Comparing the detailed results provided in Table~\ref{tab:error-values_N_var} (Appendix~\ref{appendix:cp_model_n_unknown}) with those of Table~\ref{tab:error-values}, we indeed observe that the reconstruction error remains comparable to the case in which $\nexamples$ is known. To the best of our knowledge, our attack is the first to entirely reconstruct a machine learning model's training set without a priori knowledge of its size.

\vspace{0.25\baselineskip}
\textbf{Result 7. The reconstruction attack scales well with the number of training examples $\nexamples$.} 
In Appendix~\ref{appendix:additional_experiments_scalability}, we report results of additional experiments varying the size of the reconstructed training set $\nexamples$. 
They demonstrate that increasing the number of training examples $\nexamples$ improves the predictive performance and generalization of the built DP RFs, but also often benefits the reconstruction attack.
On the one hand, reconstructing larger training sets is computationally more challenging, since the CP formulation’s search space grows exponentially with the number of variables. 
On the other hand, in the DP setting, larger values of $\nexamples$ improve the signal-to-noise ratio, making leaf counts more informative as their support grows and overlaps across trees increase. Technically, because each modelled example introduces its own set of constraints, the total number of constraints grows with $\nexamples$: the search space becomes larger, but also more structured.

For instance, the attack's average reconstruction error against DP RFs with $\lvert \forest \rvert = 10$ depth-$5$ trees and $\varepsilon=5$ decreases from $0.152$ to $0.112$ when increasing the number of training examples from the Default of Credit Card Clients dataset from $\nexamples=100$ to $\nexamples=500$. 
This is due to the fact that the relative impact of the noise added to enforce DP decreases as $\nexamples$ increases, leading to more accurate leaf counts benefiting both the RF predictive performance and the proposed reconstruction attack. 
In the meanwhile, the solver’s running time to reach the final reconstruction error is roughly multiplied by $10$ (as shown in Figure~\ref{fig:results_scalability_anytime} in Appendix~\ref{appendix:additional_experiments_scalability}), which remains promising, since the reconstructed dataset size is simultaneously multiplied by $5$.

\vspace{0.25\baselineskip}

\updated{\textbf{Result 8. Knowledge of part of the training set attributes does not help reconstruct the others.} 
In Appendix~\ref{appendix:additional_experiments_partial_reconstruction}, we report results of additional experiments assuming knowledge of part of the attributes, for all training examples. 
This setup corresponds to the situation in which part of the examples' attributes are public information, and prior work on dataset reconstruction from (non-DP) RFs~\cite{ferry2024trained} demonstrated that knowing them facilitates the reconstruction of the others. 
We consistently observe that this is not the case for DP RFs: regardless of how many attributes the adversary knows, no advantage is gained in reconstructing the remaining ones. Therefore, although DP does not fully prevent the extraction of information useful for reconstructing the RF training dataset, it does provide a complementary layer of protection by limiting the adversary’s ability to exploit auxiliary information.
}

{\rebuttalsatmlmultiline
\textbf{Result 9. Our proposed CP formulation can be adapted to the informed adversary setup.} As discussed in Section~\ref{sec:related_work}, several works have considered an informed adversary setup in which the attacker knows $\nexamples - 1$ training examples, has white-box access to the model parameters, and aims to reconstruct the remaining example.
In Appendix~\ref{appendix:informed_adversary}, we show that our attack can be adapted to this setting and that it outperforms existing approaches across all privacy budgets $\varepsilon$. In particular, it is able to retrieve meaningful information about the unknown example for privacy budgets $\varepsilon \geq 5$. 
}

\section{Related Work}
\label{sec:related_work}

In this section, we first review the relevant literature on reconstruction attacks. 
We then summarize prior works that investigated the effectiveness of DP in mitigating such attacks.
Next, we discuss the connection between our proposed attack and previous methods for post-processing differentially private outputs. 
Finally, we provide a comprehensive overview of existing approaches for constructing DP RFs.

\vspace{0.25\baselineskip}

\textbf{Reconstruction Attacks.}
Reconstruction attacks were first introduced against database queries by~\cite{DBLP:conf/pods/DinurN03}. 
In their setup, all database columns are public except for a binary one, which the adversary attempts to recover using (noisy) responses to counting queries. 
By modeling the noise within a linear programming framework, their attack successfully retrieves the private bit for all individuals after a significant number of queries. 
Over the following decade, several works extended this approach~\cite{dwork2017exposed}.

More recently, reconstruction attacks have been adapted to machine learning models. 
Unlike database-focused ones, these attacks often aim to infer specific attributes or recover information about individual samples rather than reconstructing the entire training set~\cite{fredrikson2015model}. 
Many of these attacks also rely on auxiliary information beyond the trained model, such as intermediate gradients computed during training~\cite{DBLP:conf/atis/PhongA0WM17,DBLP:conf/uss/0001B0F020,DBLP:conf/aistats/WangLL23}, stationary points reached during optimization~\cite{DBLP:conf/nips/HaimVYSI22} or fairness-related constraints~\cite{hu2020inference,hamman2022can,ferry2023exploiting}.

In contrast, recent studies have shown that tree-based models are vulnerable to reconstruction attacks due to their combinatorial structure. 
For example, \cite{DBLP:conf/dbsec/GambsGH12} and \cite{ferry2024probabilistic} developed algorithms leveraging white-box access to a single decision tree to reconstruct a probabilistic representation of its dataset, encoding all reconstructions consistent with the tree's structure. 
More recently, \cite{ferry2024trained} proposed DRAFT, a CP-based reconstruction attack against RFs, showing that the structure of a RF can be exploited to reconstruct most of its training data. %

\vspace{0.25\baselineskip}

\textbf{Reconstruction Attacks and DP.} Differential Privacy~\cite{dwork2014algorithmic} has been widely adopted as a protection mechanism thanks to its strong theoretical guarantees. 
While it inherently bounds the risk of membership inference attacks, its effectiveness against other types of attacks remains less clear and has received comparatively limited attention. 

For example, DP has been shown empirically to mitigate reconstruction attacks on deep neural networks for medical imaging~\cite{ziller2024reconciling}. 
The authors report that although strong worst-case guarantees require very tight privacy budgets, larger budgets can still provide effective empirical protection against state-of-the-art reconstruction attacks.

In the context of language models, \cite{DBLP:journals/corr/abs-2202-07623} show that Rényi Differential Privacy (RDP)~\cite{mironov2017renyi}, a variant of DP, provides theoretical guarantees bounding the probability of reconstructing a training example. 
Using language models fine-tuned to comply with RDP, they further demonstrate that, even under large privacy budgets, RDP can meaningfully protect against reconstruction of \emph{rare secrets}, which are defined as training samples with high information content (many unknown bits), which in our setting can be interpreted as outliers.
Consistent with these findings, \cite{balle2022reconstructing} has demonstrated (in the non-private setting) that an informed adversary (possessing knowledge of all but one training example, and optionally auxiliary information about this target) can reconstruct the missing example with high fidelity given white-box access to the model, for both convex and deep learning models.
Their experiments further show that DP can effectively mitigate reconstruction success, even under relatively large privacy budgets, although tighter budgets provide stronger guarantees.

Follow-up work~\cite{pmlr-v162-guo22c} shows that RDP-compliant learning algorithms theoretically provide a lower bound on the expected reconstruction error for any attacker. 
Tighter theoretical bounds are also proposed for another privacy accounting framework, namely Fisher Information Leakage~\cite{DBLP:conf/uai/HannunGM21}. 
Empirical evaluation using linear logistic regression and neural network models demonstrates that the informed reconstruction attack introduced by \cite{balle2022reconstructing} is closely captured by the proposed reconstruction lower bounds.

Complementing these studies, \cite{hayes2023bounding} analyze reconstruction attacks in the context of DP‑SGD~\cite{Abadi_2016} under a strong threat model, in which the attacker has access to all but one training example (as in~\cite{balle2022reconstructing}) and to intermediate gradients. 
They provide theoretical upper bounds on the success of any reconstruction attack, show that these bounds are tighter than those of~\cite{balle2022reconstructing} in their considered setup and propose an empirical attack that approaches this upper bound. 
Their experiments further demonstrate that different hyperparameter choices can lead to substantially different reconstruction success rates, even for the same DP budget. 
This underscores that DP guarantees alone may not fully capture empirical vulnerability in deep learning models.

Overall, the theoretical analyses and bounds proposed in the aforementioned works are not directly applicable in our setup, since they often consider specific variants of DP and, most importantly, assume an adversarial model where the attacker knows all but one training example. 
In contrast, we do not make such an assumption, as we attempt to recover all training examples. Furthermore, DP defences often lead to significantly reduced predictive accuracy~\cite{DBLP:conf/aistats/LiuWCL25}, and the considered attacks typically do not exploit the DP mechanism explicitly. 
Previous works also do not consider tree ensembles, whose structured nature was shown particularly vulnerable to dataset reconstruction attacks~\cite{ferry2024trained}.
Our work advances this field by (1) explicitly accounting for the DP mechanism within the attack formulation, and (2) analyzing the trade-offs between the DP budget, the empirical reconstruction attack success and the DP model's predictive performances in the context of tree ensembles.

\vspace{0.25\baselineskip}

\textbf{Post-processing Differentially Private Outputs.} A line of works proposed methods to improve the utility of differentially private answers to database queries by post-processing them to comply with known structural knowledge. 
For instance, \cite{DBLP:journals/pvldb/HayRMS10} construct sequences of histogram queries such that linear relationships hold between the true answers, and subsequently post-process the noisy DP answers to enforce these consistency constraints. 
Their procedure relies on least-squares minimization under the constraints, ensuring that the corrected answers remain close to the original noisy outputs. 
Building on this idea, \cite{DBLP:journals/vldb/LiMHMR15} introduce the Matrix Mechanism, which, given a set of counting queries, identifies an alternative set of ``strategy'' queries designed so that structural constraints must hold between them. 
The noisy answers to these strategy queries are then post-processed, again via least-squares minimization under consistency constraints, and used to infer the answers to the original queries. 
Although these mechanisms improve the utility of query answers while preserving the DP guarantee, they do not explicitly leverage the distribution of the injected noise. 
In contrast, \cite{DBLP:conf/kdd/LeeWK15} propose a maximum-likelihood post-processing framework, which enforces consistency constraints while explicitly accounting for the Laplace noise distribution, and show that this approach yields significant gains in both accuracy and computational efficiency compared to prior least-squares based methods.

These approaches are closely related to our proposed attack, since we also perform maximum-likelihood optimization under structural constraints. 
However, there are three fundamental differences. 
First, these works have no adversarial objective, but rather aim to provide useful and privacy-preserving statistics. 
Second, in our setup, the adversary observes the noisy leaf counts but cannot generate alternative ``queries", since no further interactions with the training data are allowed. Third, our objective is not to recover the true leaf counts from the noisy ones, but to reconstruct the entire training dataset. Bridging these two levels of representation requires enforcing much richer combinatorial constraints, which makes the problem substantially more difficult. In fact, even in the non-DP setting, reconstructing a training dataset from its exact leaf counts is NP-hard and computationally challenging~\cite{ferry2024trained}. In this sense, our work can be viewed as a state-of-the-art adversarial maximum-likelihood DP post-processing technique aimed at full dataset reconstruction.

\vspace{\baselineskip}

\textbf{Learning DP Random Forests.}
Traditional tree induction algorithms iteratively perform a greedy selection of the attributes and values to split on at each level of the considered tree, based on some information gain criterion. 
Some works on DP RFs proceed similarly~\cite{Singh,DBLP:conf/ausdm/Fletcher015, rana}. 
However, since the training data needs to be accessed at each level of the tree to greedily determine the splits, the DP budget $\varepsilon$ must be divided between them. 
Although recent work~\cite{DBLP:journals/corr/abs-2305-15394} demonstrates that private histograms can help reduce the budget consumption, this strategy inevitably introduces a substantial amount of noise. 
To mitigate this issue, alternative approaches proposed to randomly determine all splitting attributes and values~\cite{jagannathan2012enabling,DBLP:conf/ausai/FletcherI15,fletcher2016smooth}, and only access the data to populate the leaves. 
Even if random splits may partition the data imperfectly, this reduction in the amount of noise required to ensure DP generally allows for better predictive performance. 
Note that this specific aspect makes no difference in our attack's design, since the reconstruction model does not directly exploit the way the splits are determined. 
Our proposed formulation is indeed completely agnostic to the mechanism used to determine the split attributes and thresholds.

In non-DP RFs, bootstrap sampling (bagging) is commonly used to generate each tree's training set, by performing sampling with replacement from $\nexamples$ to $\nexamples$. 
While some DP RF approaches~\cite{rana} have adopted this strategy, providing tight and thorough DP guarantees remains challenging, as bootstrapped subsets are not disjoint and often contain multiple occurrences of certain examples. 
An alternative is to divide the original dataset into disjoint subsets to independently train each tree, enabling parallel composition of the privacy budget~\cite{DBLP:conf/ausai/FletcherI15,fletcher2016smooth}.
However, this approach may harm predictive performance as each tree is trained on significantly less data. Moreover, this also results in smaller leaves' class counts, which in turn are proportionally more affected by the noise added to enforce DP. 
A standard approach is to learn all trees on the full training dataset, even though it requires dividing the privacy budget across trees~\cite{jagannathan2012enabling,DBLP:conf/ausai/FletcherI15,fletcher2016smooth,Singh,DBLP:conf/ausdm/Fletcher015}. 
Although this would require modifications in the CP formulation, our attack could be extended to handle bagging as in~\cite{ferry2024trained}.
However, if the trees are trained on disjoint subsets, a per-tree probabilistic reconstruction approach such as the one proposed by~\cite{ferry2024probabilistic} suffices, since nothing can be gained by relating the information discovered across different trees.

Finally, some approaches such as IBM's \texttt{diffprivlib} implementation~\cite{DBLP:journals/corr/abs-1907-02444}, release only (noisy) majority labels (\emph{e.g.}, using the Permute-and-Flip mechanism~\cite{mckenna2024permuteflip}) within the leaves of DP RFs to conserve the privacy budget instead of providing full class counts. 
However, this strategy prevents the use of the traditional soft voting scheme for inference, relying instead on simple majority voting, which may degrade predictive performance (as evidenced in real-world applications, \emph{e.g.}~\cite{saqlain2019voting,sherazi2021soft}). 
Furthermore, access to class counts is often necessary for auditing purposes, such as estimating prediction uncertainty or checking calibration. Adapting our attack to the setting where only majority labels are released would require modifying the objective and introducing additional variables to model the likelihood of producing each observed (noisy) majority label from the guessed class counts.

\section{Conclusion}
\label{sec:conclusion}

This work has demonstrated that even with a low privacy budget, accurate reconstruction of the training examples of an $\varepsilon$-DP RF remains possible. 
Crucially, it shows that a model can comply with rigorous DP guarantees while still providing significant information regarding its training dataset to potential reconstruction attackers. 
This highlights that privacy protection cannot be solely reduced to choosing a sufficiently small $\varepsilon$. 
Rather, the design of the DP mechanism itself plays a fundamental role in mitigating reconstruction risks. 
This aligns with recent findings on privacy attacks  (\emph{i.e.}, re-identification and attribute inference) against local-differentially private protocols, in which the attack success depends not only on $\varepsilon$ but also on the mechanism's structure~\cite{DBLP:journals/pvldb/ArcoleziGCP23}. 
Similarly, our results suggest that a careful selection of the hyperparameters used to tune the RF and the DP mechanism allows for better trade-offs between predictive performance and empirical protection against reconstruction attacks.

The research avenues connected to this work are numerous.
One first direction involves extending the proposed reconstruction framework to DP RFs trained with alternative mechanisms (\emph{e.g.}, adding noise from different distributions or distributing it adaptively across the different algorithm components) or revealing different information (\emph{e.g.}, only majority labels in leaves), as discussed in Section~\ref{sec:related_work}. 
Importantly, any available \emph{prior} on the data distribution can be seamlessly integrated into the likelihood objective, potentially strengthening the attack.
Another promising perspective is to adapt our formulation to other types of tree ensembles, such as differentially private gradient boosting decision trees~\cite{DBLP:conf/aaai/LiWWH20}, in order to better characterize how correlations between trees influence reconstruction performance. Such an extension would, however, require substantial modifications, as even the adaptation of the attack to non-DP ensembles is non-trivial.

More broadly, our approach lays the foundation for a systematic privacy evaluation methodology grounded on mathematical programming, which integrates available knowledge from the training model as constraints and guides the search towards a most likely solution according to the underlying DP mechanism specifications. 
This framework can serve as a useful and mathematically-rigorous way of benchmarking DP strategies, helping to analyze how different DP configurations influence the trade-off between privacy and model utility. 
Understanding this interplay is crucial for developing machine learning models that are not only privacy-preserving in theory but also practically usable and robust against real-world privacy attacks. 
In particular, highlighting the privacy risks that may arise when deploying DP models in sensitive applications is essential to clarify the limitations of current mechanisms and to guide practitioners in developing more robust approaches. In this sense, our work contributes to the responsible and secure deployment of machine learning systems.

\section*{LLM usage considerations}

LLMs are not part of the paper's methodology. Their use was limited to minor editorial assistance (partially rephrasing a small number of sentences), and all outputs were carefully inspected and edited by the authors to ensure accuracy and originality.

\section*{Acknowledgments}

This research was enabled by support provided by Calcul Québec and the Digital Research Alliance of Canada, as well as funding from the SCALE AI Chair in Data-Driven Supply Chains. It was also supported by the \emph{Fonds de recherche du Québec} -- \emph{Nature et technologies (FRQNT)} through a Team Research Project \emph{(327090)}.

\bibliographystyle{IEEEtran}
\bibliography{IEEEabrv,paper}

\appendices

\clearpage 

\onecolumn 

\section{Reconstruction Attack when $\nexamples$ is Unknown}\label{appendix:cp_model_n_unknown}

\subsection{Confidence Interval on $\nexamples$}

While reconstruction attacks targeting a complete dataset recovery in the literature generally assume knowledge of its size $\nexamples$~\cite{dwork2017exposed}, our approach can be slightly adapted to be applicable without this information. 
In this section, we provide and empirically evaluate a CP formulation derived from the one presented in Section~\ref{sec:attack} that does not rely on knowing $\nexamples$.
While this setup is both more generic and more realistic, it is also more challenging.

When $\nexamples$ is unknown, it becomes a variable to infer. 
To bound the search, we will first calculate a confidence interval $\llbracket \nexamples_{\textsc{min}}, \nexamples_{\textsc{max}} \rrbracket$.
According to the DP RF algorithm in Section~\ref{sec:dp_rf}, each training example is used exactly once per tree. 
Thus, for a tree $\tree \in \forest$, the sum of the non-noisy leaves' counts equals $\nexamples$.
In our case, since the Laplace noise is centered at zero, for a tree $\tree \in \forest$, we can expect the sum $\nexamples_\tree^* = \sum\limits_{\node,\class} \nodesupport[\tree,\node,\class]^*$ to lie close to $\nexamples$.

We know that the noisy counts are computed by the Laplace mechanism as follows:
\[
\nodesupport[\tree,\node,\class]^* = \textrm{int}(\nodesupport[\tree,\node,\class] + \noisevar_{\tree\node\class}) = \nodesupport[\tree,\node,\class]  +  \textrm{int}( \noisevar_{\tree\node\class}) ,
\]
in which $\noisevar_{\tree\node\class}$ is a random variable sampled from a Laplace distribution \( \text{Lap}(1 / \varepsilon_\node)\) and $\textrm{int}(.)$ is the integer part function. 
Recall that the per-tree DP budget $\varepsilon_\node$ is computed by dividing the total privacy budget among the trees: $\varepsilon_\node = \varepsilon/\lvert \forest \rvert$.
The variance of each $\noisevar_{\tree\node\class}$ is $\text{Var}(\noisevar_{\tree\node\class}) = 2 / \varepsilon_\node^2$. 
Since the $\noisevar_{\tree\node\class}$ random variables are i.i.d., the variance of their sum $\sumnoisevars_\tree$ (whose $\nexamples_\tree^*$ is a realization) is:
\begin{align*}
     \text{Var}(\sumnoisevars_\tree) &= \text{Var}\left(\sum\limits_{\node \in \leaves[\tree]} \sum\limits_{\class \in \classes} \noisevar_{\tree\node\class}\right) 
     = \sum\limits_{\node \in \leaves[\tree]} \sum\limits_{\class \in \classes} \text{Var}(\noisevar_{\tree\node\class})    
    = \vert \leaves[\tree] \rvert \cdot \lvert \classes \rvert \cdot \frac{2}{\varepsilon_\node^2},
\end{align*} 
with a standard deviation:
\[
\sigma_{\sumnoisevars_\tree} = \frac{\sqrt{2 \cdot \lvert \leaves[\tree] \rvert \cdot \lvert \classes \rvert}}{\varepsilon_\node}.
\]

Let $\nexamples^* = \frac{1}{\lvert \forest \rvert} \sum_{\tree \in \forest} \nexamples_\tree^*$ be the average noisy count across all trees. 
By the Central Limit Theorem, $\nexamples^*$ approximately follows a normal distribution for large $\lvert \forest \rvert$. 
Thus, the 95\% confidence interval $\llbracket \nexamples_{\textsc{min}}, \nexamples_{\textsc{max}} \rrbracket$ can be defined as:
\[
    \nexamples_{\textsc{max}} = \nexamples^* + t_{95} \frac{\sigma_{\sumnoisevars_\tree}}{\sqrt{\lvert \forest \rvert}} \quad \text{and} \quad \nexamples_{\textsc{min}} = \max\left(1,\nexamples^* - t_{95} \frac{\sigma_{\sumnoisevars_\tree}}{\sqrt{\lvert \forest \rvert}}\right),
\]
in which $t_{95}$ is the Student's t-coefficient for a 95\% confidence level according to $\lvert \forest \rvert$.

\subsection{Adapted Constraint Programming Formulation}

As in the formulation presented in Section~\ref{sec:attack}, for each leaf $\node \in \leaves[\tree]$ of each tree $\tree \in \forest$, we let variables $\nodesupport[\tree,\node,\class] \in \{\max(0,\nodesupport[\tree,\node,\class]^*-\gamma), \dots, \nodesupport[\tree,\node,\class]^* +\gamma \}$ and $\Delta_{\tree\node\class } \in \{ -\gamma, \dots, \gamma \}$ respectively model the (guessed) number of examples of class $\class$ within this leaf and the amount of noise added to it.
We also use variable $\smash{\widetilde{\nexamples}}\in\{\nexamples_{\textsc{min}},\dots, \nexamples_{\textsc{max}}\}$ to encode the (guessed) number of reconstructed examples.

Since $\nexamples$ is unknown, we define variables for up to $\nexamples_{\textsc{max}}$ examples in the worst-case. 
For each example $\example \in \{1, \dots, \nexamples_{\textsc{max}}\}$, $\varx[\example,\feature] \in \{0,1\}$ represents the value of its attribute $\feature \in \{1, \dots, \nattributes\}$ in the reconstruction, while $\varz[\example,\class] \in \{0,1\}$ equals $1$ if and only if it belongs to class $\class$.
Although these reconstruction variables are defined for $\nexamples_{\textsc{max}}$ examples, the attack's output is obtained from the first $\widetilde{\nexamples}$ examples.

Finally, we define a set of variables modelling the path of each example $\example \in \{1,\dots,\nexamples_{\textsc{max}}\}$ through each tree $\tree \in \forest$: variable $\varyb[\tree,\node,\example,\class] \in \{0,1\}$ indicates if it is classified in class $\class \in \classes$ by leaf $\node \in \leaves[\tree]$.
Given these variables, as previously, we define constraints to compute class counts and connect them to the assignments of examples to leaves:
\begin{align}
    &\sum\limits_{\node \in \leaves[\tree]} \sum\limits_{\class \in \classes} \nodesupport[\tree,\node,\class] = \widetilde{\nexamples} &\tree \in \forest \label{constr_count_sum_N_bis}\\
     &\sum\limits_{\class \in \classes}\varz[\example,\class]=1 &\example \in \{1,\dots,\nexamples_{\textsc{max}}\}\label{constr_exactly_one_class_bis}\\
    &\varz[\example,\class] = 0 \Rightarrow\sum\limits_{\tree \in \forest} \sum\limits_{\node \in \leaves[\tree]} \varyb[\tree,\node,\example,\class] = 0 &\example \in \{1,\dots,\nexamples_{\textsc{max}}\},~\class \in \classes\label{constr_not_in_class_counts_bis}\\
    &\example > \widetilde{\nexamples}  \Rightarrow\sum\limits_{\tree \in \forest} \sum\limits_{\node \in \leaves[\tree]}\sum\limits_{\class \in \classes} \varyb[\tree,\node,\example,\class] = 0 &\example \in \{1,\dots,\nexamples_{\textsc{max}}\}\label{constr_N_examples_in_reco}\\
    & \sum_{\example=1}^{\smash{\nexamples_{\textsc{max}}}} \varyb[\tree,\node,\example,\class] =\nodesupport[\tree,\node,\class] &\tree \in \forest,~\node \in \leaves[\tree],~\class \in \classes\label{constr_counts_flow_bis}
\end{align}
Constraint~\eqref{constr_count_sum_N_bis} ensures that the examples' counts sum up to $\widetilde{\nexamples}$ within each tree.
Constraint~\eqref{constr_exactly_one_class_bis} guarantees that each example is associated with exactly one class.
Constraint~\eqref{constr_not_in_class_counts_bis} enforces that each example can only contribute to the counts of its class. 
Note that this constraint differs from the formulation presented in Section~\ref{sec:attack}. 
Indeed, if $\varz[\example,\class] = 1$ but $\example > \widetilde{\nexamples}$, example $\example$ is actually not part of the reconstructed dataset and hence is not assigned to any leaf of any tree, as stated in Constraint~\eqref{constr_N_examples_in_reco}.
Constraint~\eqref{constr_counts_flow_bis} ensures consistency between the examples' assignments to the leaves and the (guessed) counts.

Finally, the rest of the model remains unchanged, the following constraint enforces consistency between the examples' assignments to the leaves, and the value of their attributes:
\begin{align}
\sum\limits_{\class \in \classes} \varyb[\tree,\node,\example,\class] = 1 \Rightarrow &\left(\displaystyle\bigwedge_{i\in\Phi_v^+} \varx[k,i] = 1 \right) \wedge
\left( \displaystyle\bigwedge_{i\in \Phi_v^-} \varx[k,i] = 0 \right) & \tree \in \forest,~\node \in \leaves[\tree],~\example \in \{1,\dots,\nexamples_{\textsc{max}}\},\label{constr_flow_attr_values_bis}
\end{align}
The original objective is also retained with $\Delta_{\tree\node\class} = \nodesupport[\tree,\node,\class]^{*} - \nodesupport[\tree,\node,\class]$:
\begin{align}
    \max \ \sum\limits_{\tree \in \forest}\sum\limits_{\node \in \leaves[\tree]} \sum\limits_{\class \in \classes} \sum\limits_{\noiseval = -\gamma}^{\gamma} \log(p_\noiseval) \mathds{1}_{\Delta_{\tree\node\class} = \noiseval}.
\end{align}

\subsection{Experimental Evaluation} 

To assess the effectiveness of our reconstruction attack in a context where the number of training examples $\nexamples$ is unknown, we reproduce the experimental setup as described in Section~\ref{subsec:setup}, using our modified attack formulation introduced in the previous subsection. 
The computation of the reconstruction error must be slightly adapted. 
Indeed when $\nexamples$ is known, the attack's performance is assessed by determining the correspondence between reconstructed and original examples using a minimum-cost matching. 
The average distance between these matched pairs defines the reconstruction error.
However, when $\nexamples$ is unknown, it is possible that the inferred reconstruction contains more (or less) than $\nexamples$ examples, making this pairwise matching not directly applicable. 
In the case in which the number of reconstructed examples exceeds the actual one ($\widetilde{\nexamples} > \nexamples$), we randomly subsample $\nexamples$ examples from the reconstructed dataset.
In contrast, if the number of reconstructed examples is smaller than the actual one ($\widetilde{\nexamples} < \nexamples$), we randomly duplicate some reconstructed examples. 
This process yields exactly $\nexamples$ reconstructed examples. 
The same minimum-cost matching is then performed and the average distance between pairs of matched examples defines the reconstruction error. 

The results of these experiments are presented in Table~\ref{tab:error-values_N_var} for all datasets, forest sizes $\lvert \forest \rvert$ and tree depths $\depth$. 
Comparing these results with those in Table~\ref{tab:error-values} reveals two key trends. 
First, a few more runs failed to retrieve a feasible reconstruction within the given time frame ($2$ hours), indicating that the attack is computationally more expensive in this setup. 
This is expected, as the modified CP model introduces additional variables and constraints, leading to a larger search space and calling for longer solution times. 
Still, in most cases, the attack finds a feasible reconstruction. 
Second, and more importantly, the reconstruction error remains comparable to the case in which $\nexamples$ is known. 
Despite the increased computational cost, the attack is still able to retrieve meaningful information encoded by the forest regarding its training data in most cases, thereby confirming \textbf{Result~6}.

\begin{table*}[htbp]
\centering
\caption{Experimental results without assuming the training set size $\nexamples$ is known.
The table reports the average reconstruction error across varying numbers of trees $\lvert \forest \rvert$, tree depths $d$ and privacy budgets $\varepsilon$ for the three datasets. 
The cases for which the solver did not find a feasible reconstruction within the $2$-hour time limit are indicated by ``--''.}
\label{tab:error-values_N_var}
\begin{tabular}{llccccccccc}
\toprule
                                &                     & \multicolumn{3}{c}{Adult} & \multicolumn{3}{c}{COMPAS} & \multicolumn{3}{c}{Default Credit} \\
\cmidrule(lr){3-5} \cmidrule(lr){6-8} \cmidrule(lr){9-11}
                                &                     & $d=3$   & $d=5$  & $d=7$  & $d=3$   & $d=5$   & $d=7$  & $d=3$      & $d=5$      & $d=7$     \\
\midrule
\multirow{6}{*}{$\lvert \forest \rvert = 1$}  & $\varepsilon = 0.1$ & 0.27 & 0.23 & 0.20 & 0.28 & 0.25 & 0.17 & 0.34 & 0.31 & 0.28 \\
& $\varepsilon = 1$ & 0.25 & 0.21 & 0.17 & 0.25 & 0.20 & 0.14 & 0.33 & 0.28 & 0.24 \\
& $\varepsilon = 5$ & 0.24 & 0.20 & 0.17 & 0.25 & 0.17 & 0.13 & 0.32 & 0.28 & 0.23 \\
& $\varepsilon = 10$ & 0.24 & 0.19 & 0.16 & 0.25 & 0.19 & 0.13 & 0.32 & 0.28 & 0.24 \\
& $\varepsilon = 20$ & 0.24 & 0.21 & 0.15 & 0.25 & 0.19 & 0.13 & 0.32 & 0.28 & 0.24 \\
& $\varepsilon = 30$ & 0.24 & 0.21 & 0.15 & 0.25 & 0.19 & 0.14 & 0.32 & 0.28 & 0.24 \\
\midrule
\multirow{6}{*}{$\lvert \forest \rvert = 5$}  & $\varepsilon = 0.1$ & 0.22 & 0.22 & 0.26 & 0.21 & 0.21 & 0.36 & 0.30 & 0.27 & 0.31 \\
& $\varepsilon = 1$ & 0.19 & 0.17 & 0.22 & 0.18 & 0.12 & 0.14 & 0.26 & 0.21 & 0.24 \\
& $\varepsilon = 5$ & 0.18 & 0.14 & 0.17 & 0.11 & 0.09 & 0.14 & 0.23 & 0.17 & 0.19 \\
& $\varepsilon = 10$ & 0.17 & 0.12 & 0.13 & 0.10 & 0.07 & 0.10 & 0.22 & 0.15 & 0.16 \\
& $\varepsilon = 20$ & 0.15 & 0.12 & 0.09 & 0.10 & 0.06 & 0.05 & 0.22 & 0.16 & 0.13 \\
& $\varepsilon = 30$ & 0.15 & 0.12 & 0.09 & 0.09 & 0.06 & 0.04 & 0.21 & 0.15 & 0.11 \\
\midrule
\multirow{6}{*}{$\lvert \forest \rvert = 10$}  & $\varepsilon = 0.1$ & 0.28 & 0.31 & 0.28 & 0.23 & 0.27 & 0.25 & 0.29 & 0.33 & 0.32 \\
& $\varepsilon = 1$ & 0.17 & 0.21 & 0.28 & 0.13 & 0.16 & 0.20 & 0.22 & 0.26 & 0.28 \\
& $\varepsilon = 5$ & 0.17 & 0.28 & 0.27 & 0.12 & 0.20 & 0.19 & 0.20 & 0.27 & 0.23 \\
& $\varepsilon = 10$ & 0.14 & 0.17 & 0.18 & 0.07 & 0.10 & 0.20 & 0.16 & 0.17 & 0.20 \\
& $\varepsilon = 20$ & 0.13 & 0.12 & 0.13 & 0.06 & 0.07 & 0.08 & 0.14 & 0.14 & 0.15 \\
& $\varepsilon = 30$ & 0.13 & 0.11 & 0.11 & 0.05 & 0.05 & 0.05 & 0.14 & 0.12 & 0.13 \\
\midrule
\multirow{6}{*}{$\lvert \forest \rvert = 20$}  & $\varepsilon = 0.1$ & 0.27 & 0.29 & 0.35 & 0.29 & 0.33 & 0.35 & 0.31 & 0.33 & 0.39 \\
& $\varepsilon = 1$ & 0.20 & 0.26 & 0.30 & 0.19 & 0.20 & 0.29 & 0.26 & 0.28 & 0.33 \\
& $\varepsilon = 5$ & 0.24 & 0.24 & 0.27 & 0.20 & 0.23 & 0.21 & 0.29 & 0.28 & 0.30 \\
& $\varepsilon = 10$ & 0.16 & 0.22 & 0.26 & 0.13 & 0.25 & 0.30 & 0.20 & 0.23 & 0.28 \\
& $\varepsilon = 20$ & 0.12 & 0.15 & 0.18 & 0.08 & 0.09 & 0.20 & 0.14 & 0.15 & 0.21 \\
& $\varepsilon = 30$ & 0.11 & 0.12 & 0.16 & 0.06 & 0.07 & 0.11 & 0.13 & 0.13 & 0.16 \\
\midrule
\multirow{6}{*}{$\lvert \forest \rvert = 30$}  & $\varepsilon = 0.1$ & 0.36 & 0.34 & - & 0.32 & 0.33 & - & 0.36 & 0.37 & - \\
& $\varepsilon = 1$ & 0.26 & 0.33 & 0.27 & 0.20 & 0.34 & 0.34 & 0.30 & 0.38 & - \\
& $\varepsilon = 5$ & 0.25 & 0.25 & 0.33 & 0.22 & 0.20 & 0.28 & 0.27 & 0.26 & 0.32 \\
& $\varepsilon = 10$ & 0.21 & 0.22 & 0.24 & 0.18 & 0.28 & 0.23 & 0.25 & 0.26 & 0.31 \\
& $\varepsilon = 20$ & 0.13 & 0.20 & 0.22 & 0.12 & 0.13 & 0.18 & 0.15 & 0.16 & 0.27 \\
& $\varepsilon = 30$ & 0.11 & 0.15 & 0.19 & 0.07 & 0.09 & 0.20 & 0.14 & 0.16 & 0.18 \\
\midrule
\\\bottomrule\end{tabular}
\end{table*}

\clearpage 

\section{Determining a Search Interval for Noise Values}\label{appendix:width_delta_search_interval}

\begin{figure}[htbp]
    \vskip 0.2in
    \begin{center}
    \centerline{\includegraphics[width=0.75\linewidth]{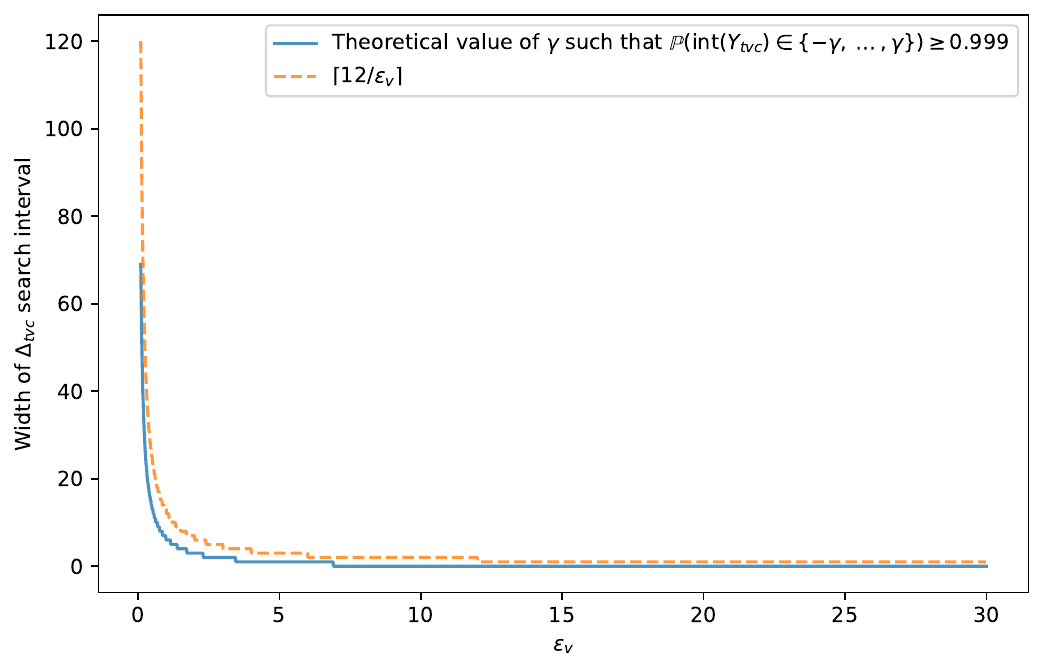}}
    \caption{Width of $\Delta_{\tree\node\class }$ search interval as a function of $\varepsilon_v$. 
    As expected, the magnitude of the noise added decreases when the privacy budget increases, resulting in a smaller search interval.}
    \label{intervalle_recherche}
    \end{center}
    \vskip -0.2in
\end{figure}

Rather than modelling all possible (integer) random noise values $\noisevar_{\tree\node\class}$ within $\mathbb{Z}$, our CP model focuses on a (sufficient) predefined range $\{ -\gamma,\dots, \gamma \}$ such that $\mathbb{P}(\textrm{int}(\noisevar_{\tree\node\class}) \in \{ -\gamma,\dots, \gamma \} ) \geq 0.999$. 
In practice, we use $\gamma = \lceil 12 / \varepsilon_\node \rceil$. 
As displayed in Figure~\ref{intervalle_recherche}, this value is suitable for all considered privacy budgets $\varepsilon_\node$.
Indeed, it always leads to the definition of slightly wider search intervals for the variables modelling the noise, theoretically guaranteeing that the probability of the actual noise value being within it is greater than $0.999$.

\section{Additional Experimental Results}\label{appendix:complete_results_tradeoffs_perf_reconstr_error}

\subsection{Comparison with State-of-the-Art RFs}
\label{appendix:rfs_perfs_comparisons}

An important insight is the quantification of the accuracy loss (on both training and test sets) attributable to the design of the DP RF algorithm~\cite{jagannathan2012enabling}, independently of the noise introduced by DP. 
Specifically, this includes omitting bootstrap sampling and selecting split values and attributes at random rather than using an information gain criterion.

We hereafter provide additional results comparing the training and test accuracy of (i) DP RFs built using the algorithm considered in our paper with a large privacy budget ($\varepsilon=30$) and (ii) non-DP RFs built using the CART algorithm implemented in the \texttt{scikit-learn}~\cite{scikit-learn} Python library, for the same hyperparameters (numbers of trees and maximum depth). 
Table~\ref{tab:dprfs_vs_sklearnrfs} reports the results obtained for a representative RF configuration of $\lvert\mathcal{T}\rvert=10$ trees of maximum depth $d=5$, for each of the three datasets considered in our experiments. 
As can be observed, for the setup considered in our paper ($N = 100$ training examples), the design of the DP RF algorithm results in some loss in training accuracy. 
This is expected, as the splits determined by CART tend to partition the data more effectively. 
However, this difference largely vanishes when considering test accuracy. 
A likely explanation is that the relatively small training set leads CART to overfit the data. This effect is mitigated in the DP RF algorithm, where splits are selected randomly.

\begin{table*}[h!]
    \centering
        \caption{Training and test accuracy of random forests with $\lvert\mathcal{T}\rvert=10$ trees of maximum depth $d=5$ trained either with our considered DP RF algorithm~\cite{jagannathan2012enabling} (for a privacy budget $\varepsilon=30$) or with the standard CART algorithm as implemented in the \texttt{scikit-learn}~\cite{scikit-learn} Python library. We report average and standard deviation over $5$ random seeds.}    \label{tab:dprfs_vs_sklearnrfs}
\begin{tabular}{@{}ccccc@{}}
\toprule
Dataset        & \begin{tabular}[c]{@{}c@{}}DP RF ($\varepsilon=30$)\\ Training Accuracy\end{tabular} & \begin{tabular}[c]{@{}c@{}}Sklearn non-DP RF\\ Training Accuracy\end{tabular} & \begin{tabular}[c]{@{}c@{}}DP RF ($\varepsilon=30$)\\ Test Accuracy\end{tabular} & \begin{tabular}[c]{@{}c@{}}Sklearn non-DP RF\\ Test Accuracy\end{tabular} \\ \midrule
Adult          & $0.84 \pm 0.01$                                                                      & $0.90 \pm 0.01$                                                               & $0.77 \pm 0.00$                                                                  & $0.78 \pm 0.00$                                                           \\ \midrule
COMPAS         & $0.74 \pm 0.00$                                                                      & $0.75 \pm 0.02$                                                               & $0.65 \pm 0.01$                                                                  & $0.63 \pm 0.01$                                                           \\ \midrule
Default Credit & $0.87 \pm 0.00$                                                                      & $0.93 \pm 0.01$                                                               & $0.78 \pm 0.00$                                                                  & $0.78 \pm 0.00$                                                           \\ \bottomrule
\end{tabular}
\end{table*}

\FloatBarrier

{\rebuttalsatmlmultiline
\subsection{Notes on the Design of the DP RF Learning Algorithm}\label{appendix:design_dp_rf_algo}

In the seminal work of~\cite{jagannathan2012enabling}, only the Laplace mechanism was considered to privatize the per-class per-leaf counts. In this appendix section, we examine whether relaxing pure $\varepsilon$-DP to approximate $(\varepsilon,\delta)$-DP (by replacing the Laplace mechanism with the Gaussian mechanism) could meaningfully improve the design or performance of the considered DP random forests. We first outline how such a modification would be implemented and recall the necessary background. We then provide qualitative and quantitative evidence showing that this relaxation offers no tangible benefit: under our setting, switching to $(\varepsilon,\delta)$-DP neither improves privacy guarantees nor enhances model utility (and in practice would degrade both).

\vspace{0.25\baselineskip}

\textbf{Learning (Approximate) DP Random Forests.}
The DP RFs considered in this paper satisfy \emph{pure} $\varepsilon$-DP and are trained using the algorithm introduced in \cite{jagannathan2012enabling}, summarized in Section \ref{sec:dp_rf}. This method constructs each tree by selecting split attributes and thresholds at random, and allocates the full privacy budget $\varepsilon$ to privatizing the per-class leaf counts. The latter are protected using the Laplace mechanism, which adds Laplace noise scaled to the query’s $\ell_1$-sensitivity. 

Replacing the Laplace mechanism with the Gaussian mechanism would instead add Gaussian noise scaled to the query’s $\ell_2$-sensitivity. However, due to the unbounded tails of the Gaussian distribution, there always exists a small region of rare outputs for which the multiplicative privacy bound may fail. For this reason, the resulting models would no longer satisfy pure $\varepsilon$-DP but rather \emph{approximate} $(\varepsilon,\delta)$-DP. More precisely, a randomized mechanism \(\mathcal{K}\) satisfies \( (\varepsilon, \delta) \)-DP if, for any pair of datasets \( \dataset_1 \) and \( \dataset_2 \) differing by the addition or removal of at most one record, and all \( U \subseteq \text{Range}(\mathcal{K}) \):
\[
\mathbb{P}(\mathcal{K}(\dataset_1) \in U) \leq e^\varepsilon \times \mathbb{P}(\mathcal{K}(\dataset_2) \in U) + \delta.
\]
Here, $\delta$ quantifies the probability of such tail events (where the privacy loss may exceed $\varepsilon$) and is typically required to be cryptographically small, e.g., $\delta = 1/\nexamples{^2}$ or less.

In our setup, observe that the $\ell_2$-sensitivity of counting queries is equal to their $\ell_1$-sensitivity, namely 1. 
Then, letting $(\varepsilon_v, \delta_v)$ be the DP budget to compute each class counts with the Gaussian mechanism, the noisy class counts are computed~as:
\[
\nodesupport[\tree,\node,\class]^* = \textrm{int}(\nodesupport[\tree,\node,\class] + \noisevar'_{\tree\node\class}) = \nodesupport[\tree,\node,\class]  +  \textrm{int}( \noisevar'_{\tree\node\class}) ,
\]
in which $\textrm{int}(.)$ is the integer part function and $\noisevar'_{\tree\node\class}$ is a random variable sampled from a Gaussian (normal) distribution \( \mathcal{N}(0, \sigma^2)\), whose variance is:
\[
\sigma^2=\frac{2\ln\!\left(1.25/\delta_v\right)}{\varepsilon_\node^2}.
\]

\textbf{DP budget using basic composition.}
Sequential and parallel composition also apply in the approximate $(\varepsilon,\delta)$-DP setting \cite{mcsherry2007mechanism}. In particular, applying several $(\varepsilon_i,\delta_i)$-DP mechanisms to the same dataset results in an overall guarantee of $(\sum_i \varepsilon_i, \sum_i \delta_i)$-DP, while applying them to disjoint subsets of the data yields $(\max_i \varepsilon_i, \max_i \delta_i)$-DP. The analysis of the DP guarantees for the modified learning algorithm therefore follows the same structure as in the pure DP case
: the global privacy budget $(\varepsilon,\delta)$ is first split across the trees in the forest.
Since the leaves of each tree have disjoint support, parallel composition applies within each tree. Likewise, within any given leaf, the per-class counting queries also operate on disjoint subsets and therefore satisfy parallel composition. 
As a result, the DP budget of the whole algorithm is $(\varepsilon, \delta)$, with:
\[
\varepsilon = \lvert \mathcal{T} \rvert  \varepsilon_v \quad \text{and} \quad \delta = \lvert \mathcal{T} \rvert  \delta_v.
\]

\textbf{DP budget using advanced composition.} Approximate DP allows the use of a different form of composition, known as advanced composition, instead of standard sequential composition. For large numbers of composition steps, advanced composition yields a tighter overall privacy bound. In our setup, when applying advanced composition across trees (combined with parallel composition within each tree), the DP budget of the entire algorithm becomes $(\varepsilon, \delta)$, with:
\[
\varepsilon = \sqrt{2 \lvert \mathcal{T} \rvert \ln{(1/\delta')}} \varepsilon_v + \lvert \mathcal{T} \rvert \varepsilon_v (e^{\varepsilon_v}-1) \quad \text{and} \quad \delta = \lvert \mathcal{T} \rvert  \delta_v + \delta'
\]
for any $\delta' > 0$.

\textbf{Visualization of the noise magnitude.}
The probability density functions of the noise added by the Laplace and Gaussian mechanisms (under either basic or advanced composition) are shown in Figure~\ref{fig:pdfs_noise_comparison}, for a DP RF with $\lvert \mathcal{T} \rvert = 10$ trees and a global DP budget of $\varepsilon = 10$ (and $\delta = 1/\nexamples^{2} = 10^{-4}$ for approximate DP). The resulting DP budgets allocated to compute each count (as determined by the composition techniques described above) are reported in the caption. As visible, the Gaussian mechanism requires adding noise of larger magnitude than the Laplace mechanism. Moreover, advanced composition does not help here – it actually worsens the per-count budget. Because the $\ell_1$ and $\ell_2$ sensitivities coincide and sequential composition occurs over only a small number of trees, using the Gaussian mechanism in this setting would degrade utility by injecting more noise, while simultaneously weakening the privacy guarantee. This makes the Laplace mechanism a preferable study target for the considered DP RF construction.

\begin{figure}[h!]
    \centering
    \includegraphics[width=0.6\linewidth]{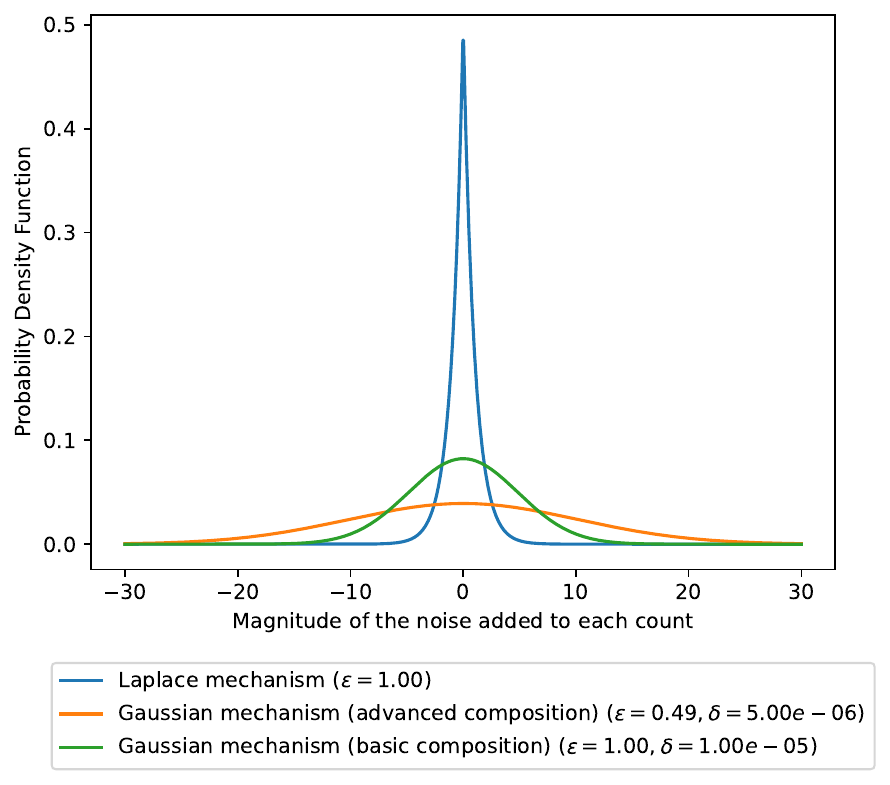}
    \caption{\rebuttalsatml{Comparison of the Probability Density Function of the noise added by different DP mechanisms to each per-class per-leaf count to protect a RF with $\lvert \mathcal{T} \rvert = 10$ trees with global DP budget $\varepsilon=10$ (and $\delta=10^{-4}$ for approximate DP).}}
    \label{fig:pdfs_noise_comparison}
\end{figure}

}

\subsection{Running Times}\label{appendix:all_running_times}

\updated{The average running time (over all performed experiments) to find a feasible reconstruction is below two minutes. %
Table~\ref{tab:solution-times} details the running time required by the CP solver to find a feasible reconstruction, averaged over the 5 random seeds for all considered datasets, numbers of trees $\lvert \forest \rvert$ within the target forest, tree depths $\depth$ and privacy budgets $\varepsilon$. 
Several general trends can be observed. 
Increasing the number of trees or their depth leads to larger CP models, including more variables, more constraints and larger search spaces. 
This logically increases the solver's running time. 
Considering tight privacy budgets also widens the solver's search space and running time. 
Nevertheless, in all cases, the CP solver is able to find a feasible reconstruction using only a fraction of its predefined 2-hours timeout, and can then use the remaining time to refine it and improve its objective function (corresponding to the noise likelihood, as explained in Section~\ref{sec:attack}).}

\begin{table*}[htb]
\centering
\caption{\updated{Average running time (s) to get a feasible reconstruction for different numbers of trees $\lvert \forest \rvert$ within the target forest, tree depths $\depth$ and privacy budgets $\varepsilon$ for the three datasets. The situation in which the solver failed to find a feasible reconstruction within the $2$-hour time limit are marked with~``--''.}}\label{tab:solution-times}
\begin{tabular}{llccccccccc}
\toprule
                                &                     & \multicolumn{3}{c}{Adult} & \multicolumn{3}{c}{COMPAS} & \multicolumn{3}{c}{Default Credit} \\
\cmidrule(lr){3-5} \cmidrule(lr){6-8} \cmidrule(lr){9-11}
                                &                     & $d=3$   & $d=5$  & $d=7$  & $d=3$   & $d=5$   & $d=7$  & $d=3$      & $d=5$      & $d=7$     \\
\midrule
\multirow{6}{*}{$\lvert \forest \rvert = 1$}  & $\varepsilon = 0.1$ & 1.81 & 9.73 & 24.98 & 1.95 & 7.10 & 22.77 & 2.65 & 9.48 & 28.37 \\
& $\varepsilon = 1$ & 0.56 & 3.90 & 16.57 & 0.51 & 2.75 & 10.68 & 0.56 & 3.63 & 15.53 \\
& $\varepsilon = 5$ & 0.45 & 3.15 & 13.57 & 0.49 & 2.61 & 9.89 & 0.53 & 3.60 & 14.84 \\
& $\varepsilon = 10$ & 0.47 & 3.16 & 13.78 & 0.47 & 2.65 & 10.03 & 0.51 & 3.69 & 16.69 \\
& $\varepsilon = 20$ & 0.46 & 1.95 & 13.35 & 0.39 & 2.26 & 9.66 & 0.37 & 1.65 & 12.21 \\
& $\varepsilon = 30$ & 0.41 & 1.57 & 12.69 & 0.43 & 2.50 & 9.74 & 0.48 & 2.78 & 15.54 \\
\midrule
\multirow{6}{*}{$\lvert \forest \rvert = 5$}  & $\varepsilon = 0.1$ & 14.35 & 47.58 & 123.09 & 10.41 & 39.33 & 96.10 & 12.94 & 41.43 & 128.34 \\
& $\varepsilon = 1$ & 6.61 & 22.23 & 60.73 & 4.84 & 24.05 & 67.07 & 4.52 & 23.36 & 56.40 \\
& $\varepsilon = 5$ & 3.72 & 23.87 & 53.88 & 3.72 & 25.14 & 71.13 & 3.56 & 28.15 & 79.50 \\
& $\varepsilon = 10$ & 4.23 & 26.25 & 75.20 & 3.82 & 22.80 & 73.11 & 3.00 & 24.55 & 88.45 \\
& $\varepsilon = 20$ & 2.95 & 24.62 & 85.41 & 3.32 & 20.51 & 63.55 & 3.82 & 22.55 & 54.54 \\
& $\varepsilon = 30$ & 2.68 & 24.47 & 110.77 & 3.60 & 27.20 & 51.81 & 3.27 & 22.18 & 118.82 \\
\midrule
\multirow{6}{*}{$\lvert \forest \rvert = 10$}  & $\varepsilon = 0.1$ & 34.90 & 104.13 & 284.07 & 27.20 & 107.22 & 422.43 & 31.08 & 112.55 & 442.63 \\
& $\varepsilon = 1$ & 17.79 & 47.11 & 157.09 & 18.73 & 44.59 & 103.73 & 14.70 & 46.66 & 109.73 \\
& $\varepsilon = 5$ & 10.88 & 46.25 & 93.64 & 9.19 & 38.99 & 83.29 & 12.10 & 45.57 & 97.78 \\
& $\varepsilon = 10$ & 10.88 & 34.03 & 98.43 & 10.72 & 59.48 & 125.38 & 12.11 & 65.36 & 122.64 \\
& $\varepsilon = 20$ & 10.35 & 42.09 & 106.39 & 8.62 & 41.33 & 119.52 & 12.33 & 51.44 & 124.82 \\
& $\varepsilon = 30$ & 8.68 & 49.37 & 90.21 & 12.32 & 48.60 & 87.92 & 9.17 & 116.85 & 284.96 \\
\midrule
\multirow{6}{*}{$\lvert \forest \rvert = 20$}  & $\varepsilon = 0.1$ & 90.69 & 293.24 & 1033.74 & 84.66 & 282.49 & 968.48 & 97.05 & 261.93 & 1022.18 \\
& $\varepsilon = 1$ & 32.68 & 92.48 & 357.22 & 26.43 & 89.08 & 300.26 & 30.97 & 85.97 & 354.49 \\
& $\varepsilon = 5$ & 22.18 & 65.19 & 167.35 & 22.18 & 68.28 & 206.02 & 17.92 & 66.48 & 225.78 \\
& $\varepsilon = 10$ & 22.39 & 68.52 & 234.35 & 27.08 & 67.73 & 254.61 & 19.43 & 71.50 & 242.74 \\
& $\varepsilon = 20$ & 28.65 & 225.42 & 360.96 & 25.57 & 125.94 & 279.20 & 21.58 & 243.11 & 890.48 \\
& $\varepsilon = 30$ & 35.18 & 233.34 & 387.37 & 26.55 & 72.33 & 285.36 & 25.58 & 127.86 & 511.48 \\
\midrule
\multirow{6}{*}{$\lvert \forest \rvert = 30$}  & $\varepsilon = 0.1$ & 185.31 & 842.61 & - & 170.20 & 699.23 & - & 150.73 & 752.15 & - \\
& $\varepsilon = 1$ & 47.45 & 152.51 & 611.71 & 39.69 & 146.58 & 666.69 & 41.66 & 158.16 & 451.77 \\
& $\varepsilon = 5$ & 29.43 & 101.53 & 405.87 & 26.72 & 95.50 & 259.35 & 31.40 & 89.66 & 438.50 \\
& $\varepsilon = 10$ & 31.83 & 92.64 & 363.51 & 28.17 & 86.97 & 333.50 & 30.04 & 96.03 & 380.74 \\
& $\varepsilon = 20$ & 34.24 & 87.08 & 357.17 & 58.27 & 212.95 & 375.59 & 32.66 & 119.71 & 413.38 \\
& $\varepsilon = 30$ & 45.91 & 410.98 & 536.70 & 32.32 & 217.46 & 828.98 & 36.58 & 232.30 & 862.32 \\
\bottomrule\end{tabular}
\end{table*}
\subsection{Trade-offs between Predictive Performance and Reconstruction Error}
\label{appendix:tradeoffs_detailed}
\begin{figure*}[htb]
\centering
\begin{subfigure}{0.5\textwidth}
    \centering
    \includegraphics[width=1.0\linewidth]{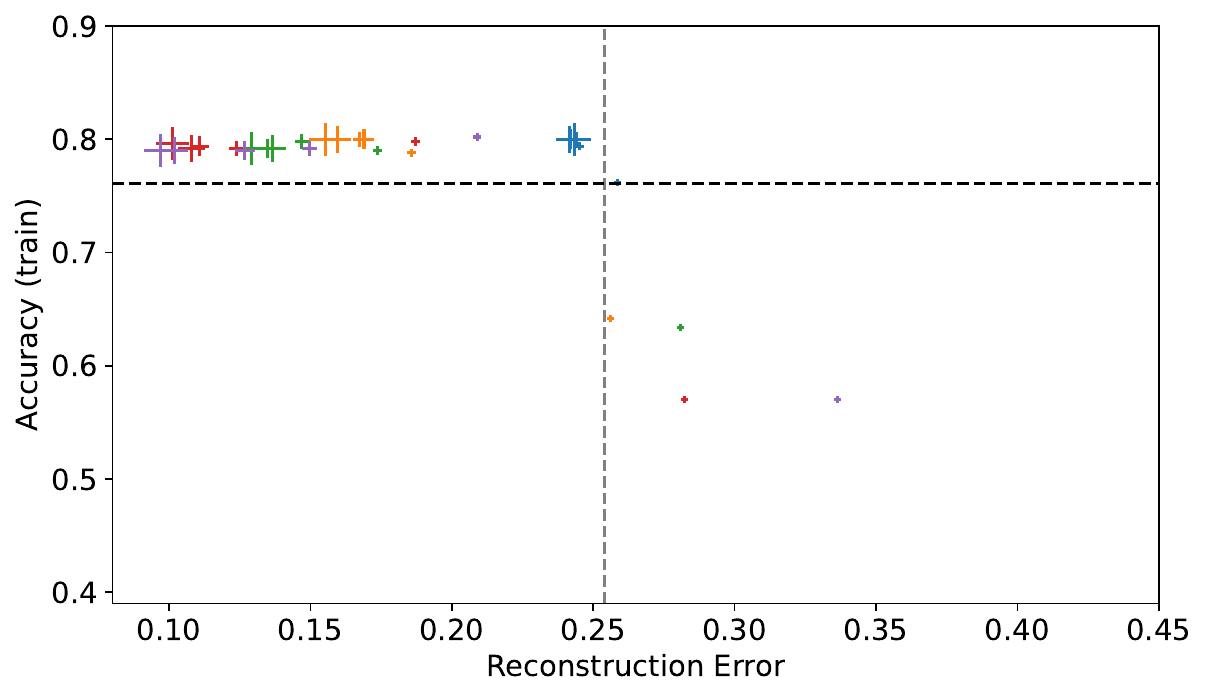}
    \caption{UCI Adult Income dataset, depth 3 decision trees.} \label{fig:results_tradeoffs_acc_reconstr_adult_depth_3}
\end{subfigure}%
\hfill
 \begin{subfigure}{0.5\textwidth}
    \centering
    \includegraphics[width=1.0\linewidth]{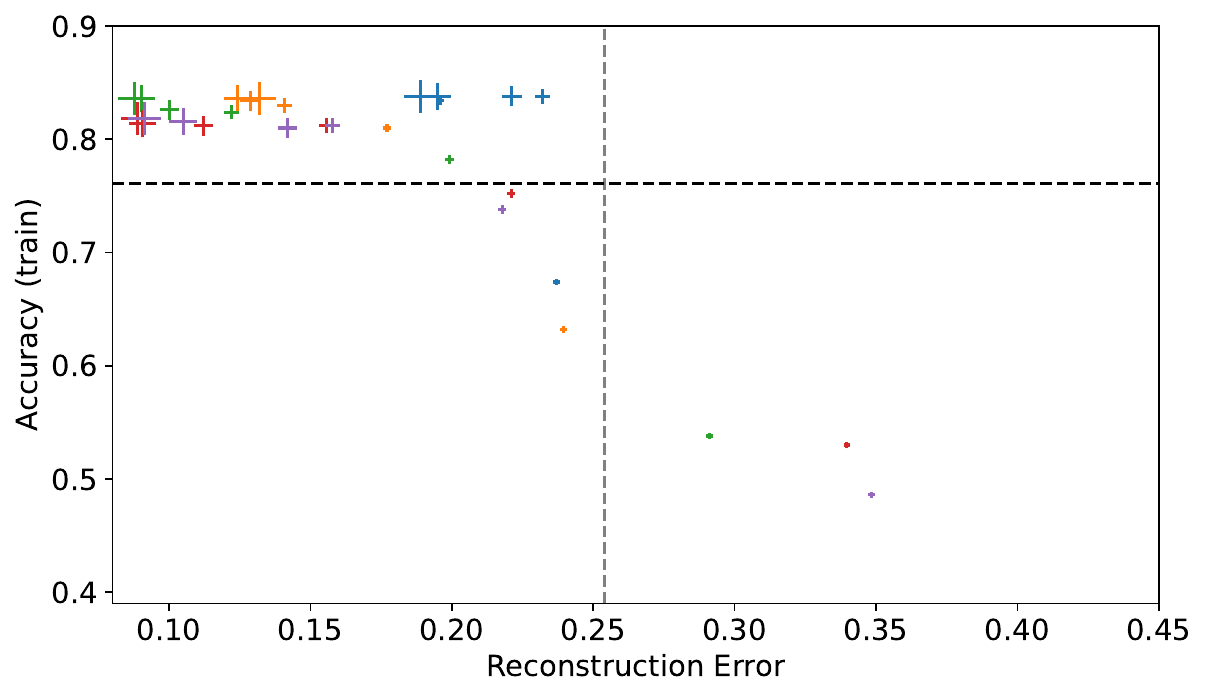}
    \caption{UCI Adult Income dataset, depth 5 decision trees.}
\label{fig:results_tradeoffs_acc_reconstr_adult_depth_5}
\end{subfigure}%

\vspace{10pt}

\begin{subfigure}{0.5\textwidth}
    \centering
    \includegraphics[width=1.0\linewidth]{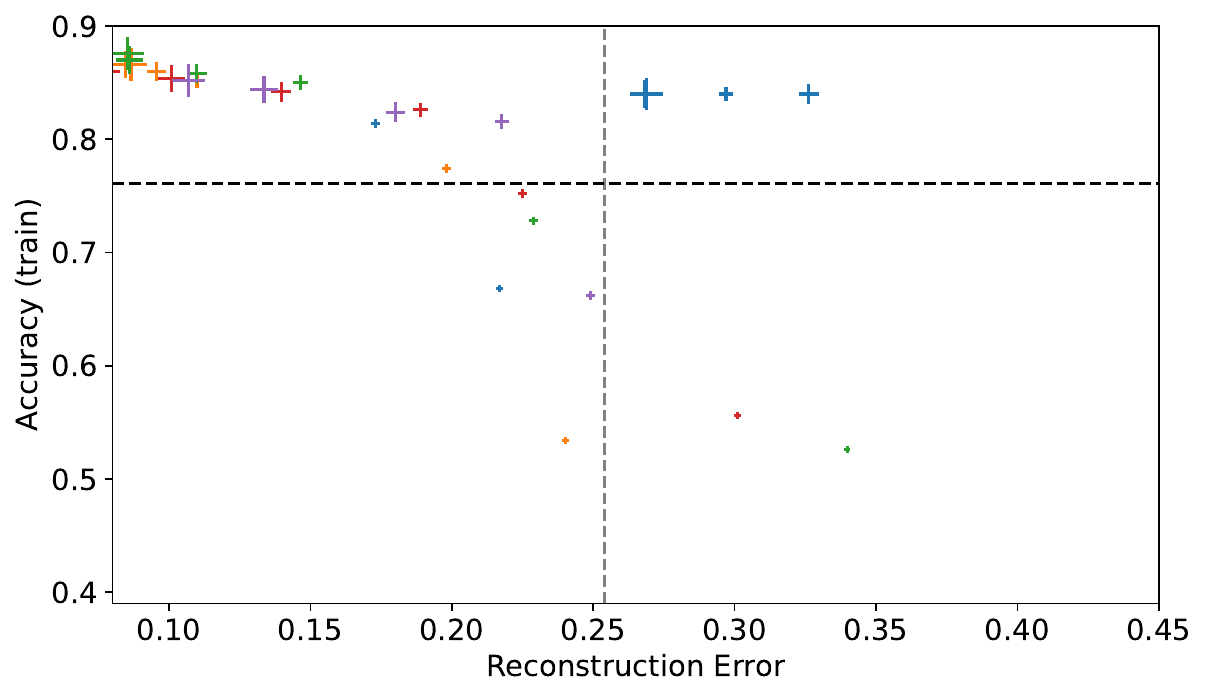}
    \caption{UCI Adult Income dataset, depth 7 decision trees.}
    \label{fig:results_tradeoffs_acc_reconstr_adult_depth_7}
\end{subfigure}

\vspace{10pt}

\begin{subfigure}{\textwidth}
    \centering
    \includegraphics[width=0.6\linewidth]{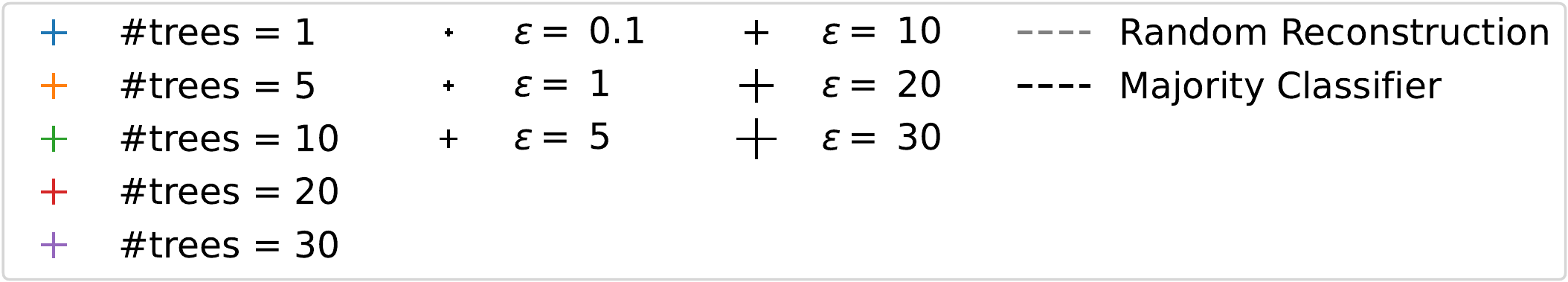}
\end{subfigure}
\caption{Average training accuracy of $\varepsilon$-DP RFs as a function of the reconstruction error of our attack for different numbers of trees $\vert \forest \rvert$ and privacy budgets $\varepsilon$, for the considered datasets and tree depths values.}\label{fig:results_all_tradeoffs_predictive_perf_vs_recontr_error}
\end{figure*}

\begin{figure*}[t!] %
\begin{subfigure}{0.5\textwidth}
    \centering
    \includegraphics[width=1.0\linewidth]{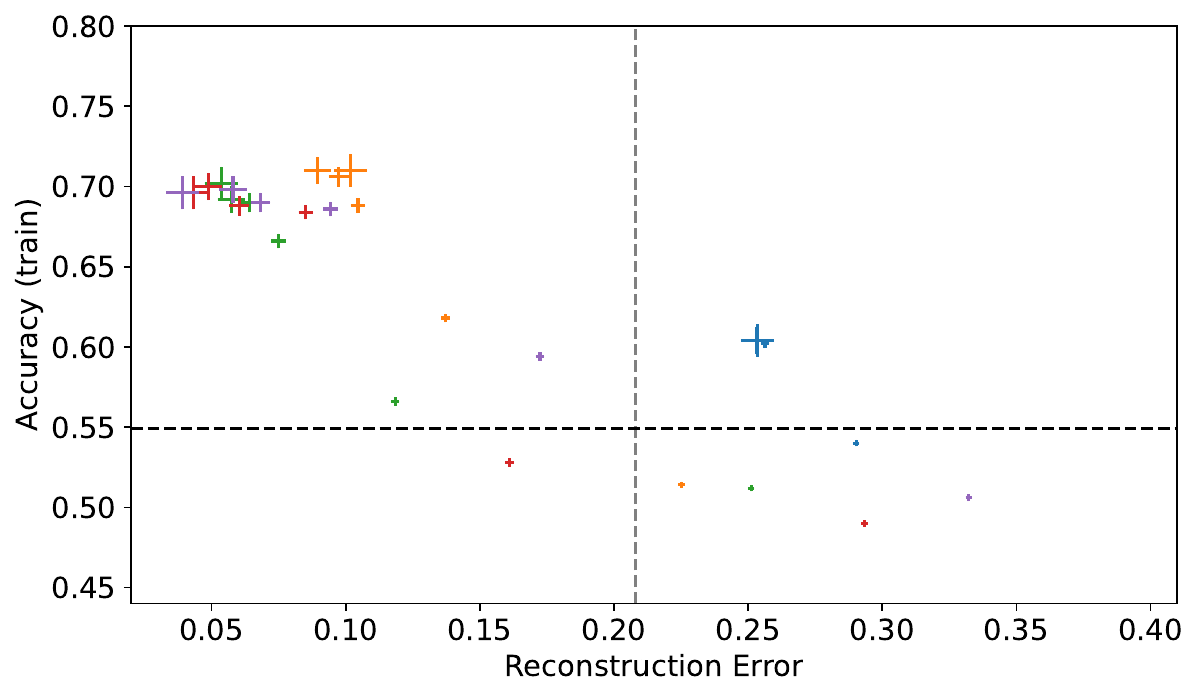}
    \caption{COMPAS dataset, depth 3 decision trees.}
    \label{fig:results_tradeoffs_acc_reconstr_COMPAS_depth_3}
\end{subfigure}
\hfill
\begin{subfigure}{0.5\textwidth}
    \centering
    \includegraphics[width=1.0\linewidth]{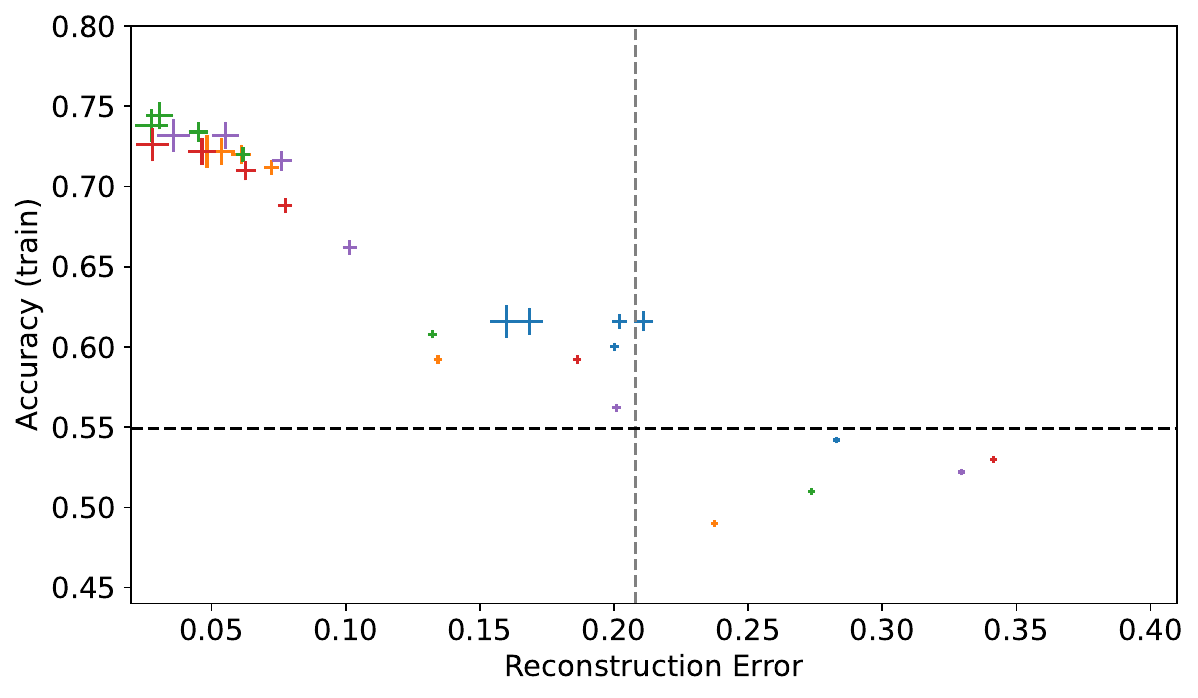}
    \caption{COMPAS dataset, depth 5 decision trees.}
    \label{fig:results_tradeoffs_acc_reconstr_COMPAS_depth_5}
\end{subfigure}

\vspace{10pt}

\begin{subfigure}{0.5\textwidth}
    \centering
    \includegraphics[width=1.0\linewidth]{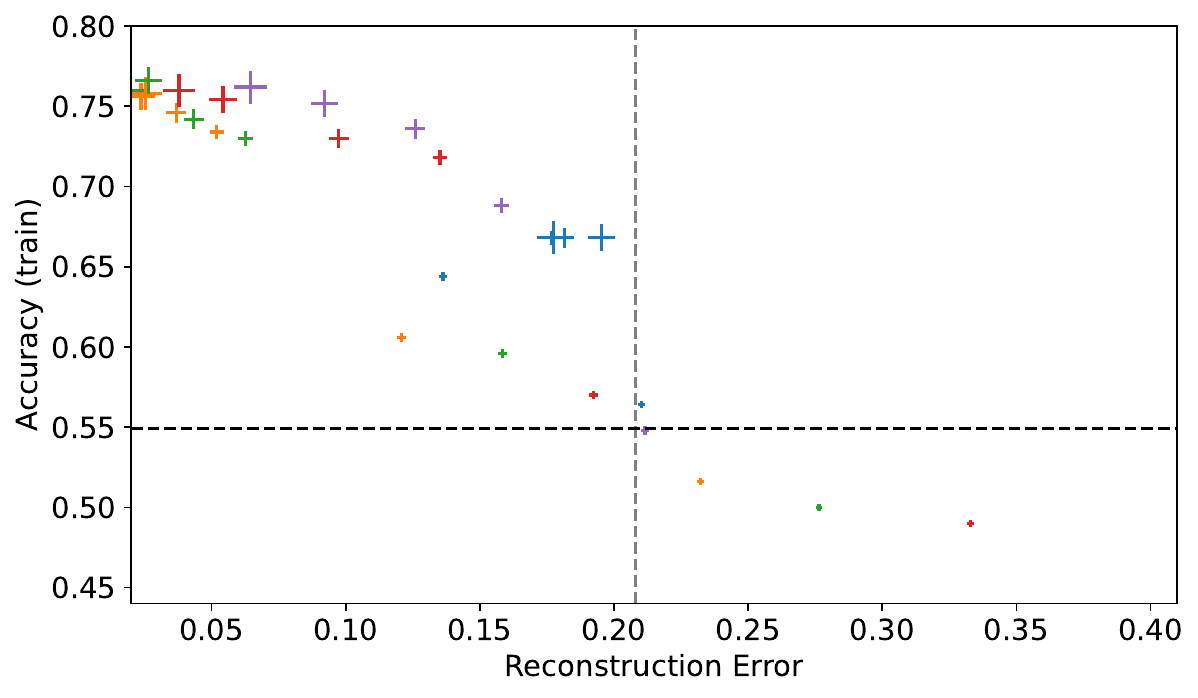}
    \caption{COMPAS dataset, depth 7 decision trees.}
    \label{fig:results_tradeoffs_acc_reconstr_COMPAS_depth_7}
\end{subfigure}
\hfill
\begin{subfigure}{0.5\textwidth}
    \centering
    \includegraphics[width=1.0\linewidth]{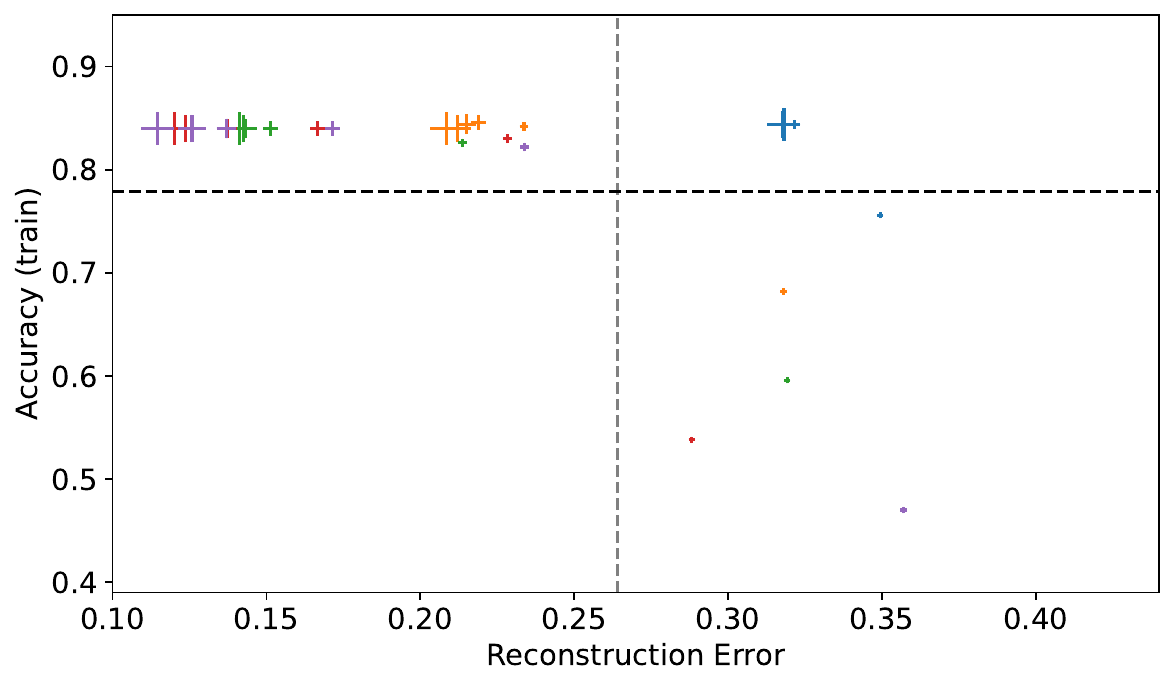}
    \caption{Default Credit dataset, depth 3 decision trees.}
    \label{fig:results_tradeoffs_acc_reconstr_default_credit_depth_3}
\end{subfigure}

\vspace{10pt}

\begin{subfigure}{0.5\textwidth}
    \centering
    \includegraphics[width=1.0\linewidth]{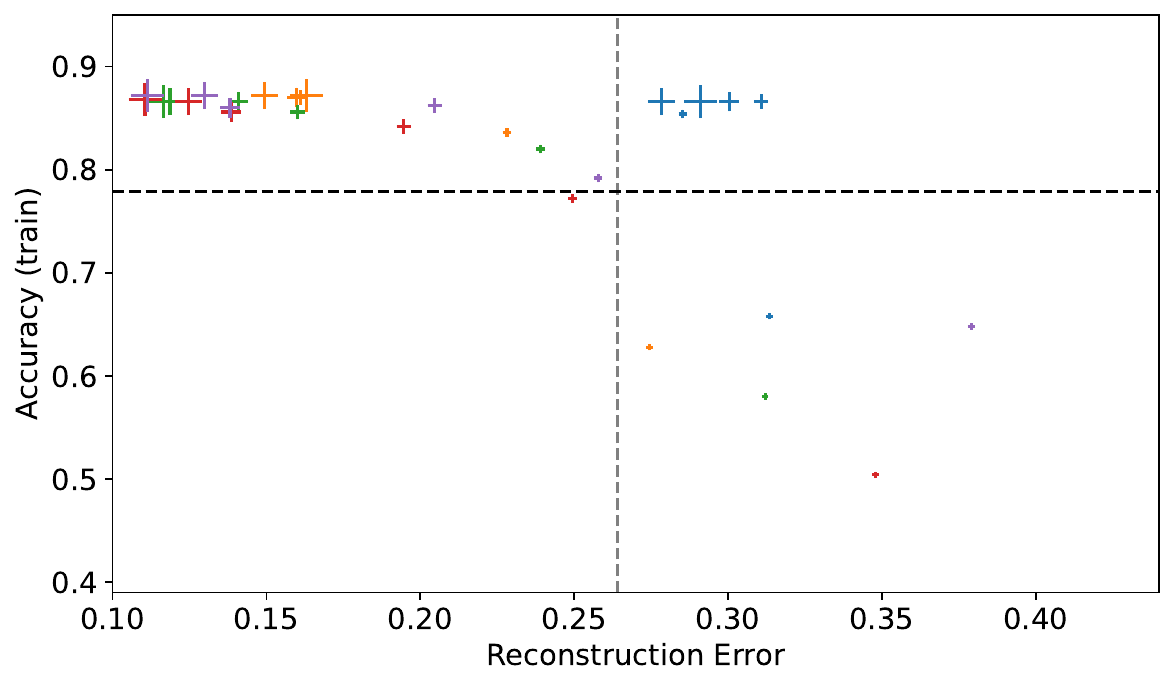}
    \caption{Default Credit dataset, depth 5 decision trees.}
    \label{fig:results_tradeoffs_acc_reconstr_default_credit_depth_5}
\end{subfigure}
\hfill
\begin{subfigure}{0.5\textwidth}
    \centering
    \includegraphics[width=1.0\linewidth]{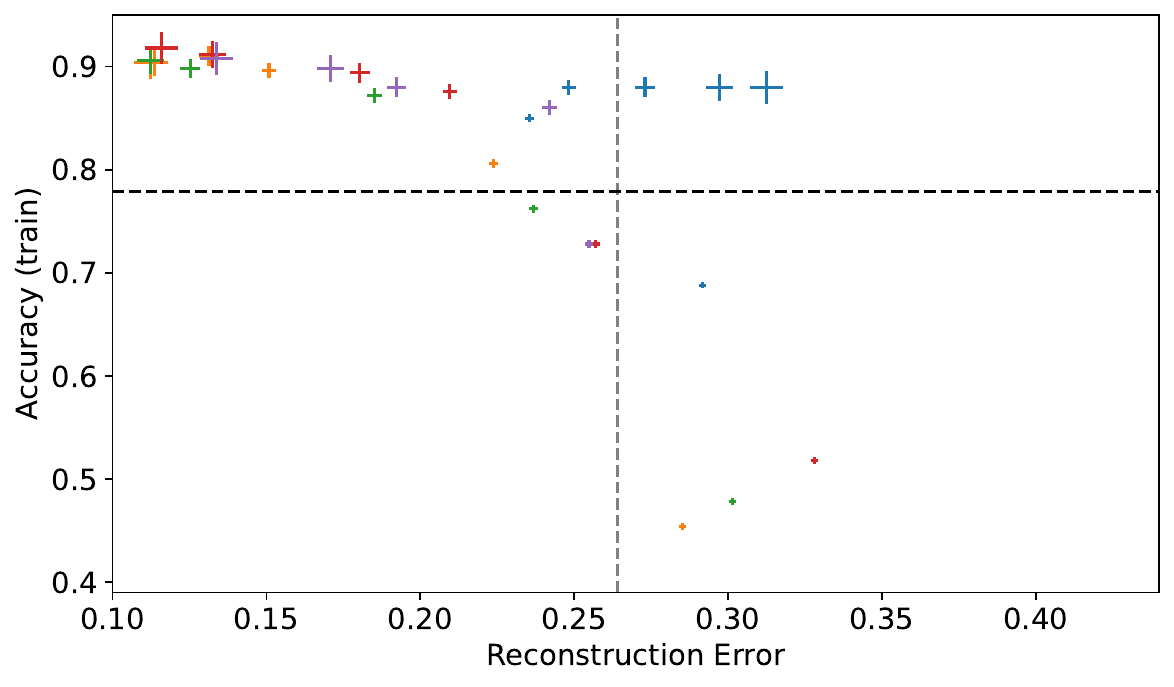}
    \caption{Default Credit dataset, depth 7 decision trees.}
    \label{fig:results_tradeoffs_acc_reconstr_default_credit_depth_7}
\end{subfigure}

\vspace{10pt}

\begin{subfigure}{\textwidth}
    \centering
    \includegraphics[width=0.6\linewidth]{Figures/legend_tradeoffs.pdf}
\end{subfigure}

\caption{Average training accuracy of $\varepsilon$-DP RFs as a function of the reconstruction error of our attack for different numbers of trees $\vert \forest \rvert$ and privacy budgets $\varepsilon$, for the considered datasets and tree depths values (continued).}\label{fig:results_all_tradeoffs_predictive_perf_vs_recontr_error_bis}
\end{figure*}

Figures~\ref{fig:results_all_tradeoffs_predictive_perf_vs_recontr_error} and \ref{fig:results_all_tradeoffs_predictive_perf_vs_recontr_error_bis} present the detailed trade-offs between the RFs' predictive performance and the empirical reconstruction error of our attack across all experiments. 
These figures confirm the key trends discussed in the main paper (in particular \textbf{Result~3}), and are consistent across all datasets, privacy budgets, tree depths and forest sizes. 
In particular, most RFs are located either in the top-left or bottom-right corners of the plots. 
The top-left region corresponds to RFs with non-trivial predictive performance, giving some advantage to the reconstruction attack compared to the random baseline, while the bottom-right region includes RFs that provide no significant advantage for the reconstruction attack but perform worse than a majority classifier.
\updated{Interestingly, the few RFs that exhibit non-trivial predictive performance while remaining resilient to reconstruction attacks (top-right corner) are also the sparsest, mostly consisting of a single tree ($\lvert \forest \rvert = 1$). 
This suggests that information from multiple trees can typically be aggregated, substantially enhancing reconstruction attack success.}%

\FloatBarrier

\section{Privacy Leak Analysis}\label{appendix:privacy_leak}

As discussed in Section~\ref{subsec:results}, our proposed reconstruction attack is often able to generate reconstructed datasets that are closer to the actual RF training set $\dataset$ than a random baseline uniformly sampling the attributes' values within their respective domains. 
While this suggests that the RFs provide information regarding $\dataset$ even with thorough DP guarantees, this information can either be distribution-wide or specific to $\dataset$. 
While the former case can still be problematic if the entity owning the data does not want to leak its distribution, it arguably does not correspond to a \emph{privacy leak}, which occurs if information specific to the actual RF training set $\dataset$ can be retrieved~\cite{DBLP:conf/innovations/CohenKMMNST25}. 
An appealing way to quantify the latter would be to compute the reconstruction error of a baseline attacker with full knowledge of the training data distribution, and then measure the improvement of our attack relative to this baseline. 
However, this methodology is inappropriate for two main reasons. 
First, our attack setup does not assume knowledge of the data distribution, and thus comparing it to a distribution-aware baseline (which relies on a different adversarial model) would therefore make little sense. 
Second, such a comparison implicitly assumes that our attack recovers the complete distributional information of the DP RF training set, with any additional reduction in reconstruction error interpreted as training set–specific information. 
In practice, this assumption does not hold: whenever our attack achieves lower reconstruction error than the random baseline (see Table~\ref{tab:error-values}), the information it extracts about the DP RF training set reflects a mixture of partial distributional knowledge and dataset-specific details.
In this appendix, we instead take a rigorous and systematic approach to quantifying privacy leakage, by investigating whether the reconstructed dataset is closer to $\dataset$ than to other datasets drawn from the same distribution in a statistically significant manner.

\updated{For each performed experiment, we compute the error (i) between the reconstructed dataset and the actual RF training dataset (as reported in Section~\ref{subsec:results}) and (ii) between the reconstructed dataset and datasets of the same size sampled from the same distribution. 
Intuitively, any decrease in the measurement (i) compared to (ii) is an indication that the attack is able to retrieve information specific to the particular individuals belonging to the RF training set, rather than solely general information about the data distribution. 
In such a case, a privacy leak can be identified, since, while an ML model may reasonably leak distributional information, it should not encode specific information about individual members of its training set.}

\begin{figure*}[h!]
  \centering
    \begin{subfigure}[t]{0.49\textwidth}
    \centering\includegraphics[width=\textwidth]{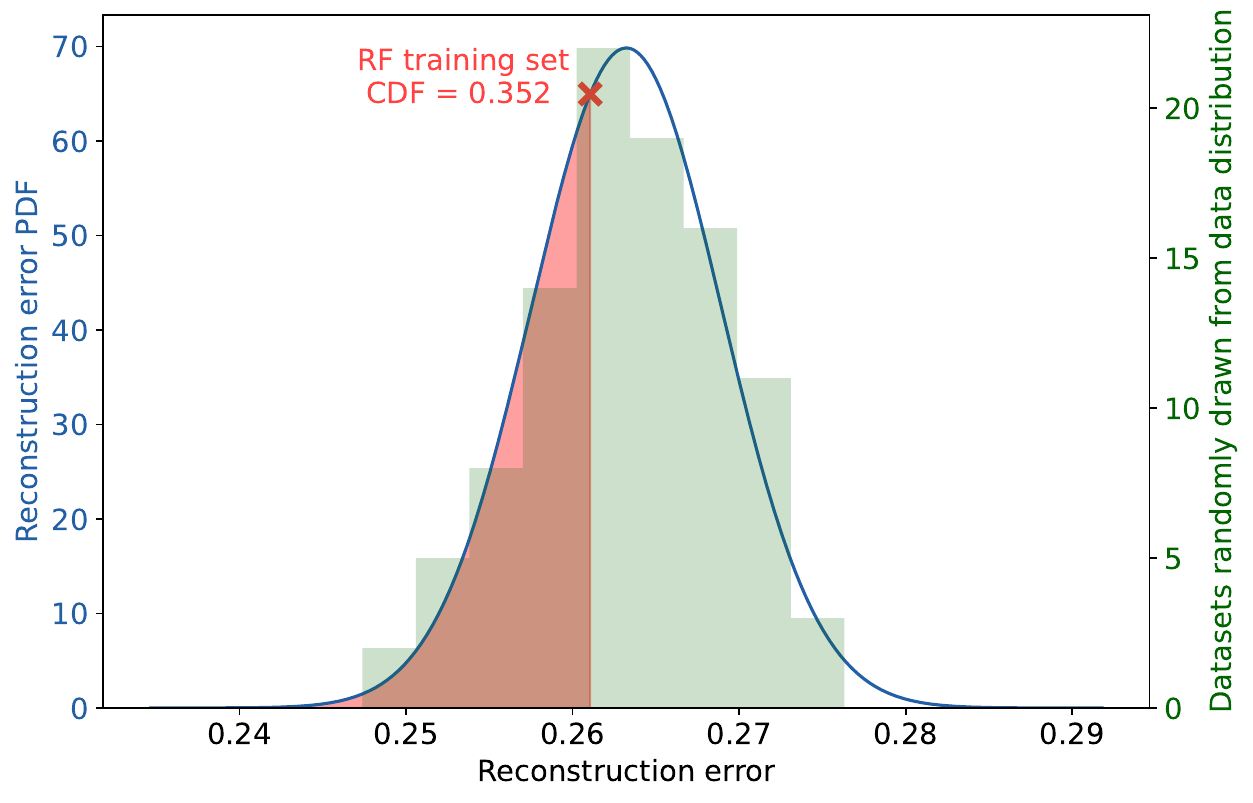}
    \caption{$\varepsilon=0.1$}\label{fig:results_privacy_leak_epsilon_0.1}
  \end{subfigure}
    \begin{subfigure}[t]{0.49\textwidth}
    \centering\includegraphics[width=\textwidth]{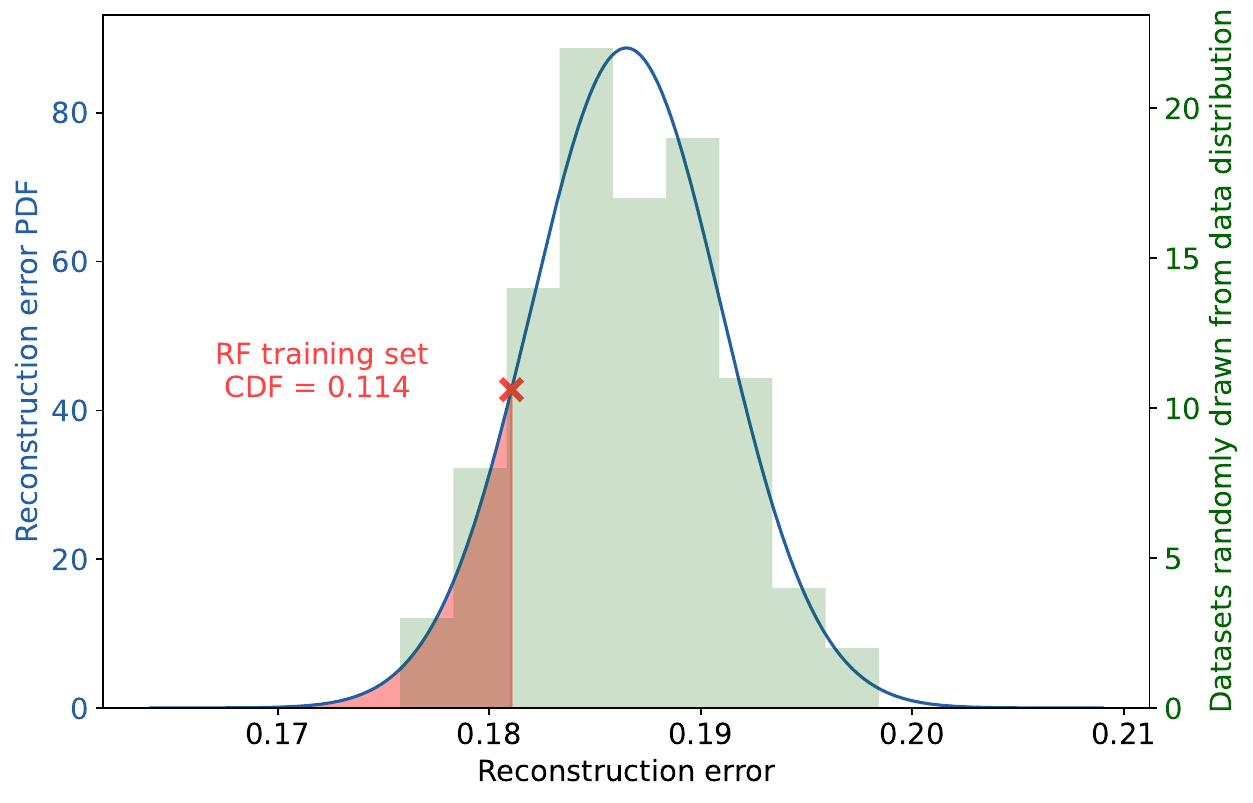}
    \caption{$\varepsilon=1$} 
    \label{fig:results_privacy_leak_epsilon_1}
  \end{subfigure}

  \vspace{10pt}
  
    \begin{subfigure}[t]{0.49\textwidth}
    \centering\includegraphics[width=\textwidth]{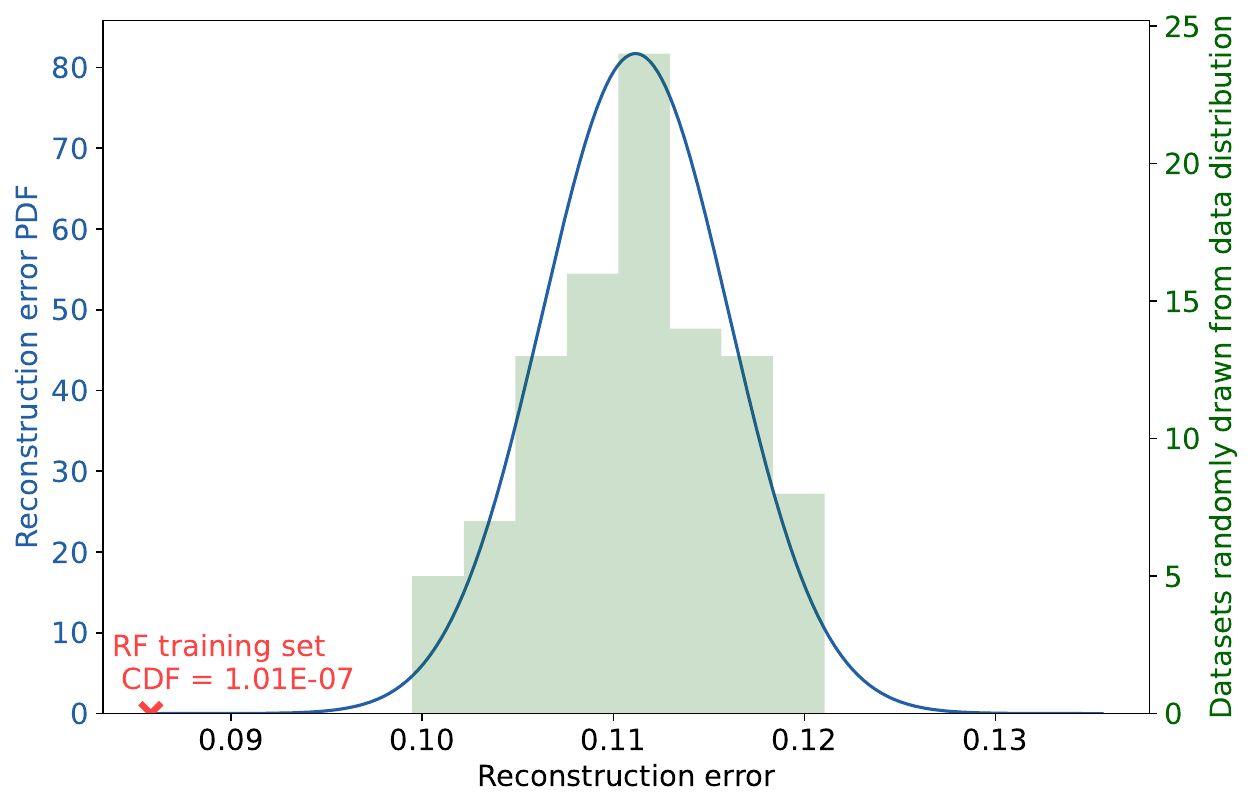}
    \caption{$\varepsilon=30$} 
    \label{fig:results_privacy_leak_epsilon_30}
  \end{subfigure}

\caption{\updated{Distribution of the errors measured between the reconstructed dataset and datasets randomly drawn from the data distribution. We report the actual proportions in green, and the fitted distribution (probability density function -- PDF) in blue. We then compute the cumulative density function (CDF) at the point corresponding to the reconstruction error measured using the actual RF training dataset. This value quantifies \emph{how likely it is to observe a reconstruction error lower or equal to the one measured on the actual RF training set, using a dataset randomly drawn from the same data distribution}. We display this process for the experiments on the UCI Adult Income dataset, using $\lvert \forest \rvert = 10$ trees of depth $5$, with three distinct DP budgets $\varepsilon$ (for a given random seed).}}\label{fig:results_privacy_leak}
\end{figure*}

\updated{In practice, we randomly sample $100$ datasets containing $\nexamples$ examples from the same distribution as $\dataset$, and compute the errors between each of them and the reconstructed dataset. 
We then fit a normal distribution from this list of errors, estimating how likely it is to measure a given reconstruction error for a dataset randomly drawn from the data distribution. 
We finally compute the cumulative density function (CDF) of this fitted distribution for the actual reconstruction error, which quantifies \emph{the likehood of observing a reconstruction error lower or equal to the one measured on the actual RF training set, using a dataset randomly drawn from the same data distribution}. 
Hence, it is somewhat analogous to a $p$-value, with very small values corresponding to observations that are unlikely to be observed purely by chance.
We illustrate this process on Figure~\ref{fig:results_privacy_leak} for a representative setup using DP RFs built of $\lvert \forest \rvert = 10$ trees of depth $5$, on the UCI Adult Income dataset. 
As it can be seen, the normal laws fit the distribution of the measured errors very well.
More precisely, we report results for three particular experiments, using very tight $\varepsilon=0.1$ (Figure~\ref{fig:results_privacy_leak_epsilon_0.1}), tight $\varepsilon=1$ (Figure~\ref{fig:results_privacy_leak_epsilon_1}) or large $\varepsilon=30$ (Figure~\ref{fig:results_privacy_leak_epsilon_30}) privacy budgets.}

\updated{In the first case ($\varepsilon=0.1$), the results previously presented in Table~\ref{tab:error-values} show that the reconstruction is not more accurate than the random baseline. 
Thus, no information is retrieved from the trained RF.
Unsurprisingly, the reconstructed dataset is as far from datasets randomly drawn from the data distribution as it is from the actual RF training dataset. 
However in the two other situations, we observed in Table~\ref{tab:error-values} a significant decrease of the reconstruction error compared to the random baseline: for $\varepsilon=1$ and $\varepsilon=30$, useful information is extracted from the trained RF. 
For $\varepsilon=1$, Figure~\ref{fig:results_privacy_leak_epsilon_1} suggests that this information is mainly distributional: the reconstructed dataset is not significantly closer to the actual RF training dataset than it is from other datasets from the same distribution. 
Formally, there is a non-negligible $11.4$\% chance that a dataset randomly drawn from the data distribution is closer to the reconstructed dataset than the actual RF training set.
In turn, for $\varepsilon=30$, the inferred information can be considered as a privacy leak.
Indeed, as evidenced in Figure~\ref{fig:results_privacy_leak_epsilon_30}, there is a negligible probability ($10^{-7}$) that the actual RF training set lies as close to the reconstructed dataset by chance (given the data distribution).}

\updated{We report the average results (estimated CDF) of all experiments in Table~\ref{tab:p-values} in Section~\ref{subsec:results}. 
They suggest that a privacy leak is \emph{likely} (CDF for the training set reconstruction error $\leq$ $5$\%) for most configurations using privacy budget $\varepsilon \geq 5$. Importantly, even for tighter privacy budgets (\emph{e.g.}, $\varepsilon=1$), the results for some configurations still suggest possible privacy leaks, highlighting the importance of the choice of hyperparameters and their influence over the privacy-accuracy trade-off.
It is worth noting that several configurations yield CDF values slightly above 5\% but remaining reasonably low -- which suggests that the reconstructed dataset is closer from the actual RF training set than from average datasets randomly sampled from the same distribution, potentially indicating a moderate privacy leak.}

\FloatBarrier
\section{Additional Experiments on Scalability}\label{appendix:additional_experiments_scalability}

\updated{In this section, we investigate the effect of the DP RFs training set size on the effectiveness of our dataset reconstruction attack. 
More precisely, we focus on DP RFs built of $\lvert \forest \rvert = 10$ trees of depth $5$, with DP budget $\varepsilon=5$. 
This configuration is representative of the experimental conditions presented in Section~\ref{subsec:setup}. 
We train such DP RFs with varying numbers of training examples $\nexamples \in \{25,  50, 100, 200, 300, 400, 500 \}$. 
The remaining of the setup is as described in Section~\ref{subsec:setup} (in particular, all results are averaged across $5$ different random seeds), apart from the maximum CPU time and the number of threads allocated to the solver, which are increased to $10$ hours and $32$ threads, respectively. This setup allows a thorough characterization of the solver anytime behaviour, even for the largest problems, as empirically evidenced in our results.}

Table~\ref{tab:scalability} displays the predictive accuracy of the built DP RFs, along with the success of the associated reconstruction attacks.
Note that the results for the configuration used in Section~\ref{subsec:results} (i.e., $\nexamples=100$) are slightly better here because the CP formulations are often not solved to optimality and the solver benefits the additional running time to find slightly better solutions. 
This suggests that the results presented throughout the paper are likely to improve with increased adversarial computational power and advancements in CP solver technologies. 
We observe that as the training set size $\nexamples$ increases, the test set accuracy of the built DP RFs improves significantly. 
Meanwhile, the training accuracy decreases and converges toward the test accuracy, indicating better generalization of the built DP RFs.
Another interesting phenomenon can be observed: the DP RFs training and test set accuracies sometimes increase simultaneously while increasing the training set size $\nexamples$ (\emph{e.g.}, from $\nexamples=300$ to $\nexamples=400$ for the Default Credit dataset). 
This is due to the fact that the relative impact of the added noise decreases when $\nexamples$ increases, allowing for better predictive performances. 

Crucially, this reduction in the relative effect of noise over the trees' parameters also benefits the reconstruction attack: the reconstruction error significantly decreases as the size of the DP RF's training set $\nexamples$ increases. 
For instance, while the reconstruction error from DP RFs trained on $\nexamples=100$ examples from the UCI Adult Income dataset is $12.3$\% on average, this number decreases to $8.5$\% (hence a relative decrease of $31$\%) when it is trained on $\nexamples=500$ examples, although the number of examples to retrieve is significantly greater. 
Indeed, as the training set size $N$ increases, the complexity of the underlying optimization problem increases as well, as the number of possible reconstructions increases exponentially with $N$. However, in the context of DP, considering larger values of $N$ also improves the signal-to-noise ratio. 
The leaves' counts also become more informative, as their support grows and allows making more overlaps with those of other trees. 
Technically speaking, since each modelled training example comes with a set of constraints, the number of such constraints increases accordingly: the search space is larger but also more constrained. This likely explains the better performance of our attack on larger training sets. 

\begin{table*}[htbp]
\centering
\caption{\updated{Results of our experiments on the reconstruction attack's scalability. For varying training set sizes $\nexamples$, we report the predictive performances of the built DP RFs, along with the resulting reconstruction error (averaged over $5$ different random seeds).}}
\label{tab:scalability}

\begin{subtable}{0.45\textwidth}
\centering
\caption{UCI Adult Income dataset}
\begin{tabular}{@{}ccccc@{}}
\toprule
\multirow{2}{*}{\#Examples} & \multicolumn{2}{c}{Reconstruction error} & \multicolumn{2}{c}{RF Accuracy} \\ \cmidrule(l){2-5} 
                            & Avg                 & Std                & Train          & Test           \\ \midrule
25                          & 0.172               & 0.008              & 0.888          & 0.766          \\ \midrule
50                          & 0.148               & 0.007              & 0.832          & 0.771          \\ \midrule
100                         & 0.123               & 0.010              & 0.824          & 0.766          \\ \midrule
200                         & 0.100               & 0.007              & 0.770          & 0.775          \\ \midrule
300                         & 0.095               & 0.008              & 0.775          & 0.770          \\ \midrule
400                         & 0.090               & 0.007              & 0.755          & 0.774          \\ \midrule
500                         & 0.085               & 0.002              & 0.758          & 0.773          \\ \bottomrule
\end{tabular}
\end{subtable}
\hfill
\begin{subtable}{0.45\textwidth}
\centering
\caption{COMPAS dataset}
\begin{tabular}{@{}ccccc@{}}
\toprule
\multirow{2}{*}{\#Examples} & \multicolumn{2}{c}{Reconstruction error} & \multicolumn{2}{c}{RF Accuracy} \\ \cmidrule(l){2-5} 
                            & Avg                 & Std                & Train          & Test           \\ \midrule
25                          & 0.111               & 0.018              & 0.712          & 0.533          \\ \midrule
50                          & 0.078               & 0.011              & 0.696          & 0.564          \\ \midrule
100                         & 0.057               & 0.004              & 0.720          & 0.635          \\ \midrule
200                         & 0.049               & 0.002              & 0.695          & 0.647          \\ \midrule
300                         & 0.039               & 0.005              & 0.693          & 0.652          \\ \midrule
400                         & 0.037               & 0.004              & 0.706          & 0.652          \\ \midrule
500                         & 0.039               & 0.004              & 0.692          & 0.655          \\ \bottomrule
\end{tabular}
\end{subtable}

\vspace{10pt}

\begin{subtable}{0.45\textwidth}
\centering
\caption{Default of Credit Card Clients dataset}
\begin{tabular}{@{}ccccc@{}}
\toprule
\multirow{2}{*}{\#Examples} & \multicolumn{2}{c}{Reconstruction error} & \multicolumn{2}{c}{RF Accuracy} \\ \cmidrule(l){2-5} 
                            & Avg                 & Std                & Train          & Test           \\ \midrule
25                          & 0.212               & 0.023              & 0.920          & 0.772          \\ \midrule
50                          & 0.174               & 0.010              & 0.876          & 0.780          \\ \midrule
100                         & 0.152               & 0.013              & 0.856          & 0.782          \\ \midrule
200                         & 0.130               & 0.009              & 0.828          & 0.782          \\ \midrule
300                         & 0.124               & 0.006              & 0.827          & 0.782          \\ \midrule
400                         & 0.117               & 0.006              & 0.828          & 0.786          \\ \midrule
500                         & 0.112               & 0.006              & 0.814          & 0.789          \\ \bottomrule
\end{tabular}
\end{subtable}

\end{table*}

\begin{figure}[h!]
  \centering

    \begin{subfigure}[t]{0.5\textwidth}
    \centering
    \includegraphics[width=\linewidth]{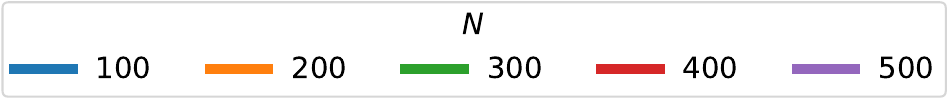}
\end{subfigure}

    \vspace{10pt}
    
    \begin{subfigure}[t]{0.48\textwidth}
    \centering\includegraphics[width=\textwidth]{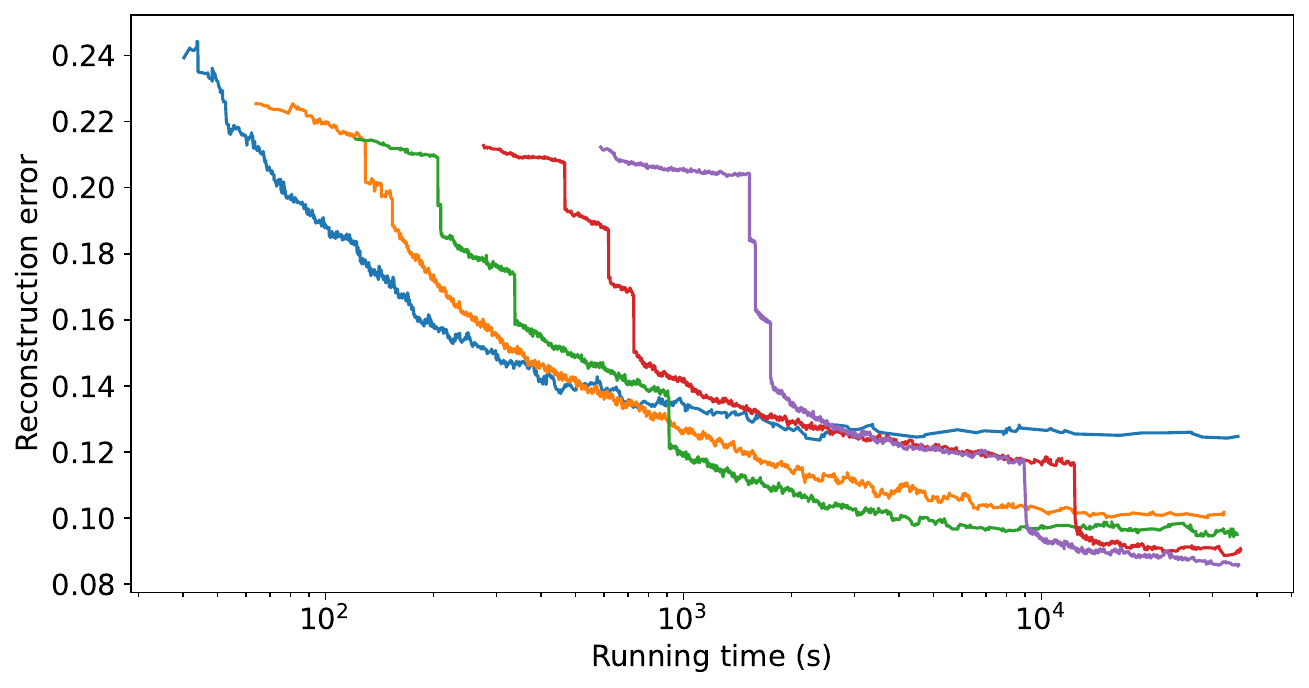}
    \caption{UCI Adult Income dataset}\label{fig:results_scalability_anytime_adult}
  \end{subfigure}  
    \begin{subfigure}[t]{0.48\textwidth}
    \centering\includegraphics[width=\textwidth]{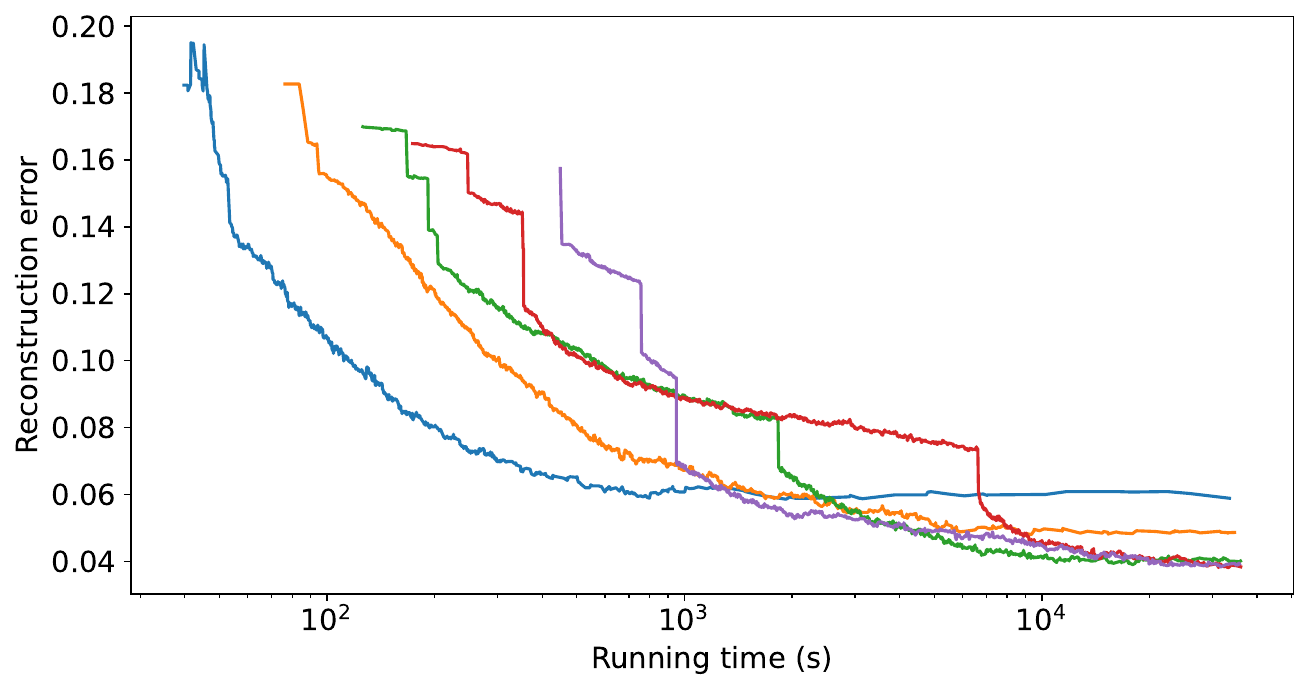}
    \caption{COMPAS dataset} 
    \label{fig:results_scalability_anytime_compas}
  \end{subfigure}
  
    \vspace{10pt}
  
    \begin{subfigure}[t]{0.48\textwidth}
    \centering\includegraphics[width=\textwidth]{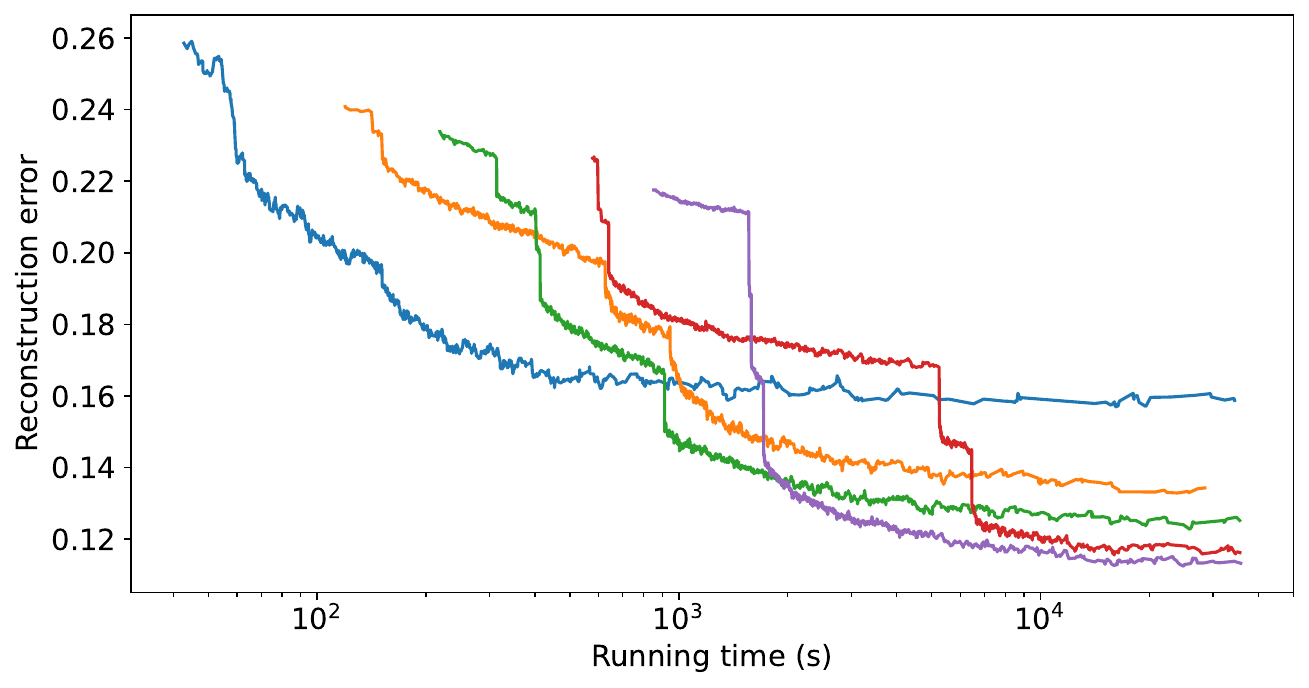}
    \caption{Default of Credit Card Clients dataset}\label{fig:results_scalability_anytime_default_credit}
  \end{subfigure}

\caption{\updated{Anytime reconstruction error for the experiments on our attack's scalability. We report the reconstruction error as a function of the running time, for varying training set sizes $\nexamples$. For these experiments, all forests are built of $\lvert \forest \rvert = 10$ trees of depth $5$, with DP budget $\varepsilon=5$. Reconstruction errors are averaged over 5 different random seeds.}}\label{fig:results_scalability_anytime}
\end{figure}

Similar trends can be seen in Figure~\ref{fig:results_scalability_anytime}, which displays the anytime performances of our attack. More precisely, we report the reconstruction error value as a function of the solver running time. 
Note that, at any given time, if some runs (\emph{i.e.}, certain random seed values out of the five considered) have not yet produced a feasible reconstruction, their reconstruction error is replaced by that of the random baseline. 
This substitution explains the ``stairs'' effect observed in some plots.
As running time increases, the CP solver progressively finds reconstructions with better objective function values (as defined in~\eqref{eq:obj_f}), which in most cases also correspond to smaller reconstruction errors. 
This illustrates the alignment between our objective (maximizing the log-likelihood of the inferred DP noise values) and the reconstruction error. 
The effect of the training set size $\nexamples$ is also apparent: reconstructing larger training sets is computationally more challenging, and feasible reconstructions are found more slowly as $\nexamples$ increases. 
Nevertheless, reconstruction errors ultimately tend to converge to better values for larger $\nexamples$ due to a more favourable signal-to-noise ratio. 

\updated{Overall, these results suggest that the attack's performance is likely to improve as the size of the target DP RF training set increases, while also demonstrating the scalability of the proposed reconstruction approach.}

{\rebuttalsatmlmultiline

Finally, Figure~\ref{fig:results_scalability_tradeoffs} displays the trade-offs between the DP RF training accuracy, test accuracy, and the reconstruction attack performance for various training set sizes. As shown most clearly for the Default of Credit Card Clients dataset in Figure~\ref{fig:results_scalability_tradeoffs_default_credit}, increasing the number of training examples $\nexamples$ simultaneously (i) decreases the training accuracy, (ii) increases the test accuracy, which indicates improved generalization of the DP RF, and (iii) reduces the reconstruction error achieved by our attack (thereby improving its success).
Larger values of $\nexamples$ also slightly improve the random baseline, since a larger candidate set makes the alignment step easier and increases the likelihood of partial matches.
However, the decrease in reconstruction error obtained by our attack is consistently larger than the decrease observed for the random baseline. This shows that our attack is genuinely exploiting the additional information provided by larger training sets and is not simply benefiting from the alignment artifact.
As discussed previously, a likely explanation is that larger training sets lead to higher leaf counts for similar noise magnitudes, which improves the signal-to-noise ratio within the trees. These results illustrate that this stronger signal benefits both the DP RF accuracy and the reconstruction attack.

\begin{figure}[h!]
  \centering

\begin{subfigure}[t]{0.7\textwidth}
    \centering
    \includegraphics[width=\linewidth]{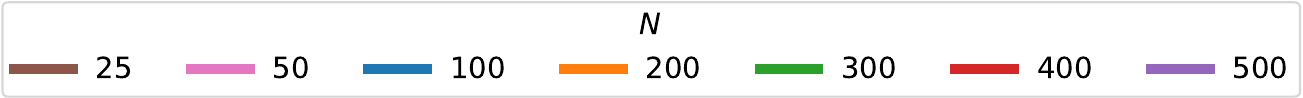}
    \includegraphics[width=0.80\linewidth]{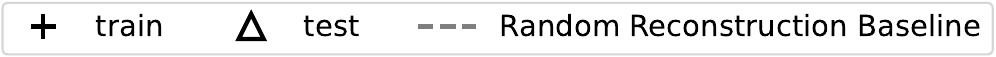}
\end{subfigure}

    \vspace{10pt}
    
    \begin{subfigure}[t]{0.48\textwidth}
    \centering\includegraphics[width=\textwidth]{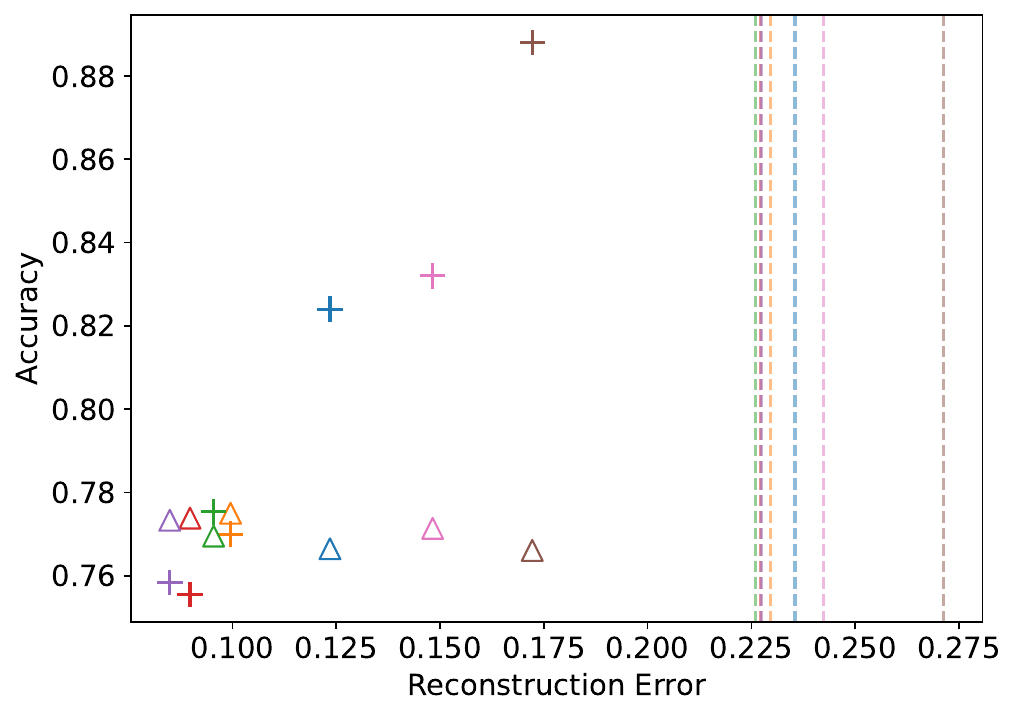}
    \caption{UCI Adult Income dataset}\label{fig:results_scalability_tradeoffs_adult}
  \end{subfigure}  
    \begin{subfigure}[t]{0.48\textwidth}
    \centering\includegraphics[width=\textwidth]{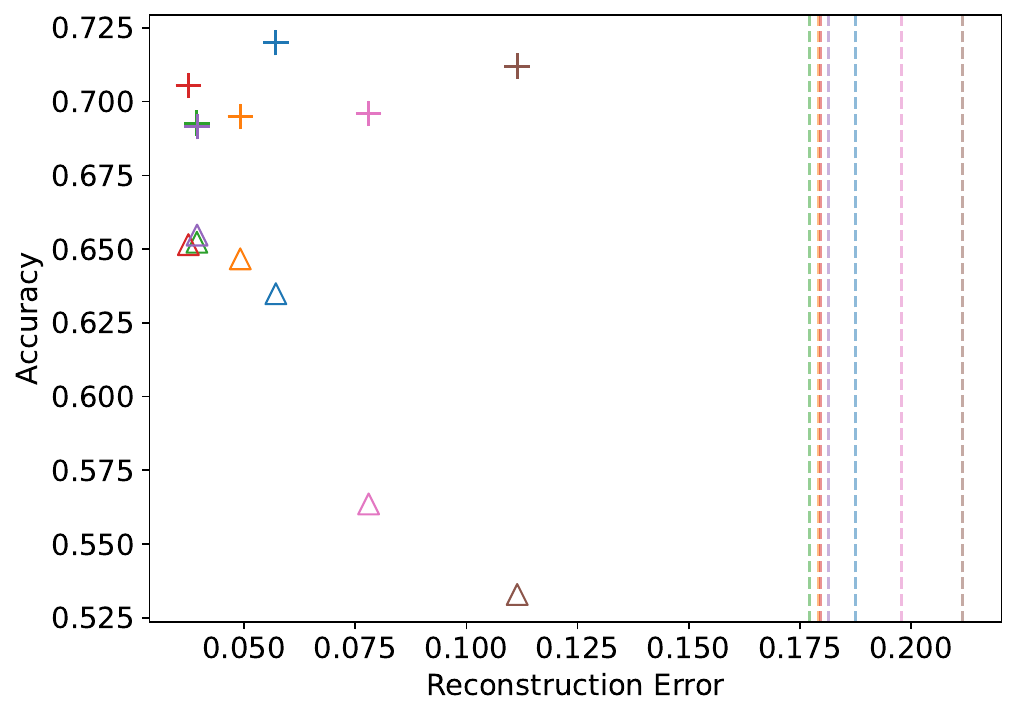}
    \caption{COMPAS dataset} 
    \label{fig:results_scalability_tradeoffs_compas}
  \end{subfigure}
  
    \vspace{10pt}
  
    \begin{subfigure}[t]{0.48\textwidth}
    \centering\includegraphics[width=\textwidth]{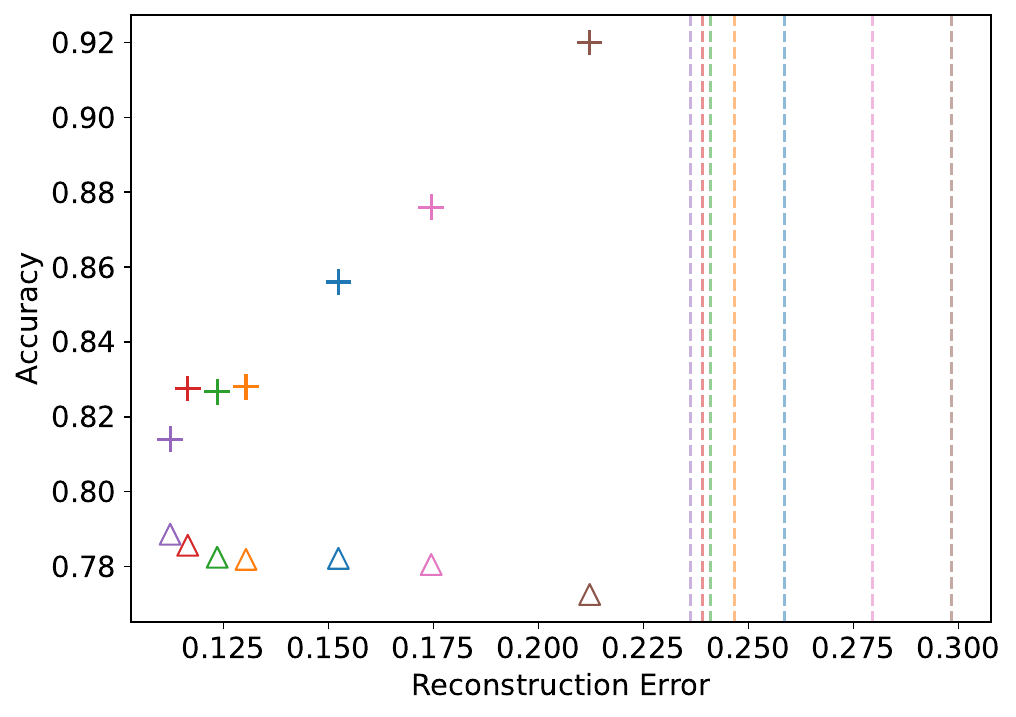}
    \caption{Default of Credit Card Clients dataset}\label{fig:results_scalability_tradeoffs_default_credit}
  \end{subfigure}

\caption{\rebuttalsatml{Average training and test accuracy of $\varepsilon$-DP RFs as a function of the reconstruction error of our attack for varying training set sizes $\nexamples$. For these experiments, all forests are built of $\lvert \forest \rvert = 10$ trees of depth $5$, with DP budget $\varepsilon=5$. All values are averaged over 5 different random seeds.}}\label{fig:results_scalability_tradeoffs}
\end{figure}

}

\FloatBarrier

\section{Additional Experiments on Partial Reconstruction}\label{appendix:additional_experiments_partial_reconstruction}

\updated{The purpose of this section is to determine whether (and to what extent) knowledge of part of the training set attributes (for all examples) helps reconstruct the others. 
As in the previous section, we focus on a representative setup with DP RFs built of $\lvert \forest \rvert = 10$ trees of depth $5$, with global DP budget $\varepsilon=5$. 
For each dataset, we vary the number of known attributes between 0 and $\nattributes_o-1$, in which $\nattributes_o$ is the number of original (\emph{i.e.}, before one-hot encoding) features. 
Indeed, binary attributes one-hot encoding the same original feature are not independent from each other and considering them separately in this partial attributes knowledge setup would bias the reconstruction results. 
The number of such original features for the COMPAS (respectively, UCI Adult Income and Default Credit) dataset is $7$ (respectively, $14$ and $16$). For each of the $5$ random seeds, we randomly sample the $m \in \{0 ,\dots, \nattributes_o-1\}$ known attributes. 
We then modify the CP formulation to fix the known attributes of each example to their true value, and solve the resulting CP model to retrieve the unknown ones.}

\begin{figure*}[h!]
  \centering
    \begin{subfigure}[t]{0.48\textwidth}
    \centering\includegraphics[width=\textwidth]{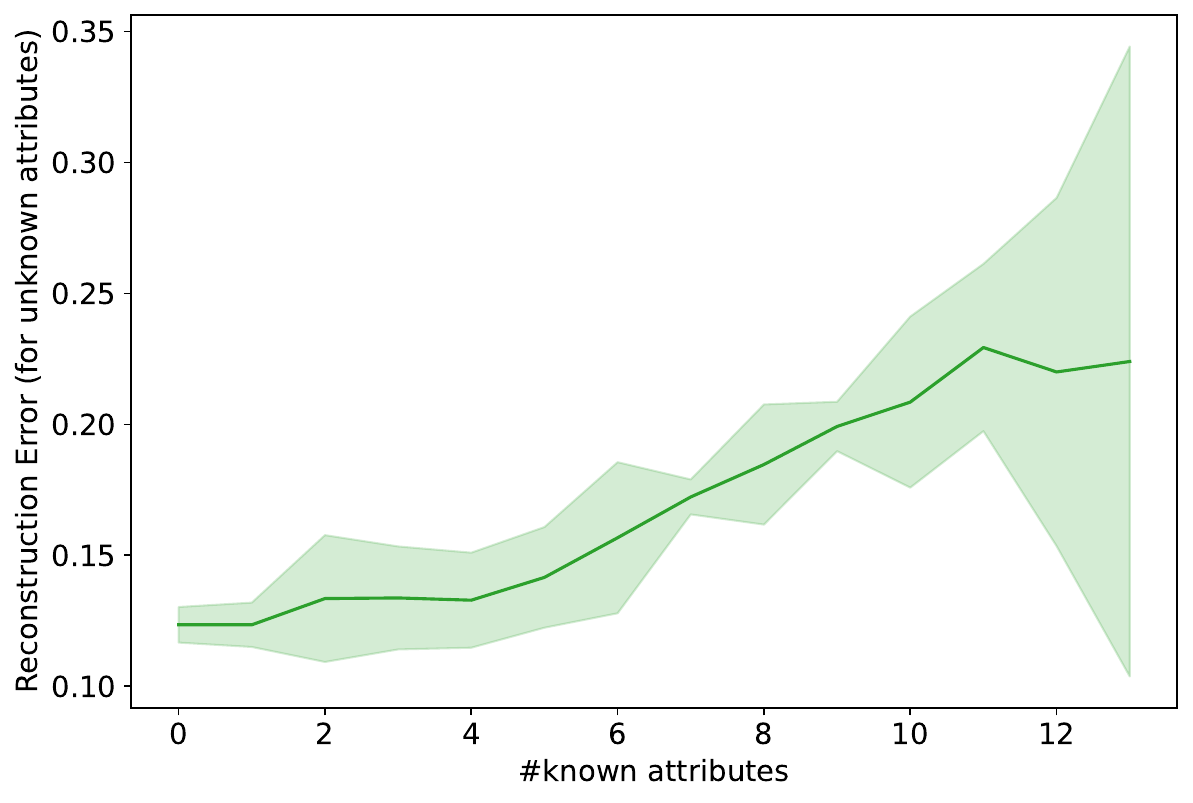}
    \caption{UCI Adult Income dataset}\label{fig:results_partial_reconstr_attrs_adult}
  \end{subfigure}
    \begin{subfigure}[t]{0.48\textwidth}
    \centering\includegraphics[width=\textwidth]{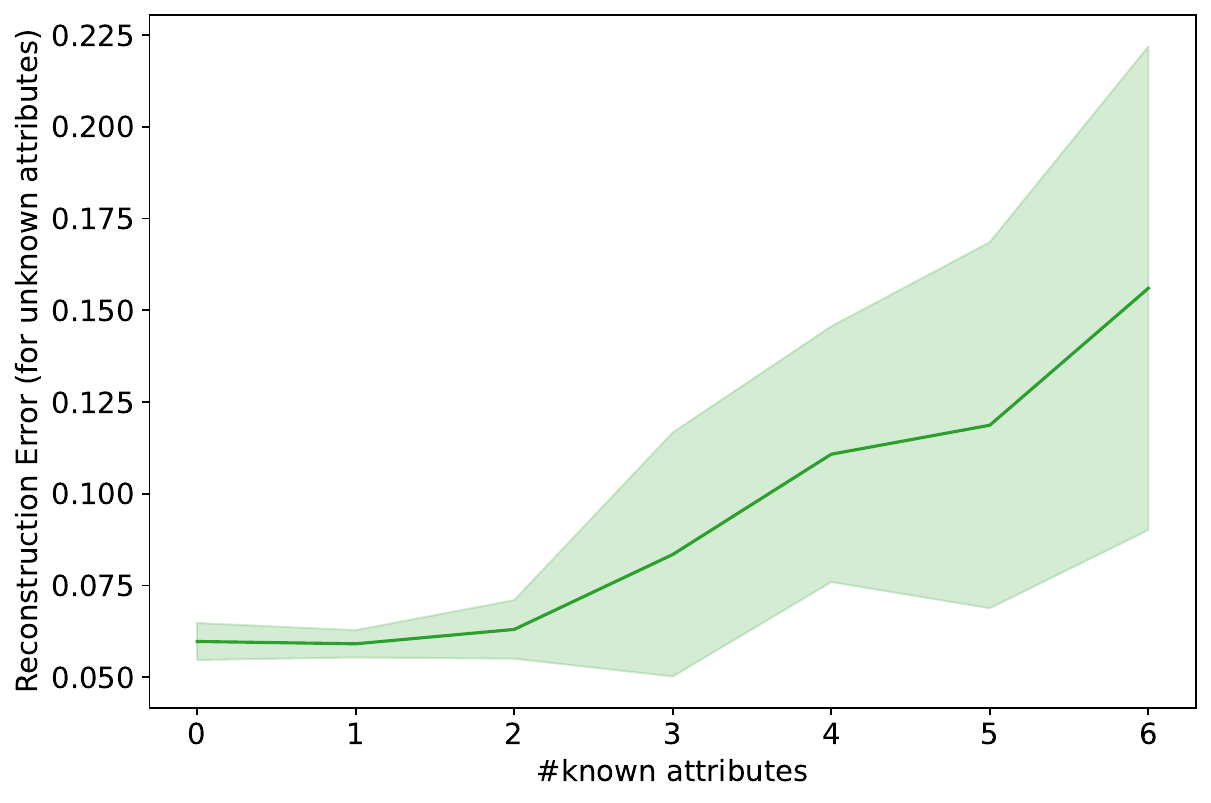}
    \caption{COMPAS dataset} 
    \label{fig:results_partial_reconstr_attrs_compas}
  \end{subfigure}
  
    \vspace{10pt}
  
    \begin{subfigure}[t]{0.48\textwidth}
    \centering\includegraphics[width=\textwidth]{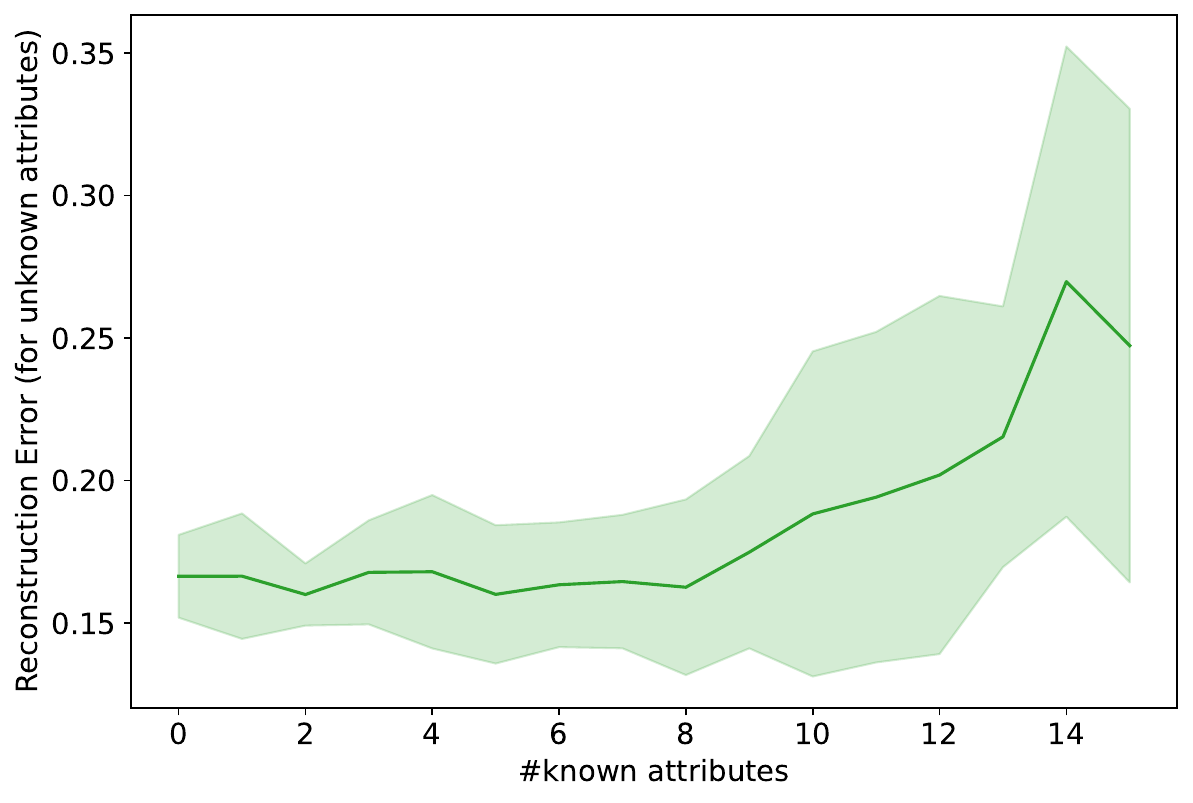}
    \caption{Default of Credit Card Clients dataset}\label{fig:results_partial_reconstr_attrs_default_credit}
  \end{subfigure}

\caption{\updated{Results of reconstruction experiments with knowledge of some of the attributes. We report the reconstruction error (for the unknown attributes) as a function of the number of known attributes in the forest's training set. For these experiments, all forests are built of $\lvert \forest \rvert = 10$ trees of depth $5$, with DP budget $\varepsilon=5$. Reconstruction errors are averaged over 5 different random seeds and we also report the standard deviation.}}\label{fig:results_partial_reconstr_attrs}
\end{figure*}

\updated{We report in Figure~\ref{fig:results_partial_reconstr_attrs} the reconstruction error (measured for the unknown attributes) as a function of the number of known attributes. 
The results for all three datasets consistently show that the knowledge of a number of attributes does not help reconstruct the others. 
The slight increase in reconstruction error when the number of known attributes increases can be explained by the alignment process: as explained in Section~\ref{subsec:setup}, to figure out which reconstructed example corresponds to which original one, we first perform a minimum cost matching, before computing the average reconstruction error for each pair of original-reconstructed examples. 
In this partial reconstruction setup, this matching is still performed on the complete examples' features vector even though the attack only retrieves part of their attributes. 
Hence, the improvement in reconstruction error induced by the matching process reduces as the number of unknown attributes decreases.}

\updated{This observation empirically verifies the theoretical guarantees offered by DP: the protection enforced for the unknown attributes can not be ``undone'' through postprocessing the DP RF, even with the auxiliary information provided by the known attributes. 
In other words, in our considered setup, and for our proposed attack, DP prevents so-called \emph{linkage attacks}~\cite{dwork2014algorithmic}. 
It is worth noting that these results contrast with the ones observed against non-DP RFs, for which knowledge of some of the attributes greatly helped reconstruct the others~\cite{ferry2024trained}.}

\FloatBarrier

{
\rebuttalsatmlmultiline
\section{Additional Experiments on Reconstruction in an Informed Adversary Setting}\label{appendix:informed_adversary}

As detailed throughout the paper, our reconstruction attack is designed to recover the entire training set of a DP-protected RF, with or without prior knowledge of its size $\nexamples$. To the best of our knowledge, no existing work has targeted full training-set reconstruction for DP-protected models. As explained in Section~\ref{sec:related_work}, a related line of work instead considers a simplified informed adversary setting, where the attacker knows $\nexamples - 1$ training examples and attempts to reconstruct the remaining one by using both the known examples and the model parameters. This setting is interesting because it focuses on reconstructing a single individual, which is closely aligned with the original protection goal of DP, and it provides a meaningful way to compare our attack with existing approaches.

In this appendix, we show that our attack can be adapted to handle this particular setup, and that it outperforms the state-of-the-art.

\vspace{0.25\baselineskip}

\textbf{Adapted CP formulation.} Two main modifications to the CP formulation provided within Section~\ref{sec:attack} can be done to adapt our attack to the informed adversary setup. 

First, since $\nexamples - 1$ examples are known by the attacker, we simply set their attributes and classes to their true value:
\begin{align}
     &\varz[\example,\class]=1 &\example \in \{1,\dots,\nexamples-1\}, \quad \class \in \classes, \quad \class = \class_{\example}\label{constr:informed_class_1}\\
     &\varz[\example,\class]=0 &\example \in \{1,\dots,\nexamples-1\}, \quad \class \in \classes, \quad \class \neq \class_{\example}\label{constr:informed_class_2}\\
     &\{\varx[k,i]\}_{\feature \in \{1, \dots, \nattributes\}}
     = \attributes[\example] &\example \in \{1,\dots,\nexamples-1\}. \label{constr:informed_attributes}
\end{align}
More precisely, for each known example $\example \in \{1,\dots,\nexamples-1\}$, Constraints~\eqref{constr:informed_class_1} and~\eqref{constr:informed_class_2} fix the one-hot variables $\varz[\example,\class]$ to the true class label $\class_{\example}$, while Constraint~\eqref{constr:informed_attributes} fixes its reconstructed attribute vector $\{\varx[k,i]\}_{\feature \in \{1, \dots, \nattributes\}}$ to the true attribute values $\attributes[\example]$.

Second, the knowledge of $\nexamples - 1$  training examples provides additional information about the underlying data distribution, which was not available in the full-dataset reconstruction setting.
This information can be leveraged to encourage the reconstructed example to be not only compatible with the DP RF counts and structure, but also consistent with the empirical distribution of the known examples. To incorporate this constraint, we add a regularization term to the objective function that penalizes the average
$\ell_1$ distance (which corresponds to the Manhattan distance in our binary setting) between the reconstructed example and the known training examples.
The resulting objective function is:
\begin{align}
    \max  \left(- \sum\limits_{\example \in \{1,\dots,\nexamples-1\}}{\lVert \{\varx[\nexamples,i]\}_{\feature \in \{1, \dots, \nattributes\}} - \attributes[\example] \rVert_{1}}\right) + \alpha \left( \sum\limits_{\tree \in \forest}\sum\limits_{\node \in \leaves[\tree]} \sum\limits_{\class \in \classes} \sum\limits_{\noiseval = -\gamma}^{\gamma} \log(p_\noiseval) \mathds{1}_{\Delta_{\tree\node\class} = \noiseval}\right)
    \label{eq:obj_f_informed_adversary}
\end{align}
where the first term acts as a regularizer that favors reconstructions lying close to the support of the empirical data distribution, and the second term is the original log-likelihood objective from \eqref{eq:obj_f}.
From a Bayesian perspective, the regularization term that penalizes the Manhattan distance to the known training examples can be interpreted as introducing a data-dependent log prior on the unknown example.
In practice, we rescale the coefficients inside the summations so that both terms have comparable magnitudes, ensuring that their relative importance is controlled solely by the parameter $\alpha$. In our preliminary experiments, we found out that $\alpha = 20 \varepsilon$ typically leads to good reconstruction results, and we retain this value throughout our experiments. For small privacy budgets $\varepsilon$, this choice places greater emphasis on producing a reconstruction that aligns with the empirical data distribution, whereas larger values of 
$\varepsilon$ increase the influence of the DP RF structure.

\vspace{0.25\baselineskip}

\textbf{Informed adversary baseline (adapted from~\cite{balle2022reconstructing}).}
The reconstruction attack studied in~\cite{balle2022reconstructing} considers an informed adversary who knows all training examples except one and has white-box access to the model parameters. Their closed-form attacks for convex models (such as generalized linear models) do not directly transfer to tree ensembles, but their reconstructor-network (RecoNN) strategy for general models can be adapted to DP RFs. Since their original work focuses on image classifiers, and does not consider tabular data or tree-based models, our adaptation to DP RFs required specific design and careful tuning.

In a nutshell, their attack trains a reconstructor network that learns a mapping from model parameters to the unknown training example. To construct training data for the RecoNN, the adversary uses the $\nexamples - 1$ known training examples (the fixed dataset), together with a set of $\nexamples'$ \emph{shadow examples}. For each shadow example, they train a shadow model on the union of the fixed dataset and that shadow example. These shadow models are then serialized into one-dimensional parameter vectors that can be fed to the RecoNN. This procedure yields $\nexamples'$ pairs of (serialized shadow model, shadow example), which are used to train the RecoNN. Finally, at attack time, the RecoNN receives the serialized parameters of the released model under attack and outputs a prediction for the unknown training example.

Serializing DP RFs into one-dimensional vectors for input to a RecoNN is a key aspect of the adaptation. A natural first approach is to flatten each tree in level-order (that is, using a breadth-first traversal from the root, visiting nodes level by level from left to right). For each internal node, we would store its split attribute and threshold, and for each leaf, its per-class noisy counts. However, this representation performed poorly in practice, even with extensive per-component normalization. The difficulty is that two DP RFs trained on slightly different datasets typically use different split attributes and thresholds at corresponding nodes, so examples traverse the trees differently and land in different leaves. As a result, without specialized architecture, the RecoNN struggled to learn meaningful correspondences between the serialized internal-node parameters and the structure of the underlying trees.

To address this issue, we adopted a simpler serialization scheme. During training, we fix the split attributes and thresholds to those of the observed DP RF, and only serialize the one-dimensional vector of per-class per-leaf noisy counts. This greatly reduces the complexity of the input representation, allowing the RecoNN to focus on learning a mapping from leaf-level signals to the unknown example. We also found that subtracting the contribution of the $\nexamples - 1$ known examples from these counts improved training stability, and we apply this correction in all experiments. Fixing the tree structure in this way enables the RecoNN to reliably associate positions in the leaf-count vector with the conditions that define each leaf and to learn the corresponding relationship to the unknown training example.

Since the information from the $\nexamples - 1$ known examples is already embedded in the serialization procedure, we found that training a single RecoNN on a larger auxiliary dataset (disjoint from the training set but drawn from the same distribution) and reusing it to reconstruct each unknown example both improves reconstruction quality and significantly reduces computational cost. In practice, we use the examples that were not selected in the DP RF training set as this auxiliary dataset.

In our experiments, the RecoNN was trained on $1{,}600$ pairs of (serialized shadow DP RF, shadow example), using batches of size $32$ for $5$ epochs. The implementation relies on the \texttt{PyTorch} library~\cite{paszke2019pytorch}, with a \texttt{BCEWithLogitsLoss} and an AdamW optimizer (learning rate $10^{-3}$, weight decay $10^{-4}$).
The RecoNN architecture is a feed-forward neural network with three hidden layers of sizes $1024$, $512$, and $256$. Each hidden layer applies a linear transformation, followed by a ReLU activation and a LayerNorm operation. The final layer is a linear projection that outputs logits for the reconstructed example. 
To extract the reconstructed example from the RecoNN, we pass the serialized DP RF through the network to obtain a vector of logits. These logits are then transformed into probabilities using a sigmoid activation. Since all attributes are binary, the final reconstructed example is obtained by thresholding each probability at $0.5$.

\vspace{0.25\baselineskip}

\textbf{Simple per-coordinate majority baseline.} For very small DP budgets $\varepsilon$ (in particular for $\varepsilon \leq 1$), the noise added to the leaves’ counts dominates the signal, and only limited information can be extracted from them (as was rigorously assessed in Appendix~\ref{appendix:privacy_leak} in the full reconstruction setup). In this regime, both our approach and the baseline from \cite{balle2022reconstructing} frequently output nearly constant examples that reflect the global data distribution rather than the specific DP RF parameters. To characterize this extreme case more precisely, we introduce a simple baseline that outputs the best constant reconstruction possible with respect to our evaluation metric. More precisely, this baseline selects the example $\mathbf{x}$ that minimizes the average Manhattan distance to all $\nexamples$ training examples, which is equivalent to choosing, for each coordinate, the empirical marginal majority among the training set. Since our reconstruction error is also measured using Manhattan distance, this baseline provides a lower bound on the performance of any attack that outputs a constant reconstruction.

\vspace{0.25\baselineskip}

\textbf{Experiments.} We focused on a representative DP RF configuration with $\lvert \forest \rvert = 10$ trees of depth $5$. For each of the three datasets, we varied the DP budget $\varepsilon \in \{0.1, 1, 5, 10, 20, 30, 10^3\}$ and ran each configuration with $10$ random seeds. We included the value $\varepsilon = 10^3$ because, in~\cite{balle2022reconstructing}, the attack required very large privacy budgets to succeed. This setting also serves as a proxy for evaluating the attacks in a regime where almost no noise is added to the DP RF parameters.

For each experiment, we trained a DP RF on $\nexamples$ training examples, with $\nexamples = 2{,}000$ (respectively, $\nexamples = 10{,}000$ and $\nexamples = 20{,}000$) for the COMPAS (respectively, Default of Credit Card Clients and UCI Adult Income) dataset. Indeed, in this setting, the computational cost of our attack is very small, enabling it to handle DP RFs trained on large datasets. However, at least $\nexamples$ examples (disjoint from the training set) must be reserved to form the RecoNN training set required by the implementation of the baseline of~\cite{balle2022reconstructing}. For each run, we randomly sample $100$ training examples and iteratively attempt to reconstruct each of them, using the remaining $\nexamples - 1$ training examples as known. We imposed a two-hour timeout per experiment (covering the reconstruction of all 100 examples), although in practice all methods completed well within this limit. %

\vspace{0.25\baselineskip}

\textbf{Results.} The results presented in Figure~\ref{fig:results_informed_reconstr} show that our attack consistently outperforms the baseline from~\cite{balle2022reconstructing}. For all privacy budgets $\varepsilon$, it systematically recovers more information about the missing example, as evidenced by lower reconstruction errors. This can be explained by the fact that our attack is tailored to exploit the full structure of a DP RF, whereas the baseline is model-agnostic (even though we made a significant effort to design a serialization and RecoNN training procedure that is effective in this setting).

\begin{figure*}[h!]
  \centering
  
\begin{subfigure}[t]{0.45\textwidth}
    \centering
    \includegraphics[width=\linewidth]{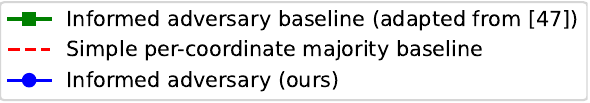}
\end{subfigure}

    \vspace{10pt}
    
    \begin{subfigure}[t]{0.48\textwidth}
    \centering\includegraphics[width=\textwidth]{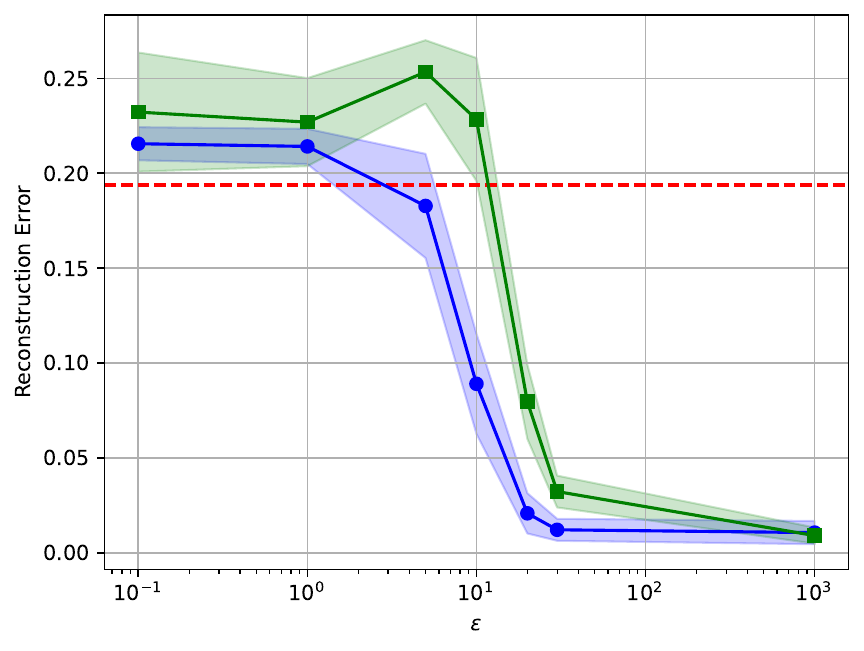}
    \caption{UCI Adult Income dataset}\label{fig:results_informed_reconstr_adult}
  \end{subfigure}
    \begin{subfigure}[t]{0.48\textwidth}
    \centering\includegraphics[width=\textwidth]{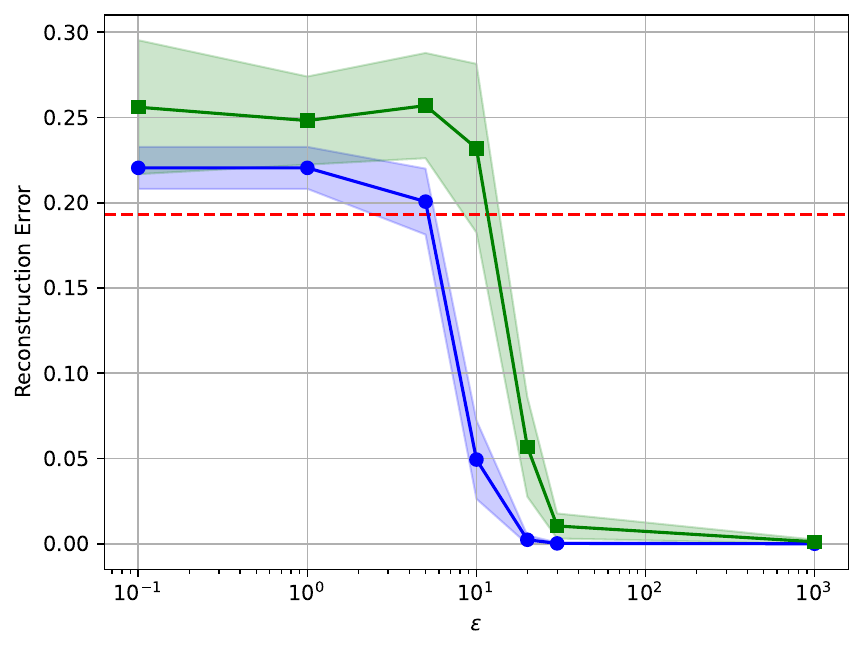}
    \caption{COMPAS dataset} 
    \label{fig:results_informed_reconstr_compas}
  \end{subfigure}
  
    \vspace{10pt}
  
    \begin{subfigure}[t]{0.48\textwidth}
    \centering\includegraphics[width=\textwidth]{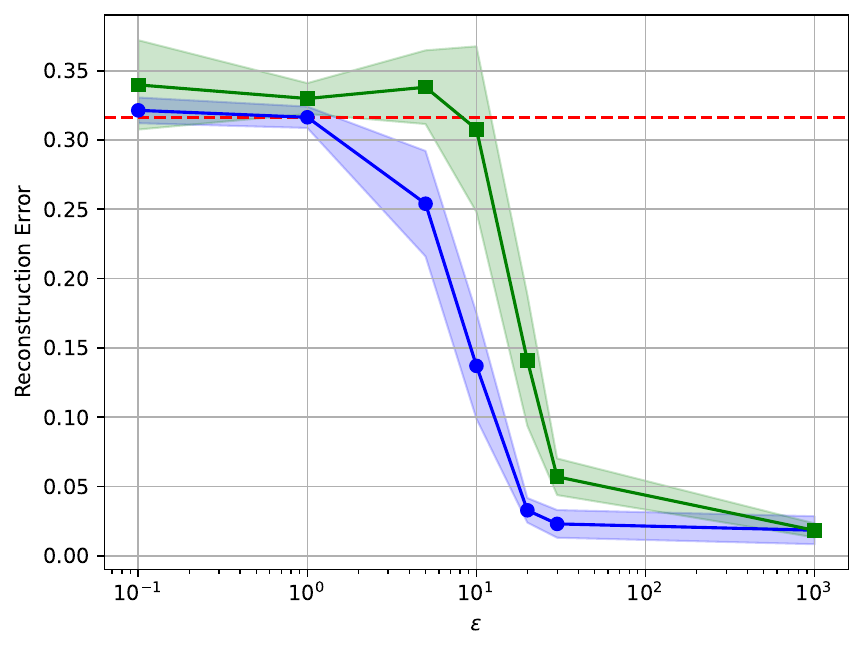}
    \caption{Default of Credit Card Clients dataset}\label{fig:results_informed_reconstr_default_credit}
  \end{subfigure}

\caption{\rebuttalsatml{Results of our reconstruction experiments in the \textbf{informed adversary} setup. We report the reconstruction error for the single unknown example as a function of the DP budget $\varepsilon$ used to train the DP RFs. In these experiments, all forests contain $\lvert \forest \rvert = 10$ trees of depth $5$. The reconstruction attacks are run iteratively over a random subset of the training examples, and the reported reconstruction errors are averaged across these examples and across $10$ random seeds. Standard deviations are also shown.}}\label{fig:results_informed_reconstr}
\end{figure*}

We observe that for privacy budgets $\varepsilon > 5$ (or $\varepsilon \geq 5$ for the UCI Adult Income and Default of Credit Card Client datasets), our attack is able to retrieve meaningful information about the unknown training example. For smaller budgets ($\varepsilon \leq 1$ for our attack, or $\varepsilon \leq 10$ for the RecoNN baseline), only distributional information is recovered, which is consistent with our observations in the full-dataset reconstruction setting (Table~\ref{tab:p-values}), as indicated by the fact that the reconstruction error is no better than that of the simple per-coordinate majority baseline.
These results highlight that for moderate ($\varepsilon \in \{5,10\}$) to large ($\varepsilon \geq 20$) DP budgets, our approach is able to recover meaningful information by thoroughly exploiting the DP RF parameters. The computational cost of our attack is also very low in this setting, with an average runtime of approximately 4 seconds (respectively, 9 and 13 seconds) for the COMPAS (respectively, Default of Credit Card Clients and UCI Adult Income) dataset.

\begin{figure*}[t!]
  \centering
  
\begin{subfigure}[t]{0.38\textwidth}
    \centering
    \includegraphics[width=\linewidth]{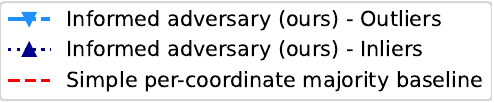}
\end{subfigure}

    \vspace{10pt}
    
    \begin{subfigure}[t]{0.48\textwidth}
    \centering\includegraphics[width=\textwidth]{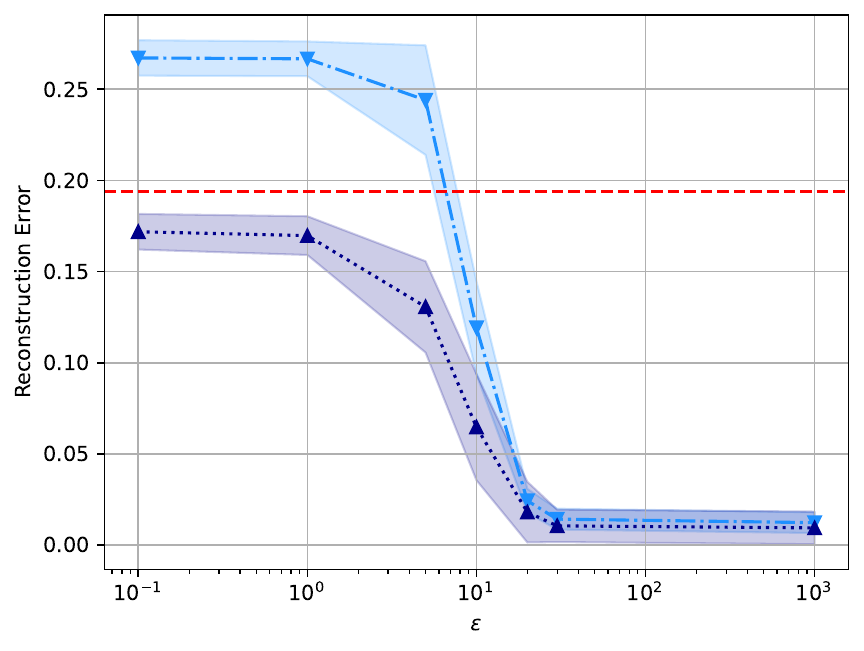}
    \caption{UCI Adult Income dataset}\label{fig:results_informed_reconstr_adult_outliers_ours}
  \end{subfigure}
    \begin{subfigure}[t]{0.48\textwidth}
    \centering\includegraphics[width=\textwidth]{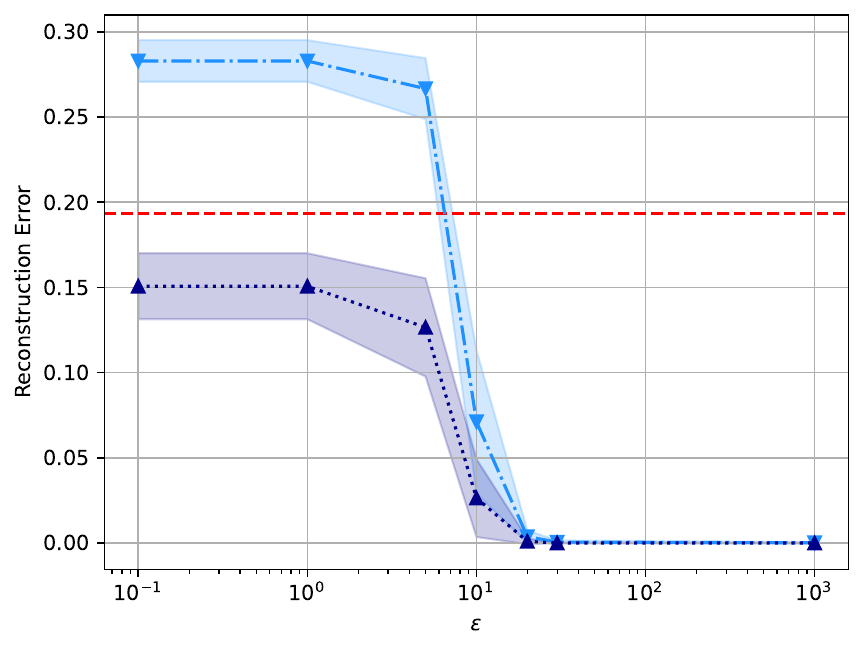}
    \caption{COMPAS dataset} 
    \label{fig:results_informed_reconstr_compas_outliers_ours}
  \end{subfigure}
  
    \vspace{10pt}
  
    \begin{subfigure}[t]{0.48\textwidth}
    \centering\includegraphics[width=\textwidth]{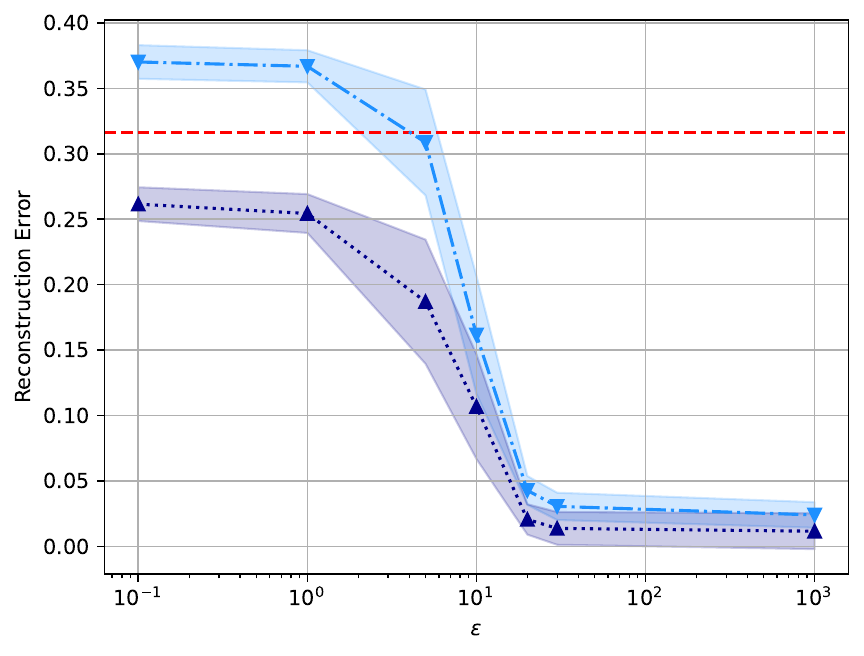}
    \caption{Default of Credit Card Clients dataset}\label{fig:results_informed_reconstr_default_credit_outliers_ours}
  \end{subfigure}

\caption{\rebuttalsatml{Results of our reconstruction experiments in the informed adversary setup. Compared to Figure~\ref{fig:results_informed_reconstr}, results are reported \textbf{only for our informed adversary}, with the average reconstruction error (and standard deviation) being computed and reported \textbf{separately for inliers and outliers}.}}\label{fig:results_informed_reconstr_outliers_ours}
\end{figure*}

\begin{figure*}[t!]
  \centering
  
\begin{subfigure}[t]{0.52\textwidth}
    \centering
    \includegraphics[width=\linewidth]{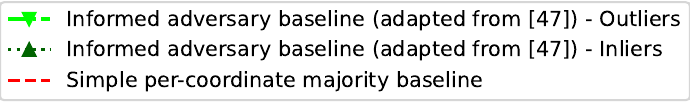}
\end{subfigure}

    \vspace{10pt}
    
    \begin{subfigure}[t]{0.48\textwidth}
    \centering\includegraphics[width=\textwidth]{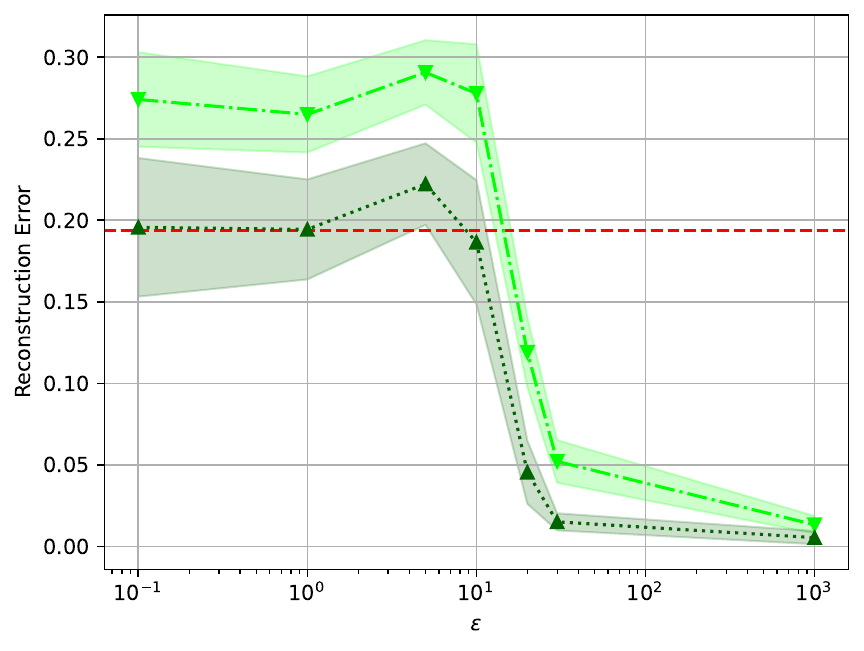}
    \caption{UCI Adult Income dataset}\label{fig:results_informed_reconstr_adult_outliers_baseline}
  \end{subfigure}
    \begin{subfigure}[t]{0.48\textwidth}
    \centering\includegraphics[width=\textwidth]{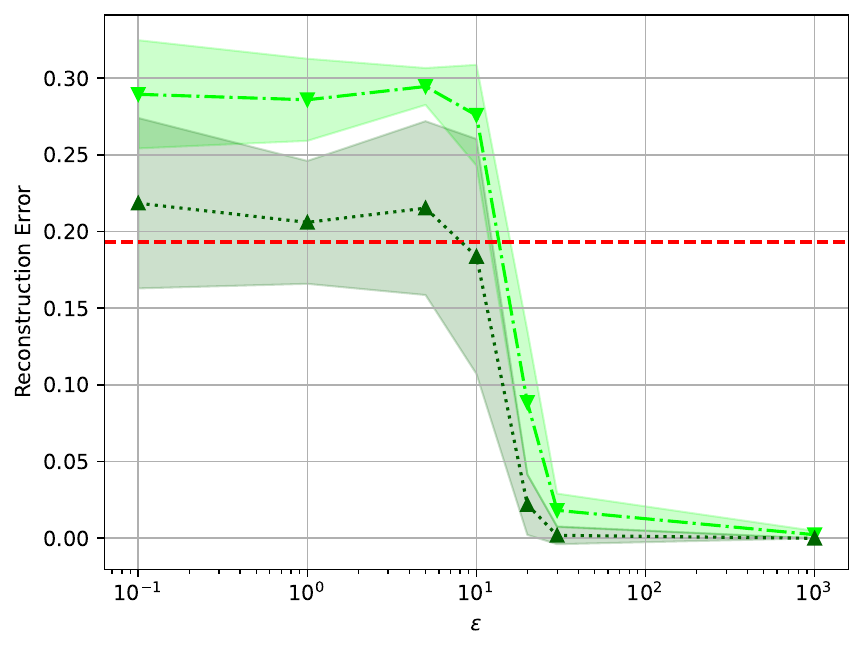}
    \caption{COMPAS dataset} 
    \label{fig:results_informed_reconstr_compas_outliers_baseline}
  \end{subfigure}
  
    \vspace{10pt}
  
    \begin{subfigure}[t]{0.48\textwidth}
    \centering\includegraphics[width=\textwidth]{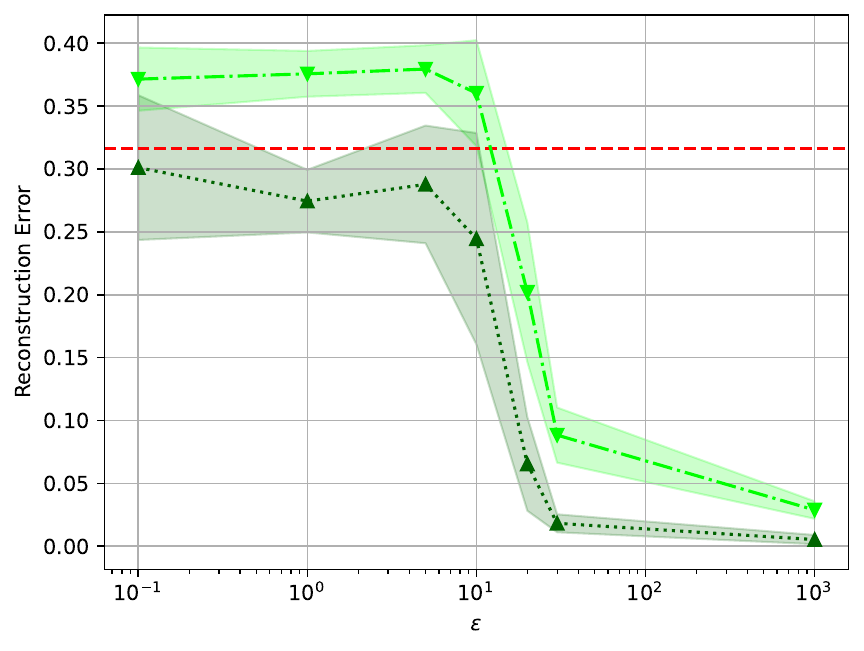}
    \caption{Default of Credit Card Clients dataset}\label{fig:results_informed_reconstr_default_credit_outliers_baseline}
  \end{subfigure}

\caption{\rebuttalsatml{Results of our reconstruction experiments in the informed adversary setup. Compared to Figure~\ref{fig:results_informed_reconstr}, results are reported \textbf{only for the baseline informed adversary~\cite{balle2022reconstructing}}, with the average reconstruction error (and standard deviation) being computed and reported \textbf{separately for inliers and outliers}.}}\label{fig:results_informed_reconstr_outliers_baseline}
\end{figure*}

\begin{figure*}[t!]
  \centering
  
\begin{subfigure}[t]{0.35\textwidth}
    \centering
    \includegraphics[width=\linewidth]{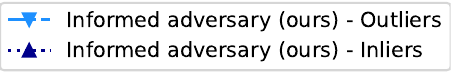}
\end{subfigure}

    \vspace{10pt}
    
    \begin{subfigure}[t]{0.48\textwidth}
    \centering\includegraphics[width=\textwidth]{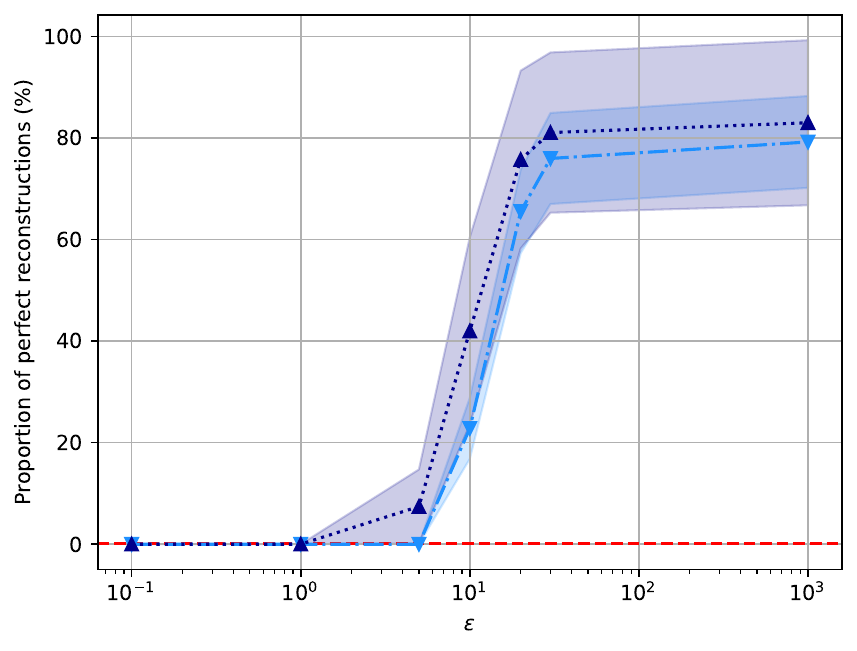}
    \caption{UCI Adult Income dataset}\label{fig:results_informed_reconstr_adult_outliers_nb_perfect_ours}
  \end{subfigure}
    \begin{subfigure}[t]{0.48\textwidth}
    \centering\includegraphics[width=\textwidth]{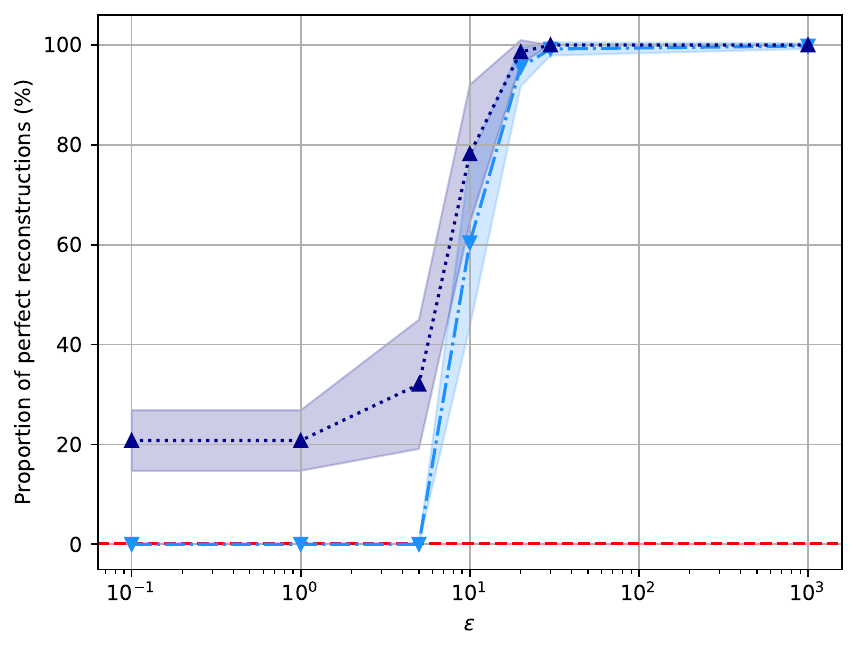}
    \caption{COMPAS dataset} 
    \label{fig:results_informed_reconstr_compas_outliers_nb_perfect_ours}
  \end{subfigure}
  
    \vspace{10pt}
  
    \begin{subfigure}[t]{0.48\textwidth}
    \centering\includegraphics[width=\textwidth]{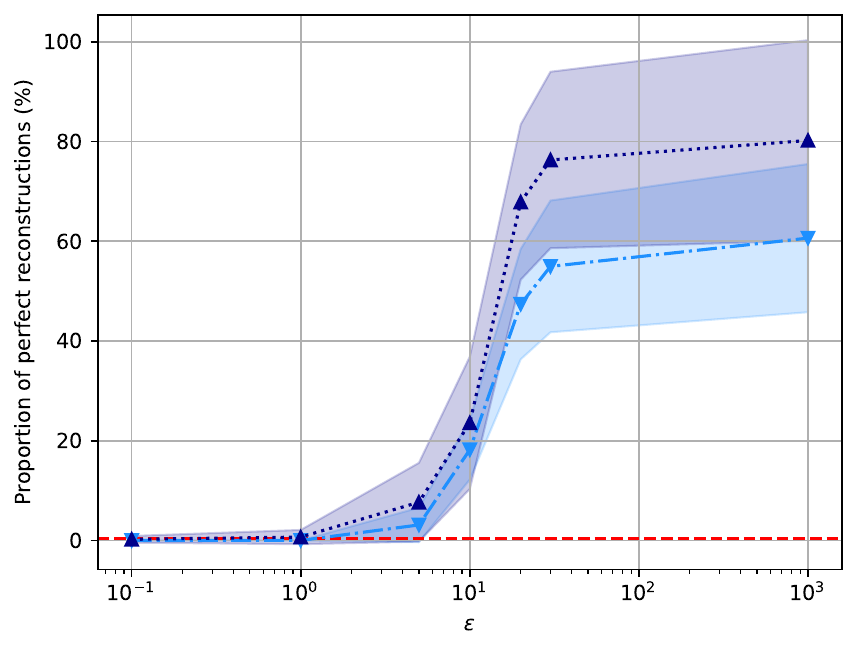}
    \caption{Default of Credit Card Clients dataset}\label{fig:results_informed_reconstr_default_credit_outliers_nb_perfect_ours}
  \end{subfigure}

\caption{\rebuttalsatml{Results of our reconstruction experiments in the informed adversary setup.  We report the average number of perfectly reconstructed examples as a function of the DP budget, for the \textbf{our proposed informed adversary~\cite{balle2022reconstructing}}, considering either \textbf{inlier or outlier unknown examples}.}}\label{fig:results_informed_reconstr_outliers_nb_perfect_ours}
\end{figure*}

\begin{figure*}[t!]
  \centering
  
\begin{subfigure}[t]{0.55\textwidth}
    \centering
    \includegraphics[width=\linewidth]{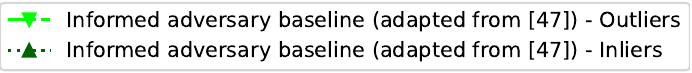}
\end{subfigure}

    \vspace{10pt}
    
    \begin{subfigure}[t]{0.48\textwidth}
    \centering\includegraphics[width=\textwidth]{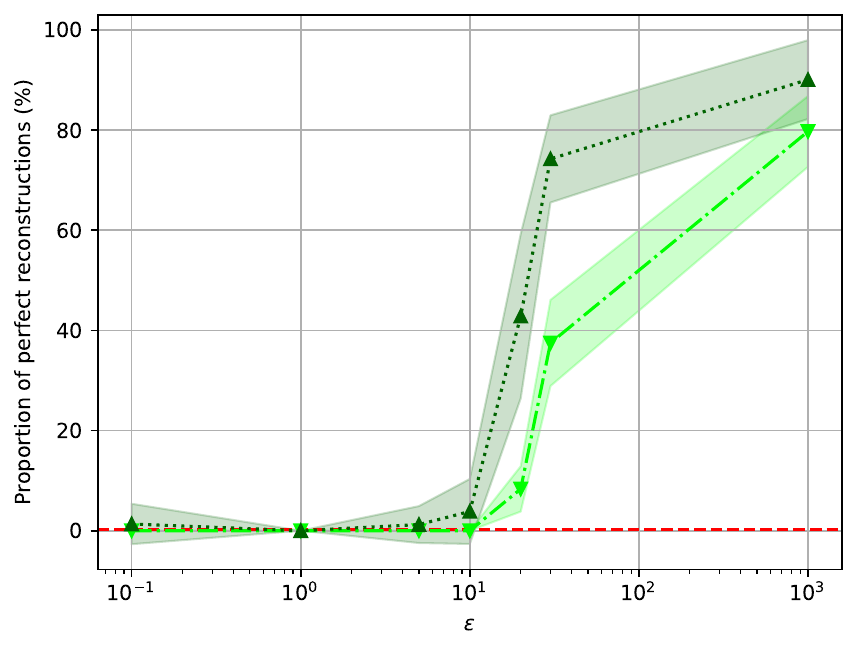}
    \caption{UCI Adult Income dataset}\label{fig:results_informed_reconstr_adult_outliers_nb_perfect_baseline}
  \end{subfigure}
    \begin{subfigure}[t]{0.48\textwidth}
    \centering\includegraphics[width=\textwidth]{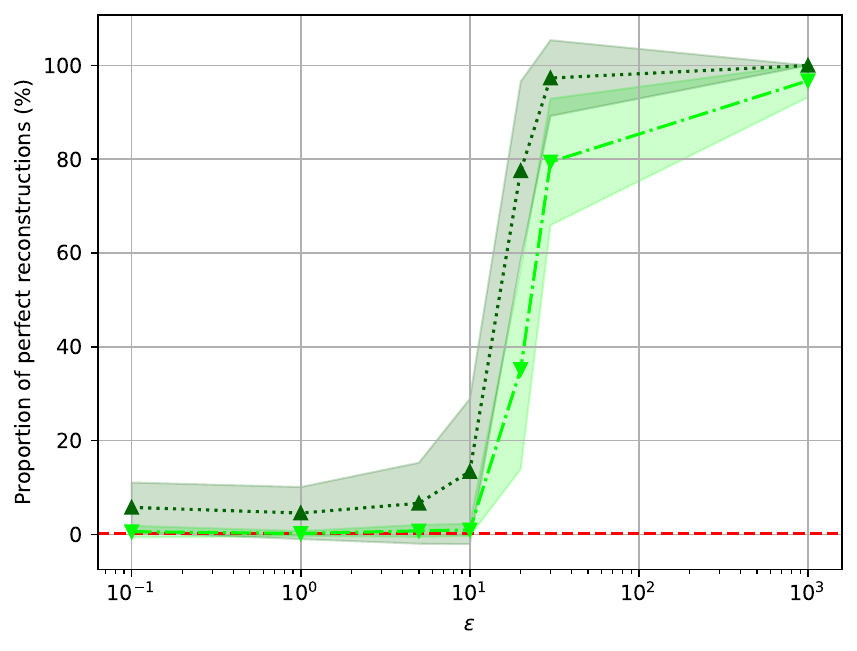}
    \caption{COMPAS dataset} 
    \label{fig:results_informed_reconstr_compas_outliers_nb_perfect_baseline}
  \end{subfigure}
  
    \vspace{10pt}
  
    \begin{subfigure}[t]{0.48\textwidth}
    \centering\includegraphics[width=\textwidth]{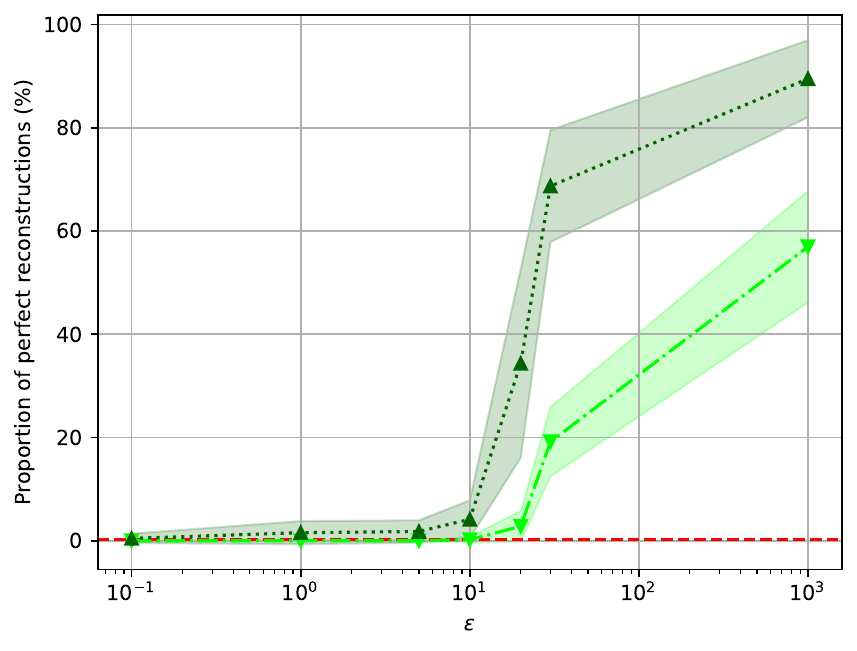}
    \caption{Default of Credit Card Clients dataset}\label{fig:results_informed_reconstr_default_credit_outliers_nb_perfect_baseline}
  \end{subfigure}

\caption{\rebuttalsatml{Results of our reconstruction experiments in the informed adversary setup.  We report the average number of perfectly reconstructed examples as a function of the DP budget, for the \textbf{baseline informed adversary~\cite{balle2022reconstructing}}, considering either \textbf{inlier or outlier unknown examples}.}}\label{fig:results_informed_reconstr_outliers_nb_perfect_baseline}
\end{figure*}

Figures~\ref{fig:results_informed_reconstr_outliers_ours} and~\ref{fig:results_informed_reconstr_outliers_baseline} refine these results by reporting them separately for inlier and outlier examples, for our informed adversary and the informed adversary baseline adapted from~\cite{balle2022reconstructing}. As detailed in Section~\ref{subsec:results}, each training example was classified as either inlier or outlier using isolation forests as implemented in the \texttt{scikit-learn}~\cite{scikit-learn} Python library, with their default hyperparameters. 

The observed trends are consistent with \textbf{Result~5} (and Table~\ref{tab:individual_reconstruction_results}) from Section~\ref{subsec:results} in the full reconstruction setup. Indeed, even in the informed adversary setting, the average reconstruction error is consistently lower for inliers. This can be explained by the fact that the distributional information extracted by the attack facilitates the reconstruction of inliers, while offering only limited benefit for outliers.
The same qualitative behavior is observed for the baseline informed adversary.

These results further confirm our previous observations. For privacy budgets $\varepsilon > 5$ (or $\varepsilon \geq 5$ for the Default of Credit Card Clients dataset), our attack can recover meaningful information about the unknown training example, even when it lies outside the data distribution. This indicates that, in these privacy regimes, the attack infers example-specific information beyond distributional patterns.

The superiority of our adversary is also apparent, as the baseline is only able to extract non-trivial information about outliers for substantially larger privacy budgets ($\varepsilon \geq 20$). The reconstruction error of outliers also quickly converges towards that of inliers (close to 0) (for $\varepsilon =20$ or $\varepsilon =30$, depending on the dataset) with our proposed attack. With the considered baseline, this convergence is considerably slower (as even with $\varepsilon = 1000$, a small gap remains). 

For smaller privacy budgets ($\varepsilon < 5$), the inferred information remains purely distributional and therefore only helps the reconstruction of inliers for both adversaries, as expected.

Figures~\ref{fig:results_informed_reconstr_outliers_nb_perfect_ours} and~\ref{fig:results_informed_reconstr_outliers_nb_perfect_baseline} confirm these trends by reporting the proportions of inlier and outlier examples that are perfectly reconstructed for different DP budgets, for our informed adversary and the informed adversary baseline adapted from~\cite{balle2022reconstructing}. In particular, our informed adversary is often able to perfectly reconstruct the unknown example even when it lies outside the data distribution and when the DP budget is moderate, and the proportion of perfectly reconstructed unknown examples increases rapidly with the DP budget~$\varepsilon$.

Altogether, these results show that tight DP budgets are required to prevent DP RFs from leaking dataset-specific patterns, including information about out-of-distribution examples, which is consistent with our findings in the full reconstruction setup presented in the main paper. They further indicate that the DP budget $\varepsilon$ required to prevent successful reconstruction in the informed adversary setting depends on both the dataset and the attack considered. This motivates the use of our informed adversary as a meaningful vulnerability audit, as it models all the information provided through the DP RF structure and parameters, and empirically outperforms the baseline attack.

}

\end{document}